# PATHWAY TO FUTURE SYMBIOTIC CREATIVITY

BUILDING PLATFORM TECHNOLOGIES FOR
SYMBIOTIC CREATIVITY IN HONG KONG

August 2023

# Contents

# PATHWAY TO FUTURE SYMBIOTIC CREATIVITY

## INTRODUCTION

Artworks, such as paintings, drawings, sculptures, pottery, music, dance, photographs and videos, seek to communicate universal emotions. Artists create art as an attempt to disseminate their self- expression and interpretation of their existence and environment. This sentiment was expressed by Van Gogh as 'I want to paint humanity, humanity and again humanity'. The process of art creation involves an interaction (submission, reception and response) between the artist and spectator. Since the creation of the first cave drawings, artists engaging in various art forms have expanded their means of expression. Collaborating and co-creating art with artificial intelligence (AI) has emerged as the next step in this endeavour.

The first question that emerges is 'Why should AI-based art be promoted?' Two apparent reasons can be identified. The science involved in developing AI capable of artistic creation can enhance our understanding of the cognitive processes involved in creative act and extend the boundaries of computer science. We do not envisage AI art as a replacement of the art created by humans. Instead, AI can be viewed as a supplement to the methods and tools available to human artists. Thus, we believe that the future of art may be grounded in human–machine symbiotic creation.

AI-based machine creators represent technologies that can produce their own representation of a concept. The art created by an AI agent will depend, similar to that for a human artist, on its learning method and the artefacts encountered in training. The receiving audience, as for any piece of art, will play a central role in deciding whether the AI creation is art or only a machine- created artefact with no artistic value. One central difference between an AI creator and an artist is the origin of its experiences. An artist will live and develop creativity as a response to events surrounding him/her in uncertain natural and societal environments. The art created by an artist is a commentary on the human reaction to these surroundings. In contrast, an AI creator is trained in an environment defined by humans. Traditional machine learning methods exhibit limitations when they are directly used for art content generation. Works of art make rules, which are formulated in attempts to capture aspects of the ontic open world surrounding the artist. Rules do not make works of art. This aspect represents a paradox. Traditional machine learning is aimed at automatically extracting rules, specifically, to establish models trained by big data. In this scenario, can the output of a model be artistic? This question has been the subject of extensive discussion across multidisciplinary domains.

A straightforward but oversimplified view of AI-based art creation is to view AI as a tool. If an artist determines the environment and learning rules of the AI creator, AI is simply a tool available to the artist. Similar to training his/her hand to wield the brush to paint, an artist can train AI to produce a unique work of art that reflects its creator. Recent advancements in AI, especially generative learning technologies, have demonstrated the autonomous creativity of AI. A frequently quoted example of AI-based art creation is the Portrait of Edmond de Belamy, which has been sold for USD432,000. This painting was created by training a generative adversarial network with a dataset of 15,000 portraits covering six centuries. This example demonstrates that to create art using machine learning models, extensive non-random exploration must be performed under the constraints of certain aesthetic logics. In this context, a machine creator goes beyond its role as a tool, to serve as a companion. The current era can be considered that of symbiotic creativity, in which machine and human co-create artefacts.

Inspirited by Simon Colton's hierarchy of creative systems [1], creative systems can be classified into 7 levels, 5 of which are as follows:

- **Generative system:** a system that takes certain inputs to generate varying outputs. The system parameters can be changed to alter the generated outputs.

- **Appreciative system:** a generative system that can discern the output quality, allowing aesthetic preferences to be encoded into a utility (fitness function) through a measurement matrix associated with aesthetic values such as novelty and surprise.

- **Artistic system:** an appreciative system that can invent its own aesthetic fitness functions and use them to filter and rank the outputs that it generates. Thus, this artistic system can affect the world artistically.

- **Symbiotic system:** an artistic system that can communicate with humans regarding aesthetic generation, thereby influencing human appreciation of a work of art through explanations and high-quality, surprising outputs (machine-to-human influence). Moreover, the system can learn from the human and environment to produce important interesting, and unpredictable outputs (human-to-machine inspiration).

- **Authentic system:** a symbiotic system that can remember its life experiences of creation and human appreciation through technologies such as sensors detecting human reactions and improved in situ and online human–computer interactions to record human/machine interactions with respect to a creation. The system can operationalise life experiences and outside knowledge into opinions that can be reflected in the generative processing and output.

Unlike Colton's seven-level hierarchy, we emphasise the coherent inheritance of key features. All creative systems are generative. An artistic system must be appreciative to be able to apply its own aesthetic preferences. A symbiotic system must be artistic to communicate with humans to gain insight and inspiration for developing and evolving its aesthetic fitness functions. Moreover, the experiences remembered by an authentic system pertain to its interactions with humans for co-creation. Thus, the corresponding art generation process is inherently symbiotic. This view reflects our belief that creative systems are evolving from mimicking human artists to being artists **(Figure 1)**. This shift reflects the connection between machine creativity and human–machine interaction: machine creativity increases with the enhanced richness of human–machine communication regarding aesthetic values during the creation and appreciation phases.

In the first three levels, an AI creative system acts primarily as a passive learning system, focusing on producing art by mimicking human artists. The success of creation by such systems can be tested based on the Turing test principle: if a human cannot identify whether an artwork has been created by a human or machine, we can claim that the machine acts as an artist. In this context, artistic systems can be considered Turing artists. Significant progress has been made in establishing Turing artist systems, as discussed in Section 2.

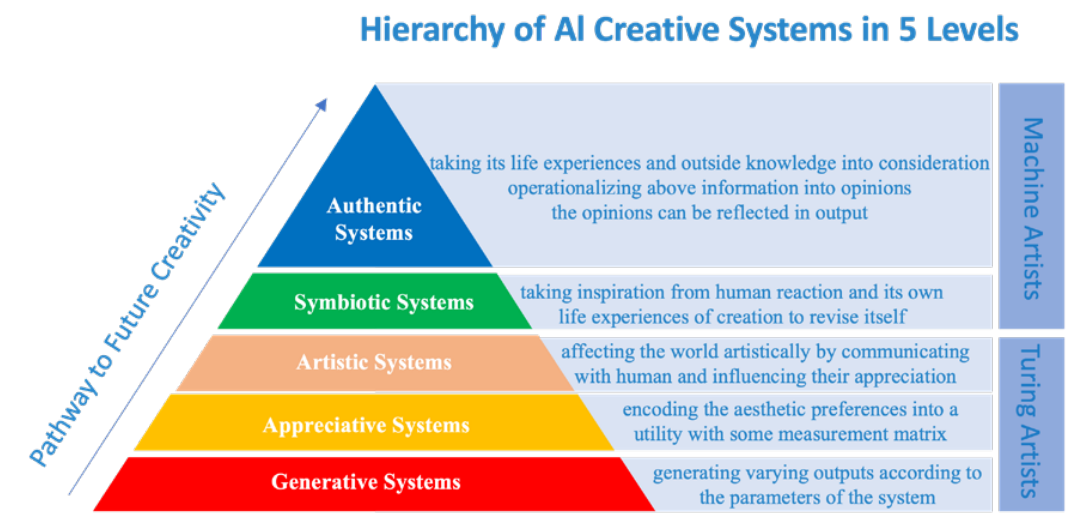

**Figure 1** Five-level hierarchy of AI creative systems.

Systems in the next two levels emphasise the communication between machine and human in the art creation process. In the realm of artistry, the harmonious collaboration between machines and humans necessitates a profound comprehension of the intricate tapestry of human mental states encompassing desires, interests, and emotions. In addition, humans must understand the machines' creative capabilities and limitations. In AI, such a human–machine bidirectional understanding is known as 'value/mental state alignment'. This bidirectional alignment is crucial to the success of human–machine collaboration. Specifically, the core concept for symbiotic and authentic systems is human–machine value alignment in aesthetics, or aesthetic alignment.

For a symbiotic system to have a profound impact on human cognition, it must possess the ability to effectively convey the aesthetic worth of its creations to humans. Moreover, it must be inspired by the human environment to continuously enrich its aesthetic knowledge and creativity. With the enhancement of the ability to remember and learn from its life experience of creative activities interacting with the human recreative environment, an authentic system is expected to have self-evolving aesthetic value, artistic style and expressive creativity. The systems in the last two levels share a distinguishing feature of communicating with human through aesthetics alignment to enable a symbiotic art creation paradigm. Thus, we term these systems machine artists.

The accelerated progress of immersive visual environments, and their evolution into the metaverse, has facilitated the creation of symbiotic art. This is due to the enhanced flexibility of bidirectional communication between artists and the manifestation environments of their art. Therefore, the development of art manifestation environments, especially in terms of the evolution of human and art interaction within the framework of the metaverse, must be reviewed to provide insights into the future development of symbiotic art creation.

This report is divided into the following parts: Part 1 presents a comprehensive review of the development of Turing artists. Part 2 discusses the human–machine interaction in art creation. Part 3 describes a framework for building machine artists. Part 4 provides an overview of a non-fungible token (NFT)-based art economic model. Part 5 discusses the ethical issues of future AI-based art generation systems.

# *Part 1*
# The Journey to Developing Turing Artists

## 1 Generative Systems: Mimicking Artifacts

Machine learning can be divided into two major paradigms: discriminative modelling and generative modelling. Taking image recognition as an example, discriminative modelling is used for classification tasks, such as object identification to identify whether a certain object (e.g. a car) is present in an image, whereas generative modelling learns about the appearance of a set of related images (e.g. pictures of cars) from a training dataset and uses the learnt knowledge of the distribution of pixels to generate novel images in the same category (e.g. new car designs). Operating on a foundation of probabilistic principles, generative modeling assumes the existence of an enigmatic probability distribution that elucidates the underlying reasons for the prevalence of specific images within the training dataset. The objective of generative modelling is to closely approximate this distribution and draw samples from it, ultimately generating a set of pixels that closely resemble those found in the original training dataset. Thus, generative modelling has the capability required for creation: that of generating new and unpredictable outputs from a learning environment.

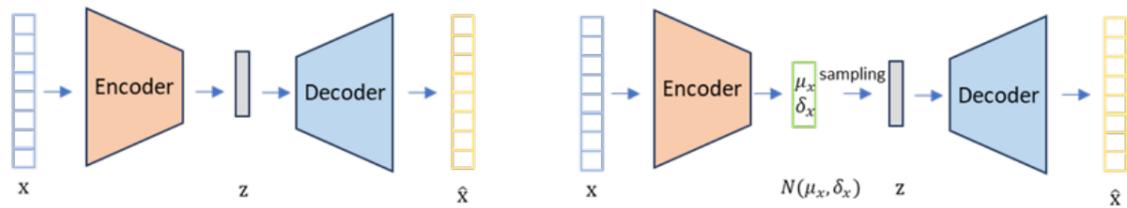

**Figure 2** Illustration of auto-encoder and variational auto-encoder (VAE)

Perhaps a basic generative deep learning method is auto-encoder. An auto-encoder is a two-part neural network composing of encoder and decoder. The encoder has the task of compressing raw input data into a hidden representation vector, while the decoder works to reconstruct the original domain by expanding the given representation vector through decompression **[2-3]**. The aim of such a system is to learn a proper compressed latent representation that can accurately reconstruct the input as the output. For art creation, an auto-encoder can be used to learn a compressed latent representation space from a collection of artefacts, such as paintings. Upon completion of the learning process, it becomes feasible to select any point within the hidden space and employ the decoder for generating fresh and innovative data. Such technology provides a tool for building Generative System. Direct reconstruction can hardly be regarded as creation, and the auto-encoder architecture can be refined to add creativity. The variational auto-encoder (VAE) **[2,4]** (compared with an auto-encoder in **Figure 2**) is a good example of such refinement. The encoder employs a mapping function that associates each input with a multivariate normal distribution. Conversely, the decoder obtains latent space data point by sampling from the encoded Gaussian distribution. Through the utilization of this approach, the decoder demonstrates its capability to generate a sample even when an unobserved point within the latent representation space is selected. This attribute enables higher creative freedom in the generation process.

The endeavor of generating paintings utilizing VAEs is commonly categorized as an image generation task, wherein models are trained on extensive databases of paintings procured from platforms including Web Gallery of Art **[5]**, WikiArt **[6]**, Rijksmuseum Challenge **[7]**, Art500k **[8]**, and OmniArt **[9]**. However, despite the theoretical elegance of VAEs, they have a tendency to generate blurry images due to the model's primary objective of approximating the overall distribution of the training data. To overcome this problem, besides the adversarial models to be introduced later, variants of VAE, such as IntroVAE **[10]** and VQ-VAE **[11]**, are proposed, which introduce an introspective process to estimate the differences between the real and generated images or learn discrete latent variables to improve the reconstruction ability of the decoder.

In addition to VAEs, alternative models have been developed, including PixelRNN **[12]**, PixelCNN **[13]**, and their variants **[14]**, **[15]**, which allow for direct autoregressive sampling from the dataset. PixelRNN utilizes a recurrent neural network (RNN) to generate pixels sequentially, moving from one corner to the corresponding corner on the diagonal. Meanwhile, PixelCNN speeds up the generation process by leveraging faster and parallel convolution operations during training.

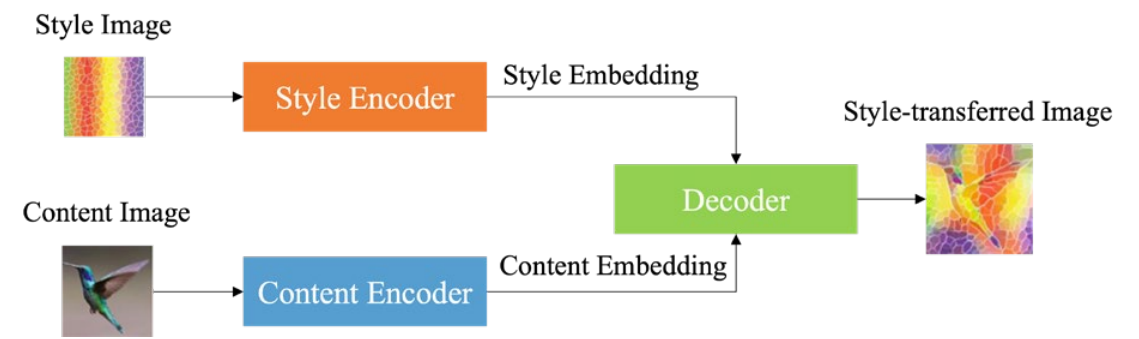

**Figure 3** Diagram of image style transfer

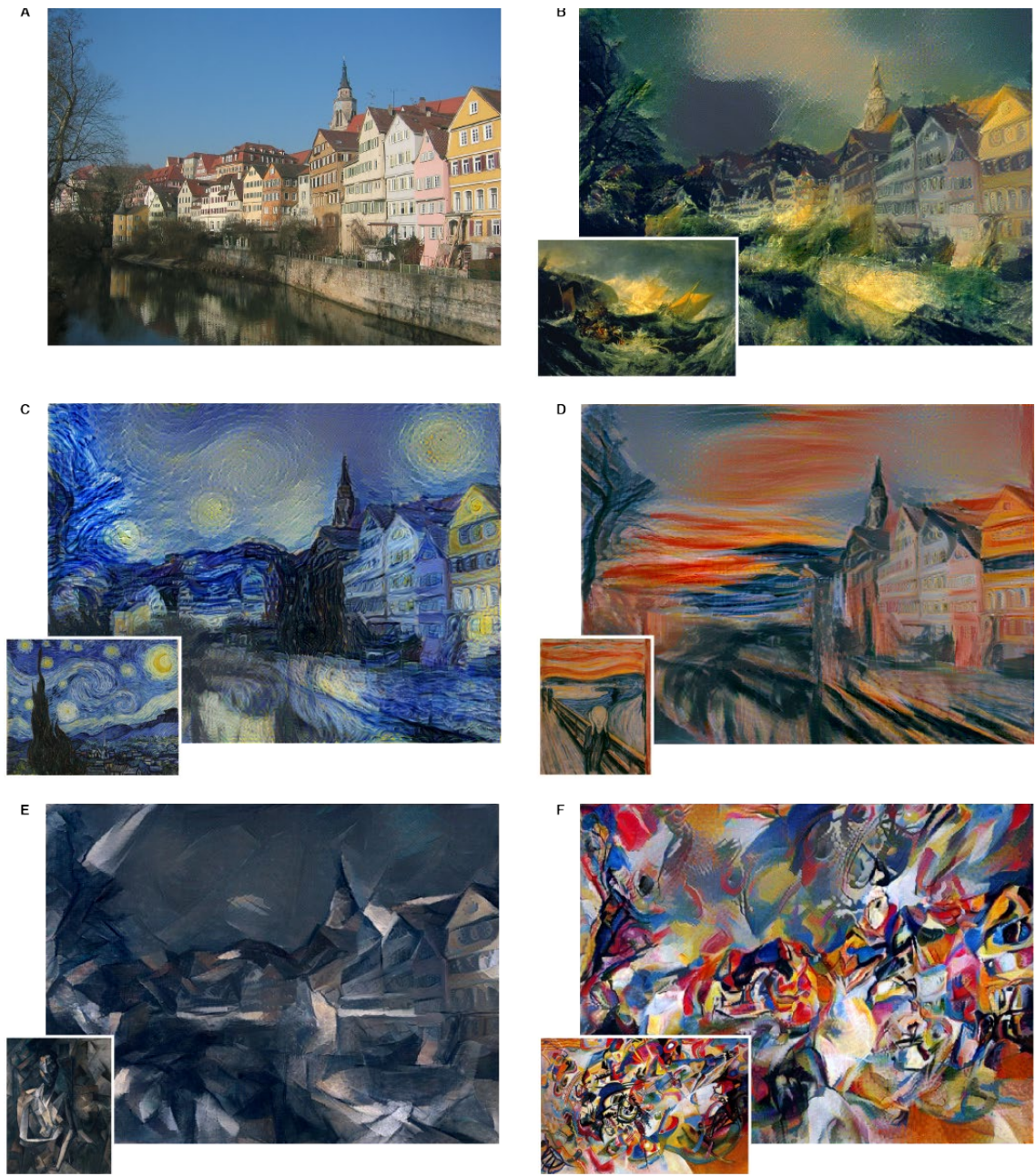

**Figure 4** Image style transfer examples in [16]

Another common method for producing paintings with artistic styles is performing style transfer. As illustrated in **Figure 3**, the model extracts the content and style information from a realistic source image and an artistic image, respectively, using two separate encoders, and then combines this information in decoding to a new image. In **[16]**, Gatys proposed a convolutional neural networks (CNN) based method that yields remarkable results, as shown in the examples in **Figure 4**. Subsequently, instance normalisation **[17]**, **[18]** was proposed to achieve fast stylisation by removing instance-specific contrast information from the content image.

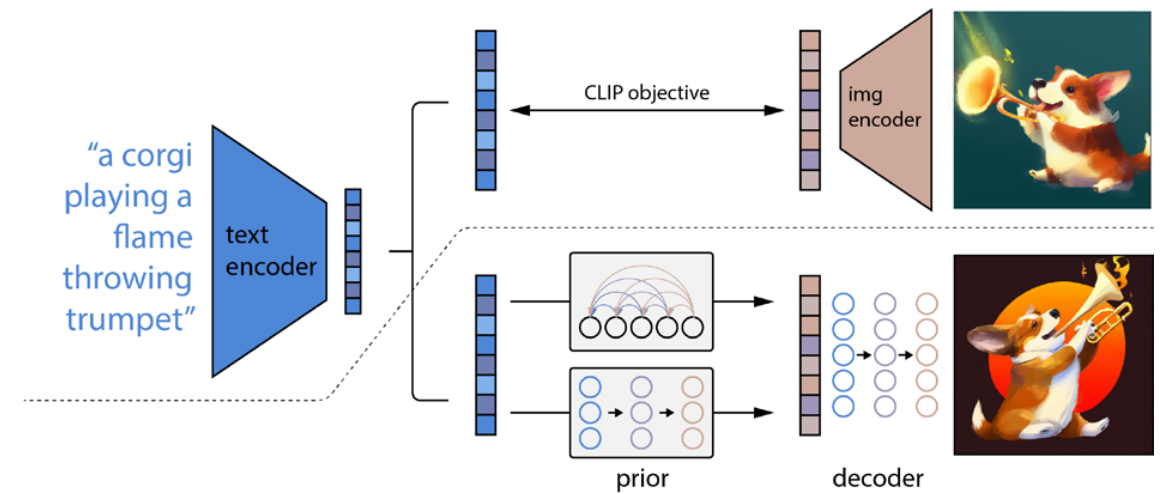

**Figure 5** Establishing congruity between textual and visual mediums **[164]**

The most recent methodologies premised on computational models exhibit a capacity to generate artistic illustrations in a manner congruent with human intent, providing a means for individuals to articulate precise attributes of the envisaged artwork at a microscopic level. **Figure 5** elucidates this process wherein the computational model **[164]** establishes congruity between textual and visual mediums by leveraging the training methodology underpinning the CLIP model **[165]**. During the inferential phase, the guide text is processed through a sequence involving an image encoder, a pre-trained model, and subsequently, an image decoder, culminating in the formation of an image that accurately represents the original instruction. Comparable methodologies in the domain of text-to-image translation encompass Textual Inversion **[166]**, DreamBooth **[167]**, Custom Diffusion **[168]**, E4T **[169]**, and InstantBooth **[170]**.

In this period characterized by an increasing convergence of artistic endeavors and AI—exemplified in domains such as automated image synthesis and collaborative fiction authoring—the field of AI-enabled music remains comparatively underdeveloped, particularly with regard to music comprehension. This is manifested in the minimal explorations related to profound musical representations, the paucity of extensive datasets, and the lack of a universally accepted, community-centric standard of evaluation.

Generative models have also been widely adopted for creating musical works, focusing on

(a) composing sheet music with long-term meaningful structures, and

(b) synthesising singing voices to generate virtual singers whose voices can be flexibly controlled.

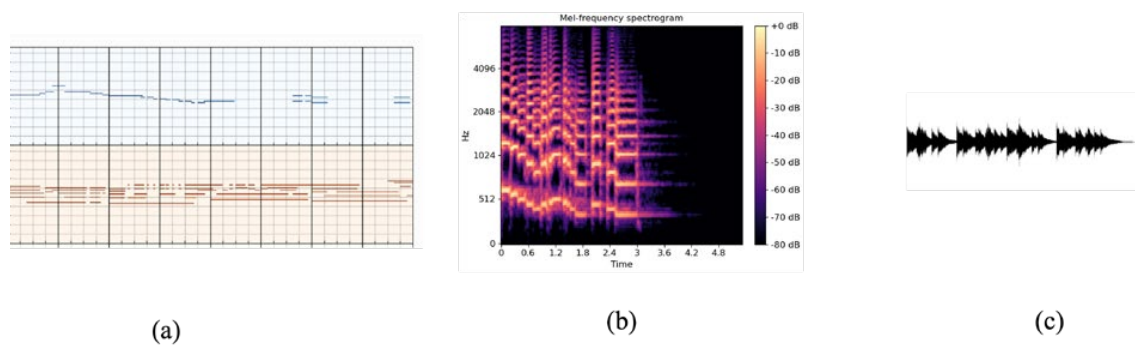

Figure 6 Different music formats: (a) symbolic piano roll, (b) Mel spectrogram, (c) waveform

Conventionally, music composition has been mainly studied within the framework of symbolic representation. To facilitate processing, the musical notes are normally expressed by the piano roll shown in **Figure 6a** in the MIDI standard, which includes the timing, pitch, velocity and instrument information, thus are regarded as a special case of natural language. Music composition is therefore coped by a symbol sequence generation, and the methods modelling word dependencies in text generation can be exploited for this task. The traditional VAE generative model served as the foundation for the development of MusicVAE **[19]**. This innovative variant incorporates smoothness constraints within the latent space and employs a novel hierarchical decoder, enabling the generation of musical outcomes that possess both localized details and coherent long-term structures. Given the success of recurrent neural networks (RNNs) in sequential modelling, numerous autoregressive techniques have been proposed for generating sheet music **[20]**-**[23]**, including hierarchical RNN, DeepBach, melody-RNN, and Anticipation RNN.

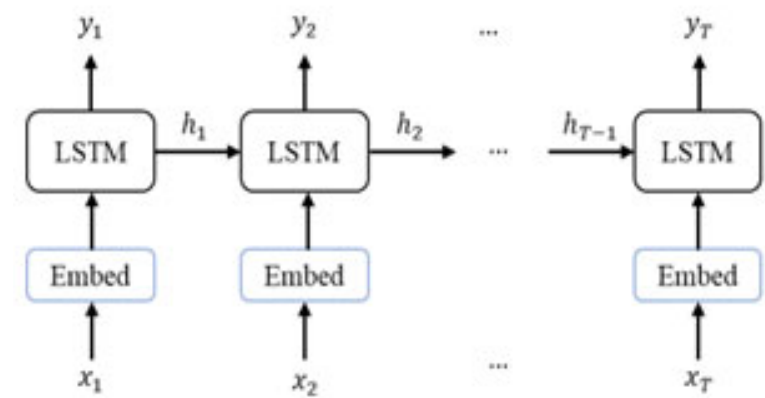

Figure 7 Serial autoregressive generative processes in the RNN model. The Long-Short-Term-Memory (LSTM) model is used.

As depicted in **Figure 7**, the main problem of autoregressive generation is the low processing speed, as the model must sequentially rely on previously generated tokens to produce new tokens. As an alternative to the RNN for sequential modelling, a transformer uses the attention mechanism for adaptive weighting over previous information. Specifically, for each token, by using linear layers, three embeddings that indicate the "Query", "Key" and "Value" of the token are obtained respectively, and the adaptive weights for each "Value" embedding are computed as the scaled dot-product of the "Query" and "Key" embeddings. The final output of the transformer is the weighted sum of the "Value" embeddings, and such processing can be duplicated to form a multi-head attention mechanism. With no autoregressive processing required, the transformer enables parallelisation during training. Accordingly, music composition methods based on non-regressive sequential modelling using transformers have been developed **[25]**, **[26]** recently, with the advantages of a longer sequence inception ability and an accelerated training and inference processes.

The symbolic approach to music composition makes the modelling problem easier by positioning it in a low-dimensional abstract space. However, it constrains the generated notes to fixed types of instruments, and fails to model the highly non-stationary dynamics, subtle changes, timbres and sound decays in different chambers that are typically expressed in the audio domain. Non-symbolic approaches have therefore been studied, in which the music is directly produced in an audio format. There have been some achievements in generating piano pieces in the audio domain **(Figure 6c)** **[27]**, **[28]** or in the spectrogram domain **(Figure 6b)** **[3]**.

The main obstacle encountered when modeling raw audio directly lies in the intricate web of exceedingly long-range dependencies, which introduces significant computational complexities in learning the intricate high-level music semantics, thus posing a substantial challenge. To mitigate this issue, one possible solution is to learn a lower-dimensional audio encoding sacrificing some unimportant information, while preserving majority musical information. This approach could generate short pieces with limited instrument types **[29]**, **[30]**. The AI composing system can be trained using an autoregressive Sparse Transformer **[24]**, **[31]** over this embedding space, where an autoregressive up-sampler is adopted to recreate the lost information at embedding to produce songs from a diverse range of musical genres, such as rock, hip-hop and jazz. A notable work following such a design is Jukebox **[32]**, which uses a hierarchical vector quantised VAE architecture and directly generates the music in an audio format, conditioned by genre and timing control.

The virtual singer is created by singing voice synthesis (SVS) techniques, which aim to generate high-fidelity, natural and expressive voices given explicit controlling factors, such as lyrics and musical notes, or by imitating the singing of another person. The latter case is specially termed as singing voice conversion (SVC). Typically, both SVS and SVC exploit an acoustic model to convert the linguistic or phonetic information into the audio Mel-spectrogram, and further rely on a vocoder to generate the audible waveforms. RNN based generative models, such as Tacotron **[33]**, DurIAN **[34]**, and the self-attention based ones including TransformerTTS **[35]**, FastSpeech **[36]**, AdaSpeech **[37]**, have become the mainstream approaches for building acoustic models, which tackles the problem using seq2seq modelling. VAE-based acoustic models are also developed, including GMVAE-Tacotron **[38]** and VAE-TTS **[39]**. It should also be noted that recently, diffusion models including Diff-TTS **[40]** and PriorGrad **[41]** have shown promising results in terms of voice quality. Many widely used vocoders also apply RNNs to model the temporal dependencies of waveforms: typical works include SampleRNN **[27]**, LPCNet **[42]** and WaveRNN **[43]**. Flow **[44]**, VAE **[45]** and diffusion models **[41]** have also been explored to design the DNN based vocoders.

For SVC, key problems also include

(a) the unavailability of parallel training data for the same content performed by different singers,

(b) missing labels for content and singer identity information, and

(c) achieving high controllability. With these difficulties in mind, we propose a new framework for unsupervised SVC as shown in **Figure 8**. Through the utilization of pretrained models, it becomes possible to extract phonetic, speaker, and pitch embeddings, which can then be employed to fully reconstruct the Mel spectrogram in an unsupervised manner. Flexible control of the pitch contour and voice timbre can be achieved by modifying the pitch and speaker embedding.

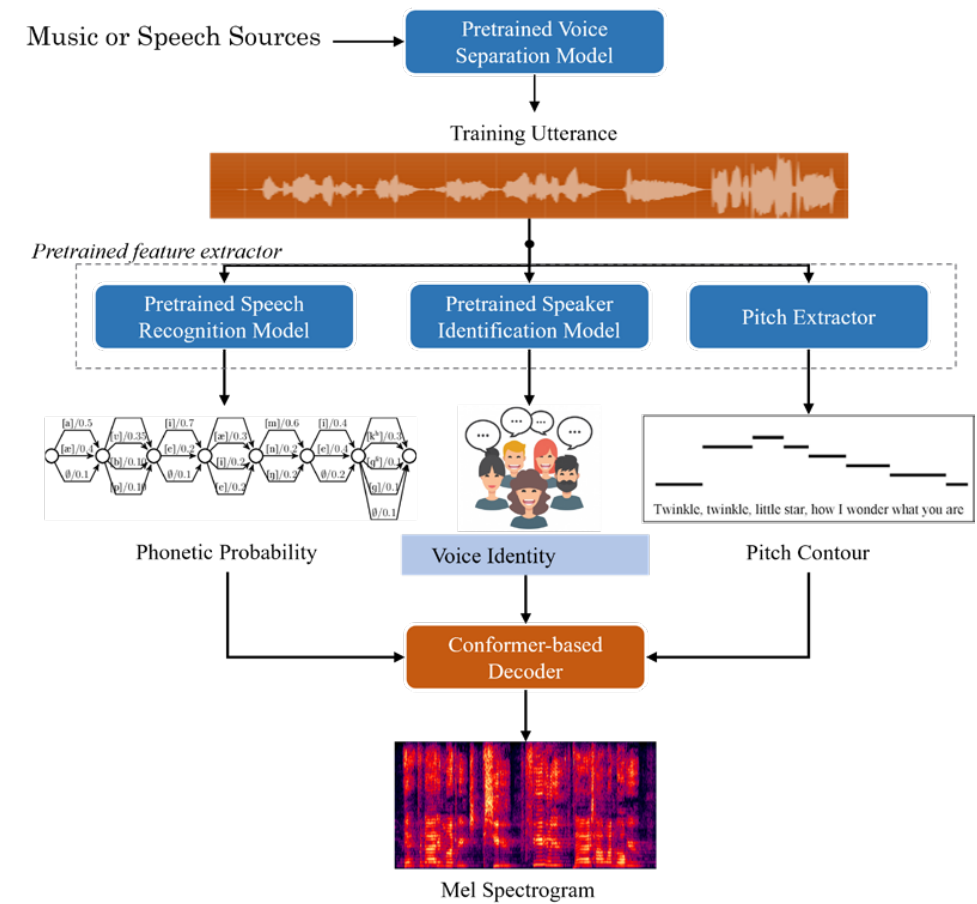

**Figure 8** A framework for unsupervised controllable singing voice conversion

Pre-trained language models (PLMs) have garnered the capability to formulate generalizable representations of data in a self-supervised learning (SSL) manner, thereby instigating impressive enhancements in the performance of natural language processing and associated domains **[171][172][173]**. Music, as an extraordinary linguistic medium facilitating cross-cultural communication **[174]**, and its inherent congruity with language as a communicative medium, constitute a promising foundation for the adaptation of PLM-oriented methodologies to model musical sequences. We propose the advantages of such an approach are dual fold. Firstly, PLMs could potentially set the groundwork to consolidate the modeling of an extensive array of music comprehension tasks, including music tagging, music transcription, beat tracking, source separation, and others, collectively referred to as Music Information Retrieval (MIR) tasks. This would eliminate the need for individualized models or features for distinct tasks. Secondly, the introduction of a PLM for music comprehension could re-distribute musical knowledge, bypassing the distribution of the data itself. This approach would circumvent the exorbitant costs of manual annotation and copyright constraints.

Scholars are exploring the interdisciplinary field of Music Information Retrieval (MIR) with the aim of developing a standard model for understanding music. The goal of MIR is to automatically decode data from raw musical audio **[175]** to make it easier to accomplish tasks such as categorizing music, recognizing emotions, estimating pitch, and analyzing musical elements like rhythm, melody, and harmony. Difficulties like copyright issues and the expense of annotation result in smaller labeled music datasets, which restricts the efficiency of supervised models. Given the demonstrated effectiveness of self-supervised learning (SSL) across a range of tasks with limited annotated datasets (such as NLP **[176]-[178]** and CV **[179]**), extensive research has been conducted on SSL-based audio representation learning **[180]-[186]** and music pre-trained models **[187]-[196]**. Established benchmarks like GLUE **[197]**, SuperGLUE **[198]**, and ERASER **[199]** in NLP, and VTAB **[200]** and VISSL **[201]** in CV, have been instrumental in promoting SSL-related research topics in their respective fields. Nevertheless, the evaluations of existing music models are fragmented and varied, lacking a comprehensive set of benchmarks. This gap makes it challenging to objectively compare and extract knowledge across different methodologies.

One of Badiou's "Fifteen Theses on Contemporary Art" wrote: "Art is the process of a truth, and this truth is always the truth of the sensible or sensual, the sensible qua sensible". "Sensibility" here refers to the formal beauty of the imitation of reality and includes the emotional connection between people and the world. AI generators not only learn the beauty of the fusion of various modal forms of art from a scientific point of view but also acquire the ability to express the beauty of ideas through learning. It is foreseeable that the future development trend of art lies in the digital and multi-modal artforms. If so, we need to start with the visual, auditory, tactile, olfactory, gustatory and interactive means of multi-modal sensory theory to study and explore how to construct a new form of multi-modal art. However, computationally modelling the human emotions evoked by art is a challenging problem.

Combined visual and auditory presentation is a multi-modal artform that is particularly suited to expressing human thoughts and emotions. The emotions and thoughts that the visual artist wants to evoke can be conveyed to the audience intuitively, and the emotions can be strengthened by the impressive channel of hearing. Different people may interpret the same artwork from different perspectives, based on their familiarity with the art style, formative experiences and current mental state and emotions. previous works such as **[46]**, **[47]**, and **[48]** have explored cross-modal art generation methods, however, the artworks generated by such models do not obtain effective feedback from individual viewers. The creativity of AI can be redefined in future through an adaptive training architecture that can incorporate artist feedback.

## 2  Appreciative Systems: Mimicking Styles

To make a generative system appreciative, it must have the ability to assess the aesthetic value of its output. As shown in **Figure 9**, one of the most significant generative modelling technologies is the Generative Adversarial Network (GAN) **[49]** which is based on the idea of using the information of appreciation to guide the training of generation. In the context of generating artefacts, we may consider a GAN as comprising a forger who produces imitations of artworks and an appraiser who evaluates the authenticity of these works. They are engaged in a competition: the forger wants the fake works to be misclassified by the appraiser as real and the appraiser wants to detect the fake works. As they compete, the forger makes many imitations and learns from what the appraiser allows to go through; meanwhile, the appraiser improves at distinguishing fake from real works by learning from the forger's tricks. The forger and appraiser are both DNNs, called the generator and discriminator, respectively, that come to understand through the interactive training process the nuances of what makes a painting real. Once human appraisers are unable to tell whether the generated paintings were created by the algorithm or by an actual artist, the training is complete.

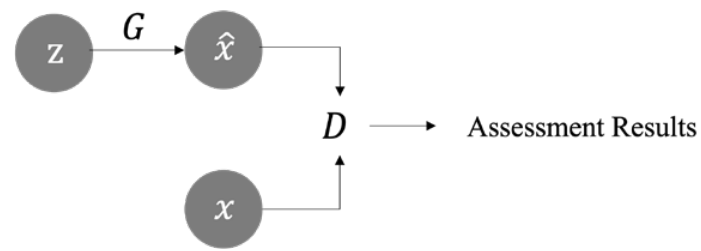

**Figure 9** GAN-based art appreciative systems, where z is the controlling factor, x and ˆx are the human- and machine-generated samples, respectively, G is the generator acting as a forger and D is the discriminator acting as the appraiser

GAN is appreciation-driven in the sense that the generator is built not merely by sampling from a concept space encoded from input data but rather based on the appreciation of the discriminator. Thus, the discriminator provides a mechanism for encoding aesthetic preferences for the generator.

The power of GAN is revealed when deep learning techniques are used to implement its components. For example, CNNs can be adopted to build a GAN. CNNs are composed of layers or filters that can extract higher-level features from an input image and generate differently filtered versions as output. This enables the transformation of images into representations that capture the high-level content, such as the objects present in the image and their arrangement, while disregarding the precise pixel values. For instance, the Convolutional Neural Network (CNN)-based StyleGAN **[50]** has the ability to independently learn high-level characteristics, including the position and identity of human facial features. Moreover, it can manage stochastic variances in the images it generates, like the presence of freckles and hair. This provides an intuitive means of controlling the image generation process, as demonstrated in the examples shown in **Figure 10**.

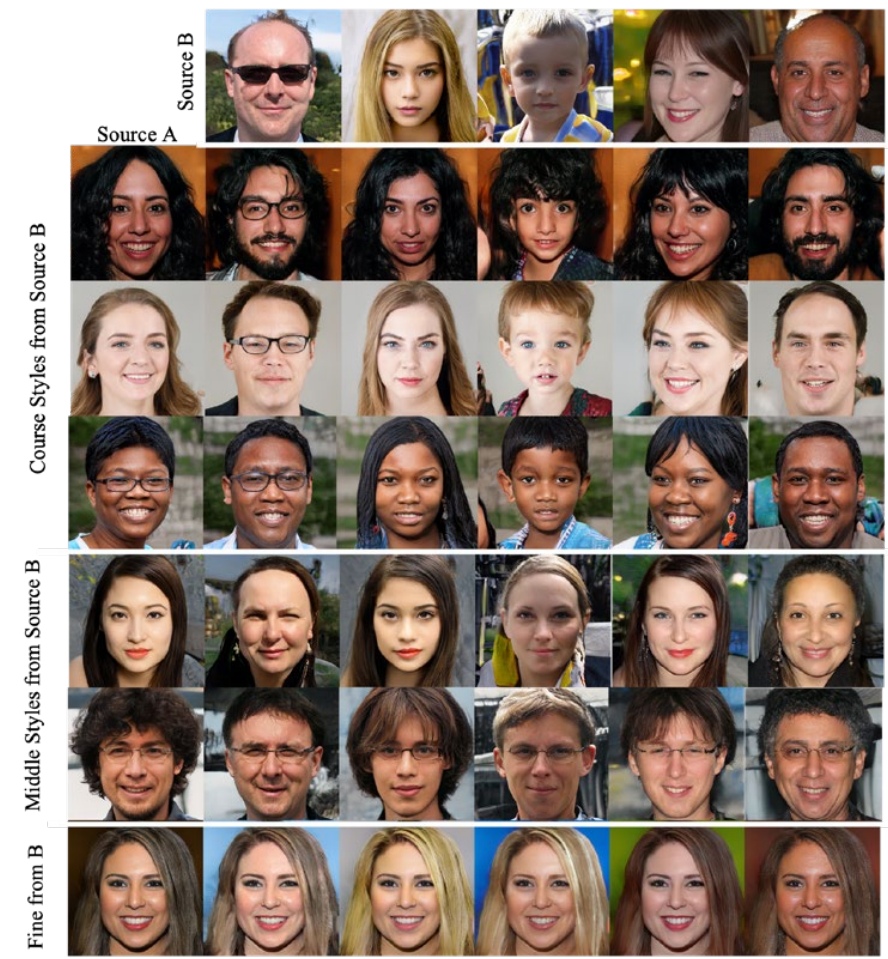

**Figure 10** Examples of controlled attributes in images generated with StyleGAN **[50]**

The use of the adversarial framework to appreciate, assess and regulate generated musical works has also been studied in some depth. For music composition, MuseGAN **[51]** and Dual Multi- branches GAN (DMB-GAN) **[52]** use a discriminator to evaluate whether composed multitrack and multi-instrument music is harmonic. In acoustic modelling for virtual singing voice generation, Multi-SpectroGAN **[53]** is proposed to use an adversarial feedback to regulate the spectrogram quality in the case of multiple singers. Many GAN-based vocoders also offer superior performance, such as MelGAN **[54]**, WaveGAN **[28]** and HiFi-GAN **[55]**.

# 3 Artistic Systems: Mimicking Inspiration

To further develop a generative system that is not only appreciative but also able to invent its own aesthetic appreciation for ranking and selecting generated outputs, an artistic system requires the ability to learn appreciation based on some given principles. Creating an artistic system that possesses an evolving aesthetic appreciation ability is a major milestone in the development of machine artists. One of the most noteworthy strides in this direction is the introduction of the Creative Adversarial Network (CAN) [56], a variant of the GAN that incorporates a revised discriminator. It goes beyond merely checking the quality based on similarity with the original training data to incorporate some rules of appreciation. As depicted in **Figure 11**, the Creative Adversarial Network (CAN) differs from the traditional GAN approach of generating artifacts that closely resemble the original ones. Instead, CAN aims to develop a generator that can produce novel creations by deviating from established style norms, while still adhering to accepted artistic principles. To achieve this, the CAN discriminator not only classifies the generated art as "art or not art," but also provides a signal of how innovative the art is in relation to established styles. Consequently, the generative model strives to maximize acceptance of the generated art, ensuring that it is recognized as art while being difficult to classify into established styles. By learning the characteristics of established styles, the discriminator is equipped to make its own aesthetic judgments. This innovative approach has been successfully applied to paintings from the fifteenth to the twentieth century, resulting in the production of abstract paintings that reveal the trajectory of art history [57].

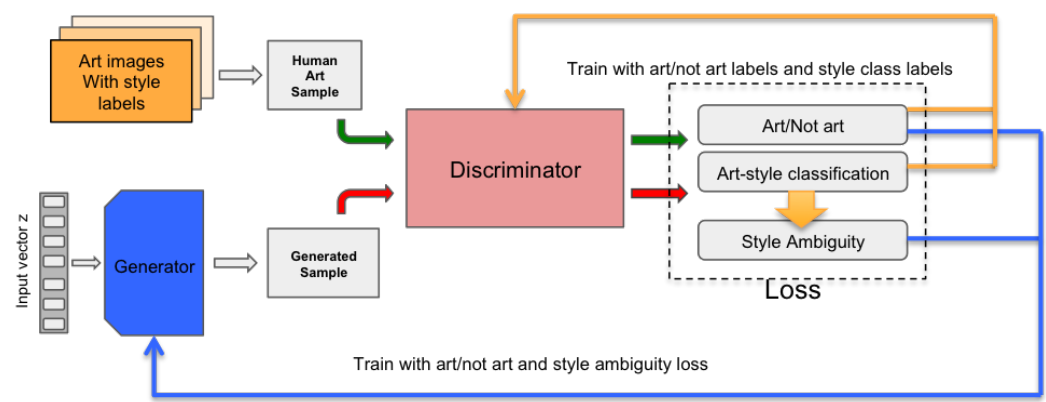

**Figure 11** Block diagram of CAN (from the original paper) [56]

Comparable concepts have been implemented to inspire originality in the created samples for musical composition. The innovation of the Musicality Novelty GAN, as proposed in [58], lies in its use of dual adversarial networks that mutually optimize the musicality and novelty of the music it generates. This model engages in a "novelty game", striving to enlarge the smallest distance between the machine-crafted musical samples and all the human-created samples present in the database.

Unique artworks are generally created by artists making small modifications to an already known style. Instead of scanning existing art pieces and opting for AI recreation, another approach by which machines can produce artwork is based on traditional brushwork. For instance, Hong Kong artist Victor Wong has made the unprecedented move of creating an AI robotic ink artist ('Gemini') that generates its own Chinese traditional ink wash paintings. With algorithmic equations inputted by Wong, Gemini showcases its artistic prowess by crafting its own compositions, skillfully wielding a brush firmly grasped in its robotic arm. With meticulous strokes of varying width and a captivating interplay of shades, Gemini artfully "paints" with water and ink on paper, creating mesmerizing works of art [59]. Whether such impressive technical improvements are to be considered as contributions to machine-generated decorations or to be counted as creative art communicating messages to the human mind can only be decided when the creation is seen in context. Does the receiving human mind experience some degree of artistic elevation and awe, or will the human spectator just see a pretty exquisite rendition? The answer will most likely depend on the cultural context.

There are many more exciting techniques in the area of art generation, such as automatic 3D modelling from a 2D image and creating a photorealistic face from simple sketches.

The works of Turing artists, as named in **Figure 1**, can inspire human artists. As pointed out by Margaret Boden, a cognitive scientist and advisor at the Leverhulme Centre for the Future of Intelligence in Cambridge, "If you have a computer that comes up with random combinations of musical notes, a human being who has sufficient insight and time could well pick up an idea or two. A gifted artist, on the other hand, might hear the same random compilation and come away with a completely novel idea, one that sparks a totally new form of composition." The question remains, however, of whether human works can inspire machines to create meaningful artworks, rather than mere artefacts of '"random combination"'. Even more interesting is the possibility of effectively realising bidirectional communication and mutual inspiration between an art machine and human artist-spectators. Norbert Wiener investigated the interplay between human beings and intelligent machines in his highly influential book *The Human Use of Human Beings* [60], which states that "'society can only be understood through a study of the messages and the communication facilities which belong to it; and that in the future development of these messages and communication facilities, messages between man and machines, between machines and man, and between machine and machine, are destined to play an ever-increasing part'". We believe that the communication between humans and machines is also at the core of the development of a future AI-empowered art community. Furthermore, this communication goes beyond mere information exchange: in the words of Susan Sontag, '"Information will never replace illumination'" [61]. For human and machine symbiotic creativity, inspiration by human aesthetic values is the illumination shared by human and machine artists.

In the rest of this report, we focus on our view of future machine artists developed on the theme of communication between machines and humans for co-creativity. We first discuss the issues involved in building a new form of art database that can serve as the basis for machines to understand human artistic appreciation. We then discuss the development of manifestation technologies that can present machine-generated artworks to humans as the essential mechanisms for machines to communicate. Finally, we propose a new framework for building novel machine learning algorithms to train machine artists.

# *Part 2*
# Art Data and Human–Machine Interaction in Art Creation

Starting from a basic artistic system, we aim not only at enabling machines to understand what makes a creative work 'beautiful' for humans but also at advancing the capacity of AI to generate inspiring artworks. For this purpose, one important task is to build art datasets with well-defined standards of curation that are rich in styles, genres and contextual information. These datasets can then be used for training generative AI systems to create artworks that can be understood and appreciated by humans. An example of an art data repository is WikiArt, a comprehensive visual art archive that, as of 2020, housed over 160,000 artworks from more than 3,000 artists. This collection encompassed works from the fifteenth century up to modern times. The WikiArt dataset has gained widespread popularity as a valuable resource for training models dedicated to the classification of visual art pieces based on 27 distinctive styles and 45 diverse genres. While its main application has been for classification tasks, endeavors have also emerged to leverage the WikiArt dataset for style transfer and generative purposes, as exemplified by the "Improved ArtGAN for Conditional Synthesis of Natural Image and Artwork" [62]. Notably, the groundbreaking achievement of the first AI-generated painting to be sold at auction, *Edmond de Belamy*, was realized by an algorithm trained on a rich collection of 15,000 portraits harvested from the WikiArt dataset.

To expand the art creation capabilities of generative AI systems to the generation of different forms of AI art and possibly even to new forms of multimedia or transmedia artworks, it is desirable to build a more comprehensive art data repository by including music, literature and other artforms alongside visual art. The proposed comprehensive art data repository will contain the data for each piece of art with all the corresponding metadata on the work and the artist/s. Furthermore, artists should be encouraged with rewards to contribute their works to the repository, which would then serve as an additional channel for them to communicate their art with audiences. Blockchain technologies can support the building of a data-sharing network with a provenance- tracking mechanism for human–machine symbiotic art creation. If a generative AI system trained with the works contributed by certain artists were to create an artwork, the provenance-tracking mechanism would be able to identify the contributing artists and thus enable some form of shared ownership.

Such a comprehensive art data repository would have a symbiotic quality. Artistic activities always involve three different roles: creators, performers and receivers. Creators generate the content by conveying their thoughts and imagination. In many types of performing arts, such as music and dance, the expressiveness and infectiousness of the performance come from the performers, who deeply understand the artwork and can subtly manipulate the details of its performance. Receivers are those who appreciate artworks, with their emotions engaged. A symbiotic art dataset would include these three roles in a closed loop.

For machine artists to effectively express humanity, technologies are needed to enable machines to understand what makes a creative work 'beautiful', 'frightening', 'elevating', 'depressing' and so on for human audiences, and to have a closed loop of understanding the response of humans towards a creation process (invention) and evolving this creation process to influence or to attract human appreciation (persuasion). Capturing human responses has become feasible with the development of modern sensing technology. Considering the different roles of artistic participants, as shown in **Figure 12**, the behavioural and biometric responses of creators, performers and receivers during different artistic activities can be captured and contribute to establishing a comprehensive database.

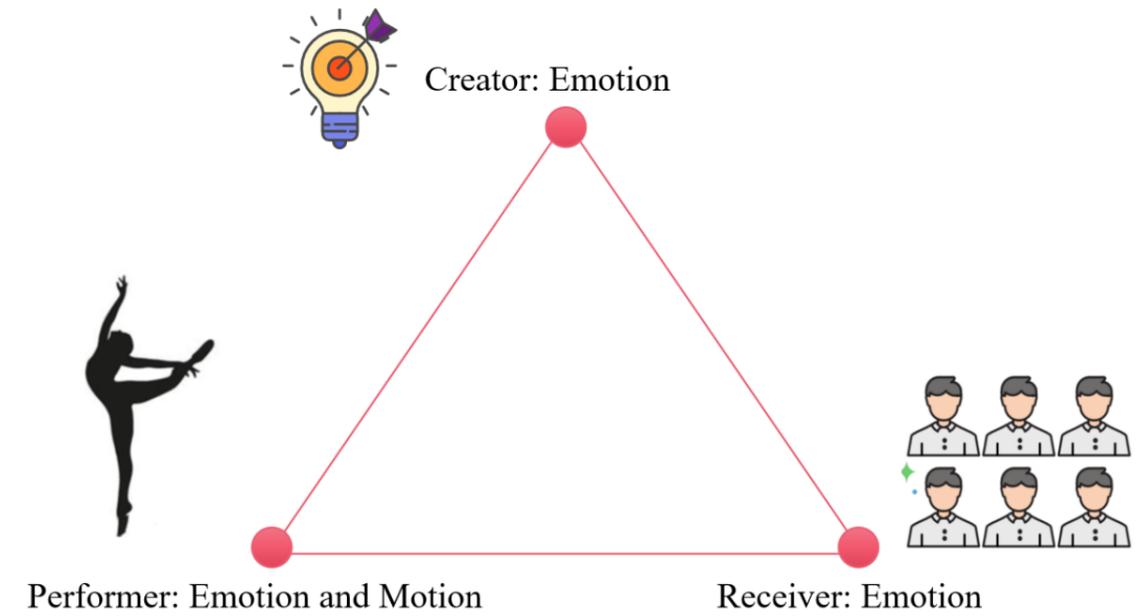

**Figure 12** The data triangle for symbiotic art data collection

The main focus of this symbiotic art data collection will be on recording the motion and emotional responses of the participants, representing the external and internal behaviour of humans, respectively, with the pairwise internal and external behaviours also collected for further investigation of their connections.

## 1 Biometric Signal Sensing Technologies and Emotion Data

The emotional responses of human participants in artistic activities are the core component of the symbiotic art database to be captured using biometric signal sensing technologies. As summarised in **Figure 13**, typical biometric signals related to human emotion that could be included in the art database are brain electroencephalography (EEG) signals, heart rates, radiation from the skin, eye movements, facial expressions and body postures.

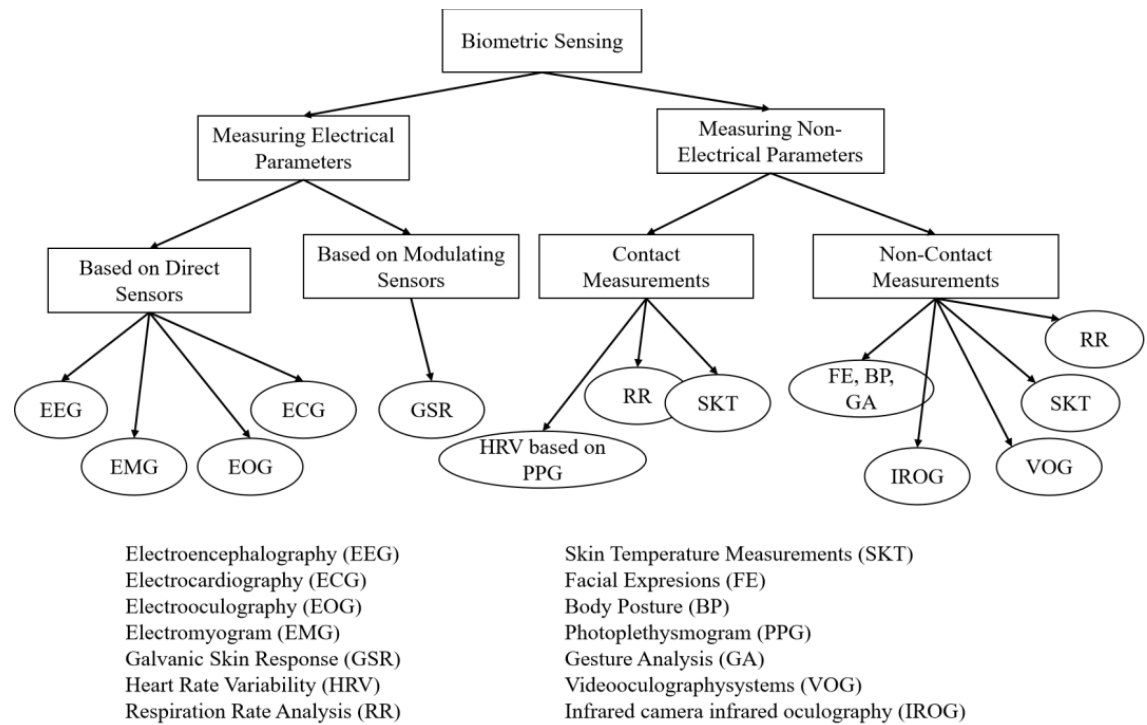

**Figure 13** Typical biometric measurements

Brain–computer interfaces (BCIs) enable electrical activities in the brain to be captured. BCIs can be generally categorised into [63]

(a) non-invasive approaches, including electroencephalography (EEG), magnetoencephalography (MEG) [64], electrooculography (EOG) [65] and magnetic resonance imaging (MRI) [66];

(b) partially invasive approaches, such as electrocorticography (ECoG) [67]; and

(c) invasive approaches, such as using a microelectrode array [68]. EEG has become the most widely adopted approach for brain signal capturing because it is non-invasive and poses low risks [69] to human subjects. Conventional EEG capturing systems conventionally rely on the application of a conductive medium, such as saline or gels, between the electrode and scalp. However, this approach can potentially result in allergic reactions or infections [70]. In response, alternative solutions in the form of semi-dry and dry EEG technologies [71]–[74] have been developed. These innovative technologies eliminate the requirement for conductive mediums, enhancing safety and user comfort. Moreover, the advent of mobile EEG technologies [75], [76] has further expanded possibilities by enabling brain activity capture during art-related processes, such as performance and appreciation, while granting participants the freedom to move unrestrictedly.

The condition of the heart can be tracked using electrocardiography (ECG) [77] and photoplethysmography (PPG) signals [78], from which heart rate variability (HRV) [79] and pulse rate (PR) [80] can be subsequently derived. The heart's state is intrinsically tied to oxygen saturation levels and blood pressure. Aside from these parameters, various other physiological signals may reflect the body's responses via skin emission, such as the galvanic skin response (GSR) [81], electrodermal response (EDR) [82], skin temperature, and electrodermal activity (EDA) [83]. Furthermore, data about eye movements and electrical signals gathered from eye-tracking devices and electrooculography (EOG) can offer valuable insights into the eye's condition. Many of these biometric signals, such as ECG, blood pressure and HRV, can be captured by commercial wearable products, such as smart watches and wristbands [84], [85], which reduces the cost and increases the feasibility of data collection. Besides these physiological signals, behavioural patterns such as facial expressions and body posture can also be captured using cameras and computer vision algorithms to provide a more comprehensive description of the emotional status of human subjects.

The aforementioned signals are particularly informative when collected in a hyper-scanning mode, which enables a comprehensive assessment of the response of an audience to the activities of AI/human performers. Collecting data in this mode allows for interdependencies to be detected among the responses of different audience members and the feedback elements related to the 'dialogue' between performers (AI and human) and the audience.

When gathered, sensor data holds the potential to illustrate emotional, mood, and stress states (EMSR) using well-established emotion theories. Using traditional labels like 'anger,' 'disgust,' 'fear,' 'joy,' 'sadness,' and 'surprise' can be useful in communicating meaning to laypersons. However, many scholars favor using multidimensional models such as valence-arousal [86] or pleasure-arousal-dominance [87] to categorize emotions within a two- or three-dimensional framework. In line with James Russell's circumflex model of emotion [88], as depicted in **Figure 14**, any emotion can be succinctly represented by its specific position within a circular layout defined by arousal and valence dimensions. Emotions can be evaluated through various indicators such as behavioral tendencies, physiological responses, motor expressions, cognitive appraisals, and subjective feelings [89]. Among these indicators, all but subjective feelings can be directly measured using biometric sensors to assess the emotional state of individuals during artistic activities. To simplify the discussion, we first classify the collection of emotional data based on whether the data are gathered in a controlled environment (laboratory) or in naturalistic settings (in the wild). Subsequently, we address some common challenges in the collection of emotional data.

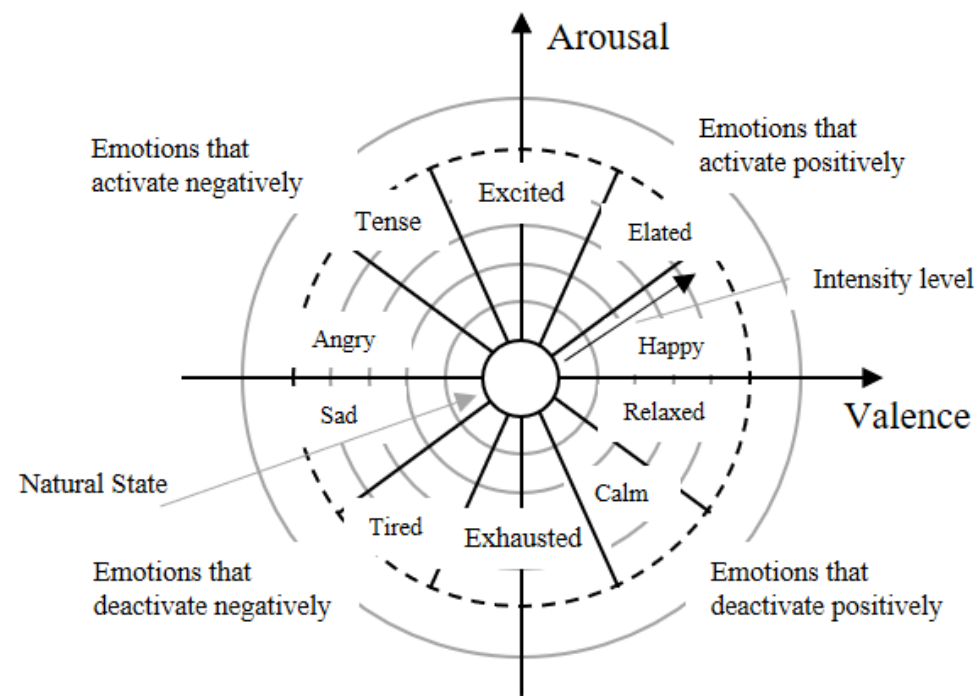

**Figure 14** James Russell's circumflex model of emotion

## 1.1 Emotion Data Collection in the Laboratory

Collecting emotion data in a laboratory setting enables the design of a specific protocol for emotion stimulation and of the standards for capturing and choosing participants. As discussed above, all three groups of people involved in artistic activities – creators, performers and receivers – will participate in the emotion data collection experiments using the following techniques.

First, the emotions and behaviour of the creators of different artworks will be recorded for use in revealing the connections between the thoughts of artists and their outputs. Taking drawing as an example, an experiment can be designed in which painters are asked to create a painting freely. The biometric signals of the painters, which can include the activities of the brain, heart, eye and skin, and the changes in the visual appearance of the artworks will be recorded throughout the creation process. This can be achieved either by using a video camera or by asking the painters to draw directly on a tablet with screen recording enabled. The painters will also be asked to mark any eureka moment of inspiration.

A sequential database for art creation can thereby be established that directly links the creators' emotions with the artistic output. It would then be of great value to establish whether a new stroke or element in a painting can be predicted by the biometric activities recorded up to a specific time, which would make mind drawing possible, or whether the creation of links between conceptually related elements is detectable at the biometric signal level, which could inspire new algorithms for the creation of conceptually meaningful artworks.

The procedures described above can be extended to music composition, in which composers' biometric activities and the musical notes they write or produce would be recorded synchronously. Artists could also be asked to carry out other tasks, such as painting and music adaptation.

Second, emotion data will be collected from performers. The main forms of art performance are stage performance (e.g. dancing, martial arts) and playing musical instruments. The laboratory environment can then be set up for motion capture ('mocap') tasks, which can be used for capturing the performance of a wide range of artworks of different types, cultures, genres and themes. With mobile EEG devices and devices such as Huawei Watch, which support capturing real-time responses from the heart, biometric information can be collected synchronously with motion data.

One problem with capturing the emotional indicators of performers engaging in large and frequent body movements, such as dancers, is that the physiological signals are strongly correlated with the performers' physical activities in addition to their emotional status. For example, vigorous activity increases the heart rate and produces significant changes in EEG signals **[90]**. We therefore propose to collect the emotion data from performers under different expressive tendencies to reveal the effects of emotional behaviour on the EEG signals and calibrate these data with those collected from the same performers at rest. The biometric data collected from musicians are less affected by motion than those from dancers or actors, as musicians' movements are often limited to the fingers and upper body and are relatively small, but it will still be useful to synchronise the MIDI data with a recording of the performing process to exclude the influence of motion.

Third, the emotional responses of observers to different artworks will be captured. Under laboratory conditions, the stimuli presented to observers can be specially designed. Generally, observers will be first asked to face a blank wall and clear their minds for acquiring baseline data without a stimulus. Subsequently, a varied selection of paintings, movies, or music will be introduced, while synchronously gathering the associated physiological signals. Similar to the data collection process involving creators, observers will be requested to indicate their moments of inspiration or breakthroughs **[91]**. Apart from capturing mere observations of the artworks, biometric data will also be acquired during instances of divergent thinking or recall, accompanied by a request for observers to document or depict the outcomes of their divergent thinking process.

The scenario of an artist creating in the presence of spectators will be of special interest. It is well known that performers such as musicians, dancers and actors are actively stimulated by the presence of an audience. The technological approaches sketched above will allow us to address how the dialogue with observers that performers experience when they perform for an audience relates to the creation process. To gain insights into the artistic value of adding AI artists to performances, two situations involving collective dynamics can be compared: human performers in the presence of spectators, and a symbiotic performance of humans with AI.

Conducting the aforementioned data collection experiments necessitates a controlled laboratory environment to ensure accuracy and reliability. Recognizing the influence of the environment on emotional responses **[92]**, meticulous attention will be given to provide comprehensive and precise descriptions of the experimental conditions.

## 1.2 Emotion Data Collection in the Wild

Unlike in laboratory environments, no explicit emotion elicitation protocol is available when collecting emotion data in the wild. Instead, participants undergo art appreciation by visiting museums, sightseeing or watching plays or movies in theatres while wearing portable biometric capturing devices, such as mobile EEG devices, smart watches and wristbands. Biometric signals are collected by these mobile devices throughout the art appreciation activities. This scenario has the significant advantage of eliciting emotions naturally in a real-world environment rather than relying on elaborately designed protocols in a laboratory, which may not stimulate the same emotional responses as natural settings.

However, there are also many challenges to data collection in the wild. First, although uncontrolled settings may be beneficial to collecting natural emotion data, they also create considerable inter-individual differences in the collected data across many control factors, making pairwise comparisons and dataset construction difficult. Second, given that a creator usually works by first perceiving the surrounding environment and the nature of information from the environment is multi-modal, it is difficult to classify the role of the participant within the experiment as solely a creator or receiver and to attribute the captured biometric signals to a specific activity. Third, it is difficult to fully describe the factors giving rise to emotional responses, and these factors are time-variant and difficult to reproduce.

## 1.3 Common Issues

In this section, we discuss three common issues in emotion data collection. The initial challenge lies in accurately assigning labels to the emotions corresponding to the captured biometric signals. One commonly employed approach is to rely on participants' self-reports of their emotions [93]. However, it is crucial to note that while the collected emotional responses derived from sensor outputs are objective, self-reported emotions are highly subjective and may be prone to inaccuracy [94]. Participants usually lack the specific knowledge to accurately define their own emotions in terms of arousal and valence based on Russell's circumplex model and it is difficult for them to recall how they felt at each moment of an artistic activity when the labelling is conducted afterward.

The inaccuracy of self-reporting might be less harmful in laboratory settings than in wild settings because in the laboratory a large group of people can be asked to conduct experiments under the same conditions and statistical corrections can be made. It is also possible to rely on physiologists to identify the underlying emotions based on the captured signals [95], but this is labour-intensive and the labelling accuracy is determined by the expertise of the annotator. From the research perspective, this raises the challenge of alleviating the reliance on accurately labelled sentiment data in developing symbiotic AI by taking emotion as the auxiliary variable and analysing the relationships between the biometric signals and artistic activities.

The second problem is how to prepare the stimuli presented to the observer, such as paintings, movies or music. To secure diverse artworks as stimuli, a comprehensive and extensive database of asymbiotic art must be established as a prerequisite for collecting affective data. There are several databases available that can be used for this purpose. Notable among them are Web Gallery of Art [5], WikiArt [6], Rijksmuseum Challenge [7], Art500k [8] and OmniArt [9], which provide a large collection of paintings. Likewise, for selecting musical stimuli, existing music databases such as AllMusic [96], Discogs [97], and Jamendo [98] provide valuable resources. In order to effectively manage master data and supplementary information, a clear protocol should be designed. In order to consider the subsequent symbiotic art content generation task, the data distribution should be analyzed before using the data as stimuli. This involves annotating genre, emotion, musical structure, and other factors. These annotations are also required when setting learning objectives or control factors for training content generation models.

There remain severe distribution imbalances in the annotated databases for art content generation. Taking music as an example, there are many rich databases for piano but far fewer for traditional Chinese musical instruments. The generated artwork should jointly consider expressiveness, structure and multi-track harmony, but datasets with annotations regarding music emotion or music structure are rare and there is a lack of expressive singing voice datasets containing styles such as high-pitch, trill, glide, folk and Chinese opera. Therefore, it is crucial to consider how to mitigate the imbalance of source data distribution when preparing the artistic data source for emotion data collection.

The third concern revolves around accounting for variations in participants' backgrounds and prior knowledge of the art pieces. Art appreciation responses are influenced by participants' background knowledge [99], which encompasses factors such as familiarity with the artworks, a long-standing interest or research background in the subject matter, or previous education in related fields. For instance, when considering the same painting, a group of professionals well-versed in aesthetics may interpret it differently compared to a group of amateurs, and these disparities can be reflected in the captured physiological signals. Therefore, it is necessary to develop a systematic framework to evaluate the expertise of the subjects.

# 2 Motion Capture Technologies and Motion Data

Mocap is used to record the movements of people or objects for various entertainment, sports, medical and artistic applications. In this report, we are interested in its application to performance capture, with special requirements for recording precise and subtle body movements (including those of the face, fingers and performing tools) in multi-person, interactive scenarios. Here, we review and discuss how state-of-the-art mocap technology enable actors to fully express their artistic intentions with their body and bring a virtual character to life, and facilitate endless possibilities in transmedia digital creations.

## 2.1 Motion Capture Technology

Mocap technology has developed rapidly in recent decades. The advent of real-time performance capture systems has opened up avenues for the development of innovative media formats, wherein a live actor controls an avatar in real time. This technological advancement has paved the way for the creation of immersive experiences that blend live performances with virtual characters seamlessly. Many companies provide various mocap options that offer different levels of accuracy and efficiency at different price points. The main options are the inertial, optical and hybrid systems as shown in **Figure 15**.

**Inertial mocap systems** (e.g. Xsens **[100]**, Perception Neuron **[101]** and Rokoko **[102]**) utilize small sensors known as inertial measurement units (IMUs). These IMUs consist of gyroscopes, magnetometers, and accelerometers, which enable the measurement of forces and rotations at precise body locations. By leveraging this technology, accurate motion data can be captured and utilized for various applications in the field of motion tracking and animation. IMU sensors can transmit data without body occlusion problems. However, IMUs are prone to positional 'drift' over time because they lack knowledge of their actual 3D coordinates.

**Optical mocap systems** (e.g. Vicon **[103]**, PhaseSpace **[104]** and OptiTrack **[105]**) use optical cameras to detect markers placed on an actor's body and calculate the 3D position of each marker in real time. To ensure accurate triangulation of a marker's 3D position, it is necessary to employ multiple cameras. Each marker should be visible to at least two cameras simultaneously, as this provides the minimum information required for triangulation. The presence of more cameras that can capture the marker enhances the precision of calculating its 3D position. In other words, the more cameras that have visibility of the marker, the greater the accuracy achieved in determining its spatial coordinates. The markers can be active or passive. Active markers emit light that the cameras can detect, with additional information such as the marker's ID encoded into the light pulse. Passive markers retro-reflect light from other sources, usually infrared light projected from the camera. Active markers are more robust under natural lighting because there is a lack of sufficient contrast from the ambient environment for passive markers to be effective. Optical systems, despite their advantages in producing high-quality results and accurate positional data, have limitations due to the need for many cameras. This requirement makes optical systems expensive and less portable compared to other alternatives. However, these systems excel in scenarios that involve multiple actors interacting in the same physical space, providing comprehensive support for capturing their movements with exceptional precision and fidelity.

**Hybrid optical/inertial systems**, which combine the benefits of each of the other two systems, are becoming available. When occlusion is severe, optical markers cannot resolve the ambiguity, whereas inertial systems continue to provide data. Inertial data can also help to reduce jitter noise (caused by measurement uncertainties and factors such as ambient light) when jointly optimised with mocap data from an optical system. This solution reduces the required number of cameras, which lowers the overall cost. Another alternative is to use an inertial suit to provide data while using one or more optical markers to provide extra positional data to alleviate the positional drift. However, fusing different data sources in this way to reconstruct natural motion remains very difficult.

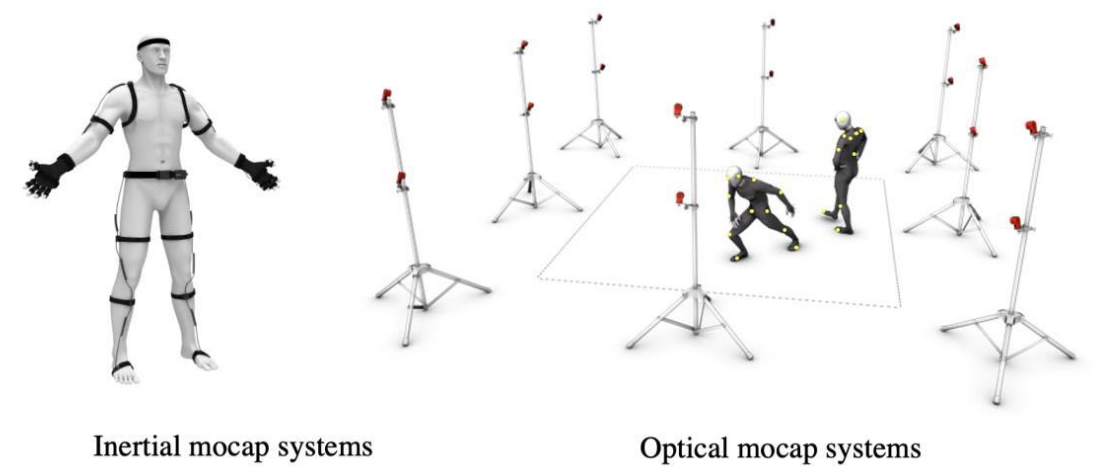

**Figure 15** Mocap systems based on inertial **[101]** and optical sensors **[105]**

## 2.2 Affordable Motion Capture and Research Directions of Motion Reconstruction Algorithms

Motion capture using RGB/RGBD cameras offers a more affordable alternative but presents technical challenges. In an image-based motion capture system, 2D images captured by one or two cameras are utilized to triangulate the 3D position of an object over time. One notable example of a successful commercial product in this domain is the Kinect camera [106], which provides a cost-effective motion capture solution with acceptable accuracy for non-professional applications. Although image-based systems may not provide a fully precise estimation of body position, they can still be valuable when a general understanding of body movements is sufficient, such as in action recognition and interactive gesture controls. However, for professional purposes, image-based solutions are primarily reliable for capturing facial expressions. By equipping an actor with a helmet-mounted camera setup, the cameras move in tandem with the head, constantly focusing on the face. Within this focused and limited capture area, the system can record facial movements with a level of detail that enables accurate real-time transfer to a digital character.

Drawing upon established models of the human body, it is feasible to estimate poses with a solitary camera. For instance, parametric human models such as SMPL [107] can be meticulously fitted via optimization methodologies employing 2D observations, including key points and silhouettes. To curtail unnatural shapes and poses, Song et al. [108] elegantly employ optimization techniques to minimize the 2D joint reprojection error. In a similar vein, Zhang et al. [109] propose an ingenious pyramidal network structure to systematically explore multiscale features, thereby explicitly rectifying SMPL parameters. While the integration of additional cameras promises heightened precision, Zhang et al. [110] ingeniously introduce a real-time estimation framework that seamlessly combines per-view parsing, cross-view matching, and temporal tracking. They adeptly address the correspondence problem by imposing an epipolar constraint while judiciously disregarding appearance cues.

Furthermore, Dong et al. [111] propose a matching algorithm that judiciously considers appearance similarity and elegantly formulate a convex optimization problem that robustly reconstructs 3D joints.

One challenge that remains for these algorithms is a lack of understanding of physics (such as gravity, collision, friction, etc.). Therefore, a possible research direction is to combine statistical body priors with deep learning models. For example, Xie et al. [112] argue that by adding laws of physics to the deep learning model, the 'foot sliding' problem in monocular approaches can be solved and more natural poses generated. A higher-quality motion reconstruction can also be achieved by modelling the distribution of motion patterns. The interaction and cooperation between different performers and the cross-modality correlations (e.g. the correlation between music and dance) can be fully explored by attention modules such as transformers to predict novel motions [113], [114] and to recover noisy and incomplete motion detections. These research directions have great potential to resolve some of the major uncertainties surrounding mocap.

## 2.3 Motion Capture Dataset

There are several public mocap datasets, including the SFU Motion Capture Database [115], ZJU- MoCap Dataset [116] and CMU Graphics Lab Motion Capture Database [117]. However, few of these databases have sufficient coverage of artistic performances for the present purposes, and the connections between performances and emotional behaviour have not been explored. As shown in Figure 16, for the creation of a symbiotic art dataset, four types of motion data can be collected using mocap, encompassing human–human and human–environment interactions from different perspectives.

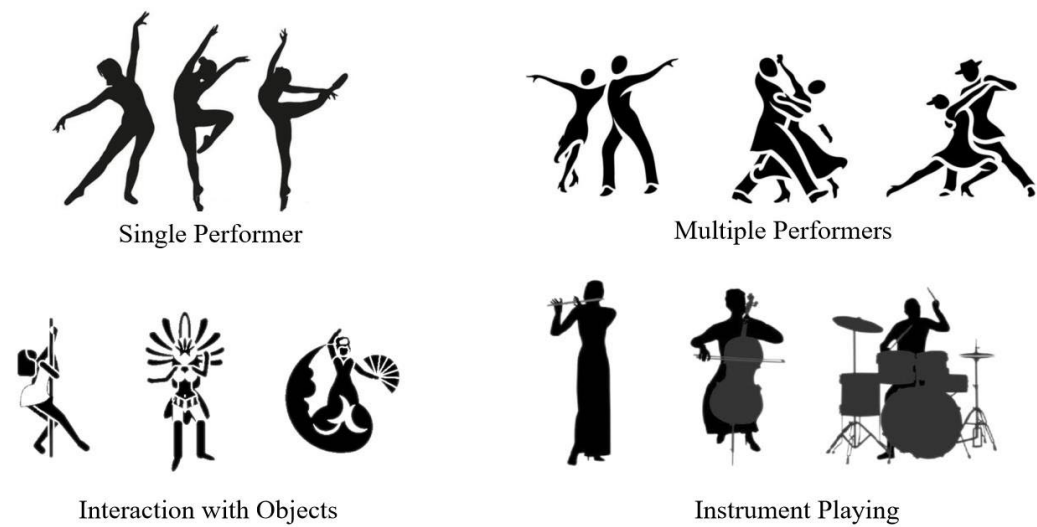

**Figure 16** Illustration of different types of captured motion data.

The first type of data will be focused on the body-scale movements of a single performer in an empty space. A wide range of performances will be collected, varying in type (e.g., dancing, martial arts), culture (e.g., Eastern, Western, African), genre (e.g., ballet, contemporary dance, tap dance) and theme, which is closely related to the emotions to be expressed by the performers. As the dataset should also consider variations in performers' body shapes and gender, different actors can be asked to deliver the same performance. Human–human interactions can then be captured by having multiple dancers perform together. Building an action database with the interactions of real performers will help in producing augmented reality (AR) environments in which virtual humans can dance in response to or in interaction with a human performer, and in having multiple virtual performers perform collaboratively in fully virtual environments by following the rules indicated by the captured human–human interactions. Interactions between humans and objects, including stage props, clothing and animals, should also be collated in the dataset. By placing markers on the objects, their physical movements during the performances will also be recorded. Another essential task will be to capture the movements of fingers, hands and arms in musical performances to establish a mocap database for the playing of different instruments. A wide variety of musical works will be selected for performance, with the MIDI scores of the played works recorded and synchronised with the motion data.

As described in more detail below, the motion data will be aligned with the emotion data to describe the behaviour of the performers. Another crucial part of the mocap database will be the complementary information. A protocol will be needed to systematically handle metadata, including information on the environment, the artwork, instruments and performers. Given that the fluency and expressiveness of a performance are related to the skill of the performer, a questionnaire will also be needed to capture the performers' levels of expertise.

## 3 Photogrammetry/Volumetric Capture

Photogrammetry is a powerful technique utilized in the domains of surveying and mapping, enabling precise measurements between objects through the application of photography [118]. Its origins can be traced back to the mid-nineteenth century when researchers made a remarkable discovery—by employing a minimum of two images, they could determine lines of sight from individual camera points to objects captured in photographs, thereby extrapolating valuable 3D data. A closely associated concept, volumetric capture, finds greater prominence in scenarios where dynamic targets are translated into holographic videos, facilitating immersive and realistic visual experiences.

**Figure 17** illustrates that the prevailing approach in 3D movie and video game production entails the seamless integration of multi-camera photogrammetry and motion capture (mocap) techniques. The 3D scenes and objects are digitally captured by photogrammetry and human targets are rigged for animation. Rigging involves creating a skeleton within a mesh for animators to manipulate characters more efficiently and accurately. The characters' movements are driven by mocap.

Different types of technologies and system setups are available for photogrammetry. Depending on budget and space limitations, certain system capabilities can be achieved through point-and- shoot or multi-camera configurations.

In the point-and-shoot setup, a scene is captured using a limited number of cameras that are strategically relocated to different positions, ensuring comprehensive coverage of the entire area to facilitate the acquisition of a complete mesh. A popular method employed in this context is video-to-photogrammetry, commonly executed using mobile phones or GoPro cameras. This approach offers notable advantages, as it allows for mesh capture in various locations and is cost-effective due to the availability of affordable equipment. However, it is important to note that the quality of the resulting mesh is currently considered suboptimal with the existing technology and software tools. Ongoing advancements are being made to improve the fidelity and precision of the generated meshes.

A more advanced approach in photogrammetry is multi-camera photogrammetry, offering the advantage of capturing objects in motion or animation more efficiently. With a static setup of well-calibrated cameras, this method yields meshes of notably higher quality compared to point-and-shoot systems. However, multi-camera photogrammetry entails higher costs, increased complexity, and reduced portability. Typically, a minimum of 40 calibrated and synchronized cameras is necessary to achieve satisfactory results. Consequently, these systems are predominantly utilized in professional studios where the infrastructure and resources are available to support such setups.

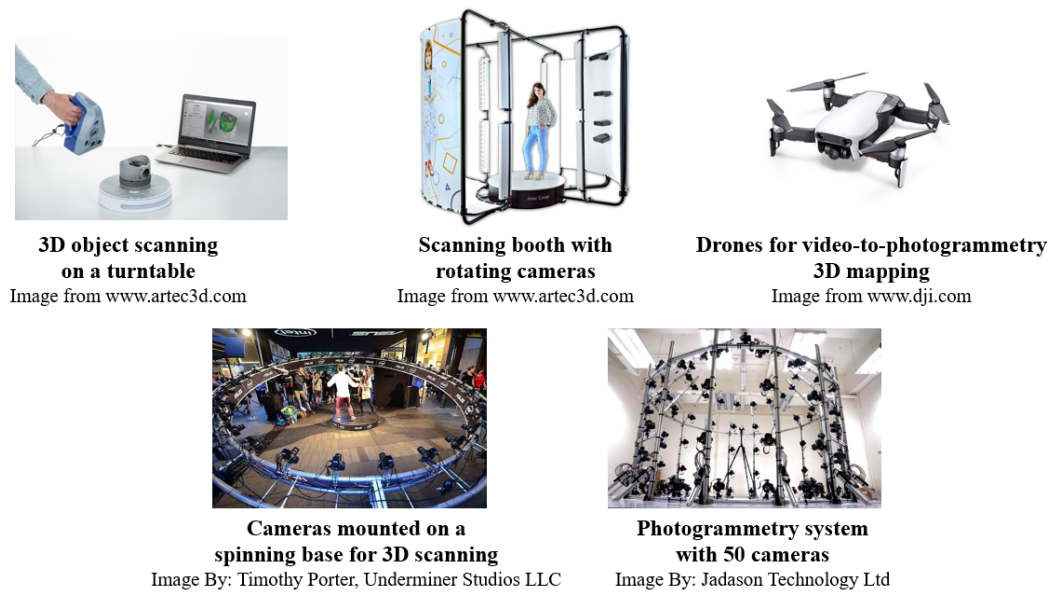

Figure 17 Various system setups for volumetric capture.

A well-known example of a volumetric pipeline is the one developed by Microsoft [119], illustrated in Figure 19. The system employs a green screen stage setup along with over 100 carefully calibrated, high-speed RGB and IR (infrared) cameras. The pipeline begins by processing stereo pairs and silhouette data extracted from the images to generate dense depth maps. These maps are then merged using a sophisticated multi-modal multi-view stereo algorithm, resulting in a comprehensive point cloud. Refinement techniques are applied locally to align the point cloud with the underlying surfaces, ensuring enhanced accuracy. From this refined point cloud, a watertight mesh is formed, which undergoes denoising and optimization for optimal quality. Keyframes are intelligently selected to enable seamless tracking of the mesh across frames, ensuring the production of visually cohesive subsequences.

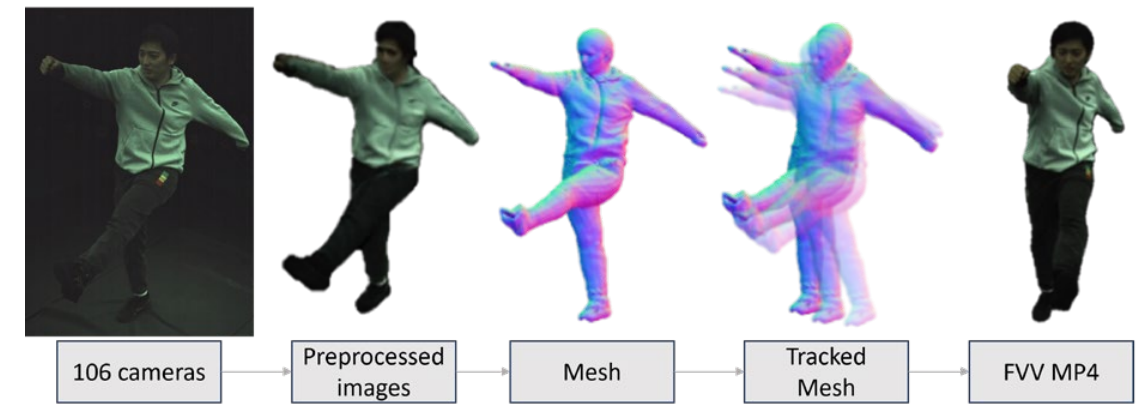

Figure 18 Photogrammetry Processing Pipeline by Microsoft.

Another well-known volumetric capture stage is Intel Studios [120], shown in **Figure 19**. The largest of its kind in the world, Intel Studios has a stage area of 10,000 square feet and can capture up to 30 people simultaneously. Its 100 8K resolution cameras can generate 270 GB/sec of raw RGB data travelling through 5 miles of fibre optic cables.

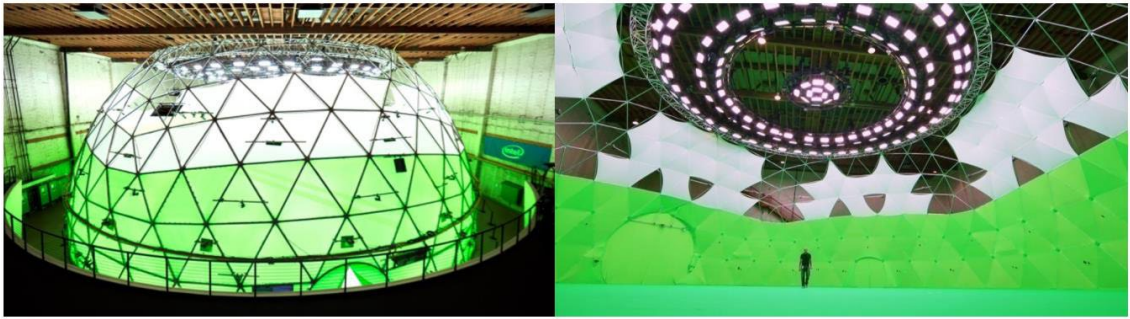

**Figure 19** The Intel Studios volumetric capture stage **[120]**

Photogrammetry technology is developing rapidly and it is likely that the current state-of-the-art technologies will be surpassed every few years. The direction is quite uncertain, but the creation and capture of light beams is the most fundamental issue. Some possible modalities include Wi- Fi, radio and the visual light spectrum.

## 3.1 Affordable Volumetric Performance Capture

Compared with expensive professional studio setups, recent developments in low-cost depth cameras, such as the Intel RealSense **[121]** Depth Camera and Microsoft Kinect Azure **[106]**, and the rapid progress in advanced AI based modelling and reconstruction algorithms show great potential in generating high-quality meshes from affordable devices. These devices have capabilities that were once restricted to very expensive high-end scanners.

In most scenarios, the target of volumetric capture is a human performer. Therefore, human- centric representations and rendering algorithms for human performance capture is a popular research topic. DynamicFusion **[122]** can capture a dynamic object using only one depth camera. Multiple point clouds can be recorded and fused into a canonical point cloud. As illustrated in Figure 20, DoubleFusion further extends DynamicFusion with a two-layer representation. This includes a layer of node graph on SMPL **[107]**, as well as a layer of far-body nodes. These additional layers enable DoubleFusion to efficiently process fast motions. Recently, Neural Body **[116]** was introduced to integrate a parametric human model SMPL **[107]** and a canonical neural radiance field (NeRF) **[124]** for performance capture. A realistic image is artfully crafted by skillfully warping a sparse set of vertices into different poses, followed by the application of an advanced deep learning technique. NDF **[125]** projects spatial rendering points to a learned deformable space surrounding SMPL for more efficient dynamic surface modelling.

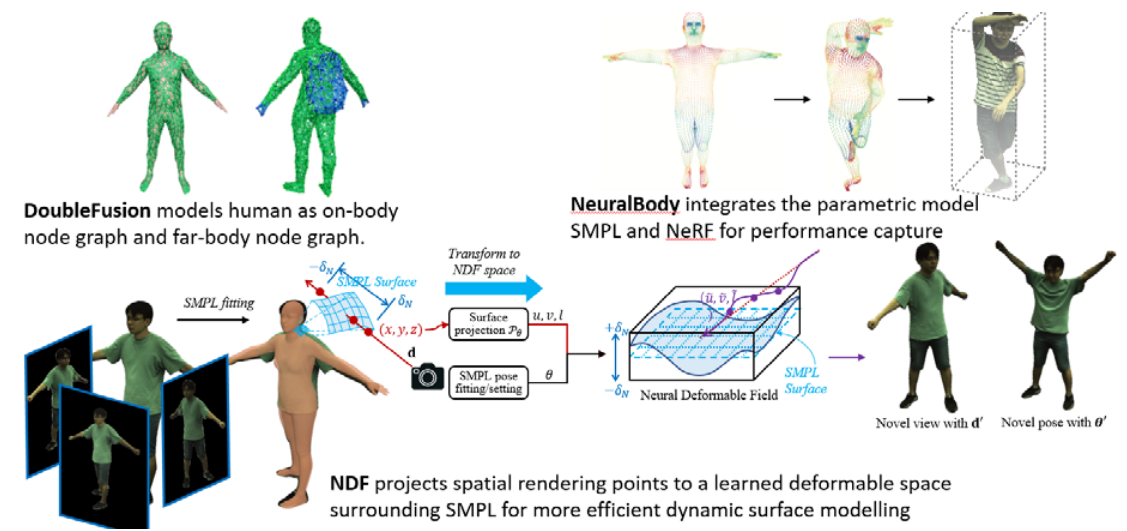

**Figure 20** Developments in novel representations for human performance modelling. BodyFusion **[123]** leverage point clouds and fusion techniques, NeuralBody **[116]** and NDF **[125]** are based on the neural representation model NeRF and parametric

## 3.2 Neural Modelling for Volumetric/Holographic capture

Conventional methods work over pixel intensity space for correspondence matching to generate a sparse point cloud. The more recent deep learning models exploit deep feature space, which is proven to be much more robust. Transformers can also be used to deduce the density distribution along the ray **[126]**, which provides better prediction of occlusion relationships.

Early endeavors in novel view synthesis primarily involved various techniques such as warping, resampling, blending, and exploiting ray-space proximity **[127]**, **[128]**, proxy geometry **[129]**, **[130]**, optical flow **[131]**–**[133]**, and soft blending **[134]**, **[135]**. These methods aimed to transform and combine source views to match target viewpoints, enabling the generation of new perspectives. These frameworks allow for high- resolution rendering and work well with a dense multi-camera photogrammetry system. However, they are unsuccessful in challenging scenarios, such as those with few cameras and in uncontrolled capture conditions.

A promising and recent research direction in overcoming these challenges involves leveraging DNNs to represent the shape and appearance of real scenes. As depicted in **Figure 21**, the weights of multi-layer perceptrons excel at mapping continuous spatial coordinates to various attributes such as signed distance values **[136]–[138]**, occupancy values **[139]**, density **[124]**, **[140]**, and colors. These neural models are trained using differentiable rendering techniques. These techniques necessitate the transformation of 3D objects into images, which are then compared with input images to provide effective supervision. Surface-based representations **[136]–[138]** exhibit gradients that are localized to a single spatial point, limiting the extent of back-propagation. In contrast,

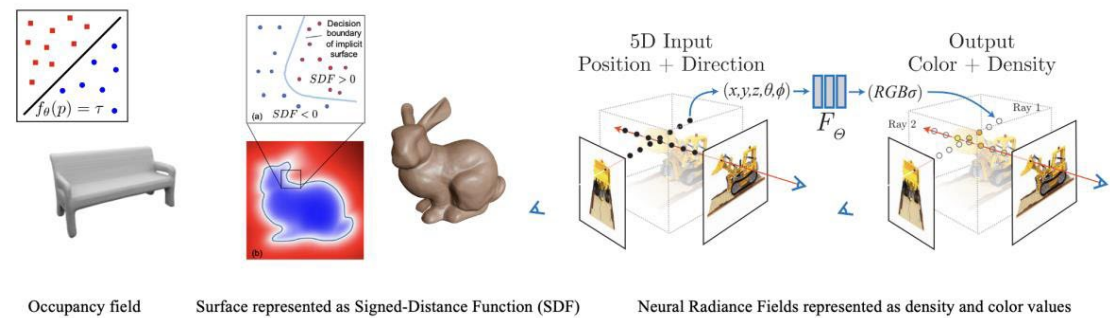

**Figure 21** Neural representations for 3D contents modelling in the forms of occupancy fields **[139]**, signed distance function **[136]–[138]**, and neural radiance fields **[124]**

volume-based representations **[124]**, **[140]** sample multiple points along each ray, generating gradient signals that facilitate the modeling of intricate structures through back-propagation. Nonetheless, extracting high-quality surfaces from implicit volumetric representations remains a challenge, necessitating further exploration of novel representations capable of efficiently capturing volumetric distribution while adhering to surface constraints.

Despite achieving an unprecedented level of photorealism in the generated images, neural representations face challenges when it comes to modelling dynamic scenes. To model the intricate deformations of non-rigidly deforming scenes, researchers typically employ a canonical NeRF (Neural Radiance Fields) model as a foundational template. Complementing this, they introduce a deformation field **[141]** for each observation, facilitating the mapping of observation-space points to the canonical space. This approach enables the acquisition of dynamic scene understanding directly from images, as it effectively captures the evolving nature of non-rigid deformations. By integrating the canonical NeRF model with deformation fields, researchers can elegantly analyze and comprehend the complexities inherent in non-rigidly deforming scenes based on image data.

Despite the aforementioned challenges, we believe that with the ongoing development of AI- based modelling and reconstruction algorithms, the reliance of photogrammetry and volumetric capture systems on expensive hardware setups will continue to decrease over time.

# 4 Aesthetic Descriptor: Labelling Artefacts with Emotion

To endow AI with art creativity, we need to understand aesthetic experience, or how human audiences enjoy artworks. Ideally, feeling something close to what the creator feels or imagines allows an audience to precisely comprehend the inspiration and creativity represented by an artwork. This concept bears a close relationship to the psychological notion of 'empathy,' as originally defined by Robert Vischer. Empathy refers to the inherent ability to comprehend or share in the experiences of another individual, perceiving their emotions and perspectives from within their own frame of reference. It entails the capacity to place oneself in the position of another person, fostering a deeper understanding and connection **[19]**.

To build empathy, artists usually leverage synaesthesia to bridge the gap between them and their audience. Synaesthesia is a perceptual phenomenon characterized by the involuntary experience in one sensory or cognitive pathway triggered by stimulation in another pathway **[23]**. For example, to help students better understand an abstract piece of music, a professor will often inspire them to imagine the close relationship between a passage of melody and rhythm and some visual object or scene, such as a torrential river or a quiet stream. Wassily Kandinsky, who is perhaps the best-known synaesthete **[142]**, defined this intrinsic relationship as 'inner beauty', which is the fervour of spirit and inner necessity of spiritual desire. Inner beauty is the hidden but central aspect of art, which can be projected into various artistic forms such as music, painting and dancing.

Synaesthesia can be explored as the means for communicating human aesthetic experiences via machines. If we can record the human sensory or cognitive information stimulated by an artwork, this record can be used to annotate an artwork with an 'aesthetic descriptor' reflecting the human aesthetic experience associated with it. Although artistic and aesthetic tastes may differ among individuals, individual aesthetic appreciation can be derived from well curated art data of a population enriched with corresponding aesthetic descriptors.

By exploring synaesthesia as the aesthetic values of artworks across various sensations, aesthetic descriptors can be used as hidden representations for information across multiple modalities. Moreover, connecting artworks with different modalities through their common aesthetic descriptors enables cross-media or transmedia creation. In transmedia art generation, an AI system receives pieces of art in one medium as an input and outputs an artwork in another medium; for example, an AI system might take some music as an input, analyse the associated emotions, and then use the results of this analysis to generate a piece of visual art or even a poem. A new generation of art creation is thus made possible with a cross-media art data repository explored by generative AI algorithms.

Advancements have already been made in the domain of synaesthesia-based cross-media and transmedia creations, including the inspiring work of Verma et al. on 'Translating Paintings into Music Using Neural Networks' **[143]**. The authors propose a system designed to learn from the artistic pairings of music as well as the associated album cover art. The ultimate objective of this system is to "translate" paintings into music and, conversely, music into paintings. By studying the relationships and connections between these art forms, the system seeks to establish a means of creative expression that enables the transformation of visual artworks into musical compositions and vice versa. To clarify, **Figure 22** presents an illustrative diagram based on our understanding of **[28]**.

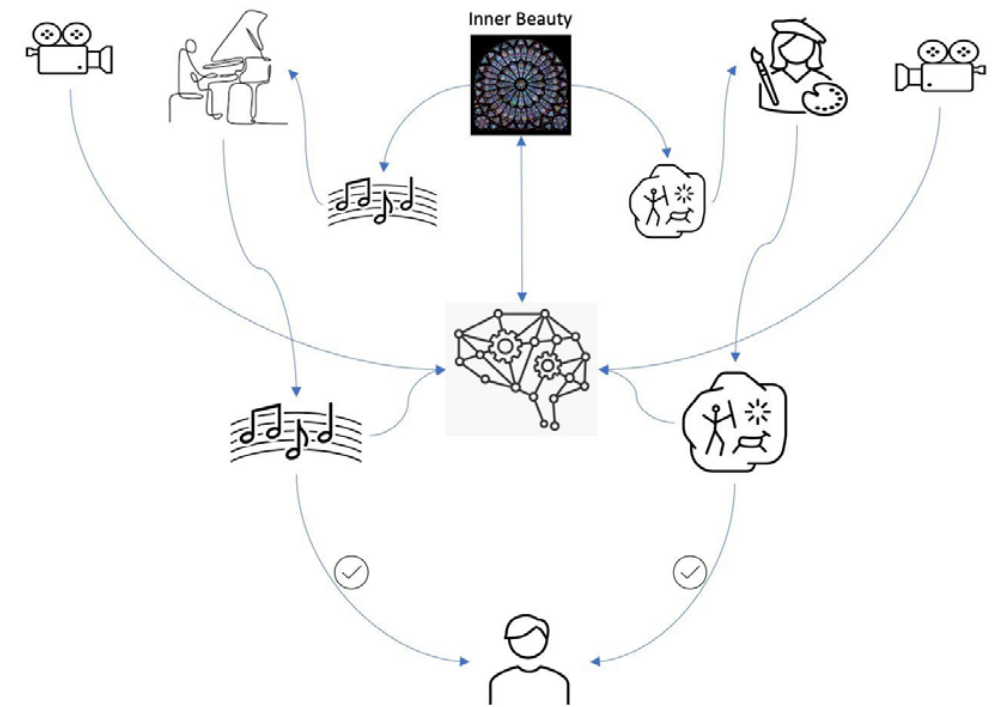

**Figure 22** Inspiration Between a Musician and a Painter via Synaesthesia

Art can be understood as the transfer of emotions and ideas from one human to another through the visual arts, music and various other forms. Although the process of human creativity remains mysterious, it is apparent that the creative process is closely related to brain activities. Furthermore, aesthetic appreciation is certainly a type of cognitive behaviour. Thus, measuring the cognitive and physiological activities connected to human experience and behaviour in art creation and appreciation is an obvious approach for building aesthetic descriptors.

This research demonstrates a close connection to EMSR (Emotion and Affect Sensing and Recognition), a field that has been extensively explored within controlled laboratory settings for several decades. EMSR entails the use of physiological signals to categorize emotional states in controlled situations. However, due to the remarkable advancements in sensor technology, particularly in the realm of wearable sensors, it is becoming increasingly feasible to deploy EMSR systems for a broader population during their daily routines. As a result, there is an increasing demand to assemble datasets in real-world environments to guarantee the dependability, applicability, and practicality of EMSR models for real-life implementations. These datasets collected "in the wild" play a crucial role in validating and leveraging EMSR models to enhance their effectiveness in real-world scenarios.

## 5 Immersive Visualisation: Machine to Human Manifestations

Since its invention in the 1890s, cinema has been the paradigm for using moving images and sound to provide audiences with an immersive engagement with narrative content. The history of cinema has involved a continuous development of screening technologies, from Cinemascope to Cyclorama to Omnimax. The situation changed dramatically in the late twentieth century as the development of computer technology allowed screened content to be fully digitised and algorithmically coded, generating new possibilities for the creation of enveloping and lifelike 3D worlds. Cinema has proven multifarious in its applications, as an artform, as entertainment, in documentary and as a medium of education. Multimedia will continue to expand across areas of public engagement and offer a new range of content, including virtual museums, intangible and tangible cultural heritage, big data archives and multimedia performance. This signifies a notable technological shift towards a more widespread and advanced utilization of the broader multisensory spectrum. The continuous and unwavering quest for increasingly immersive and convincingly perceptual media technology has spurred innovative developments. These advancements cater to the inherent human desire for richer, more engaging experiences that deeply resonate with our senses. As a result, we witness a growing integration of technologies that aim to captivate and envelop us in multisensory realms, pushing the boundaries of what is perceptually possible. In multimedia applications, this immersion is not restricted to the visual realm but rather extended to all the human senses.

A quick sketch of the history of immersive visualisation systems could go back as far as prehistoric cave paintings and murals painted around AD 20 in Pompei. The invention of single point perspective in the Renaissance was another milestone, with the 360-degree painted panorama patented by Robert Barker in 1787 becoming the apotheosis of what could be achieved with such spatial rendering mathematics.

A further essential aspect of the immersive panoptic experience is the perception of 3D depth by means of stereoscopy, which creates a 3D illusion from multiple 2D images. The stereoscope, a popular consumer product in the late nineteenth century, is the photographic antecedent of today's head-mounted display (HMD), and together with 360-degree panoramic painting is the precursor of virtual reality.

Immersive visualisation comes in many projection forms today. One example is the hemispheric planetarium theatre, which is a panoramic 360-degree cinema that evolved from Cinemascope and Disney's Circle-Vision. A significant breakthrough in merging immersive imaging, 3D stereoscopic perception, and positional tracking of the viewer's perspective was achieved with the invention of the multi-sided projection CAVE (Cave Automatic Virtual Environment) at the Chicago Electronic Visualization Laboratory of the University of Illinois in 1992. This pioneering system laid the foundation for the development of fully projection-mapped indoor and outdoor architectures, which have now become the prevailing standard for situated immersive experiences. The CAVE technology inspired the creation of environments where projections are meticulously mapped onto various surfaces, enabling a seamless integration of virtual content with physical spaces. These advancements have revolutionized the way we engage with immersive experiences, providing a heightened sense of presence and interactivity. Other configurations include the planetarium- type dome projection systems, which can also be experienced as upright hemispheres (iDomes) and can render almost any surface shape with the geometry correction capabilities of the latest projectors. Parallel with these outwardly projected strategies has been the development of optical systems that have their antecedents in Renaissance experiments with anamorphosis. Originating in the 1960s in the context of military flight simulation, HMDs were initially complex and expensive but have now become consumer products. In this short time, HMDs such as Oculus and Vive have made the experience of virtual reality ubiquitous, and especially so in conjunction with the similar evolution of the requisite tracking and sensing technologies, such as those built into smart phones.

Another essential dimension to the topic of immersive visualisation is the relationship between representation and reality. The history of art is in essence an exploration of that relationship and a series of attempts to bridge the gap. Exemplary in this respect are the trompe-l'œil and illusionistic techniques of Baroque churches, which blurred the borders between painting, sculpture, architecture, the space of the observer, and even the heavens. In today's

50    51

terminology, this relationship is the subject and ambition of AR and mixed reality (MR) technologies. The range of visualisation, tracking and sensing techniques that are now available can be combined with optical strategies to create perfect alignments and conjunctions between the real (physically present) world and any form of represented (immaterially imagined) world, such that they co-exist as a holistic narrative experience. We are only now beginning to explore the creative potential of these extraordinary capabilities.

Extended reality (XR) technologies can also be combined with biometric signal sensing techniques for building the proposed art database. Given that XR technologies are used in art creation, capturing the human responses within an immersive environment will help to evaluate the interaction between humans and virtual spaces and objects. Meanwhile, XR can facilitate constructing databases with reduced complexity compared with signal capturing in realistic environments because the contents and environments presented to the participants can be more precisely controlled for the collection of synchronised biometric signals.

The next generation of interactive and immersive media platforms will create a fully interactive immersive experience in which the visual and sonic array is calibrated moment by moment at the will of the spectators/interactors, as manifested through their movements, facial expressions, actions and verbal communication. Human interaction combined with real-time immersive visualisation will create MR environments in which cinematic narrative (and its visual expressions) can evolve in real time through sensorial and intelligent engagement with the audience. At the heart of this development are the co-evolutionary narrative potentials of a hybrid multi-modal space of interaction between human and machine agents, within which human intentionality can enter into a dialogue with empathetic AI. This innovative undertaking to bring digital media into the twenty-first century will be afforded by fully immersive display technologies and the prodigious real-time imaging capabilities of digital cinema, amalgamated with versatile sensing systems and the algorithmic extension of human–machine interactions. "Mimetic engineering" refers to the combination of two interconnected elements: the desire for increasingly immersive and realistic viewer/interactor experiences, and the utilization of emerging engineering techniques for human-machine interactions. This term captures the fusion of these aspects, aiming to create technologies that replicate and simulate the real world in a highly authentic manner. The focus is on engineering systems that blur the line between the virtual and physical domains, enabling users to engage in experiences that closely resemble real-life interactions. Mimetic engineering represents the integration of the pursuit of immersion with innovative engineering approaches for seamless human-machine interactions.

# Part 3
# Towards a Machine Artist Model

Artificial intelligence (AI) capabilities have evolved from passive recognition (perception) to active reasoning and decision-making (machine behaviour and cognition), and creative intelligence (machine creation) abilities have recently emerged. Creative intelligence is strongly correlated with reasoning, and Alexander Gottlieb Baumgarten described this relationship when introducing the core concept of aesthetic taste as 'Aesthetic taste is not only sensory but also intellectual' **[144]**.

Artistic creativity has so far been unique to human beings. Endowing machines with the capability of art creation is an academic dream that has been actively pursued by AI scientists, cognitive scientists, artists, psychologists and neuroscientists. Research on machine creativity is expected to advance the exploration of not only machine intelligence but also the cognitive psychological mechanisms of human beings. To facilitate effective human–machine communication based on appropriate ethical standards and moral values, machines must be able to fully understand human emotions and artistic values.

The art creation ability of machines may be different from that of humans. However, from the design perspective, the artwork created by machines must satisfy human aesthetic values. Moreover, the enhancement of machine creativity can promote the evolution of human aesthetic values and abilities.

The following questions are fundamental to the development of future machine artists: How does one define machine creativity? What is the inherent logic of a machine artist model? What is the relationship between the machine artist model and existing generative-based Turing artist models?

## 1 Challenges in Endowing Machines with Creative Abilities

**Sample scarcity.** A large number of data samples are required to train traditional AI models. The acquisition of high-quality samples is challenging in many application scenarios. In particular, in the art domain, artists cannot realistically be asked to label their own artefacts. If the artwork is considered to be a mapping from contents to emotions, labelling the content with its invoked emotion may prove to be challenging. Moreover, the labelling performed by an amateur artist may lead to improper training of the art generation model.

**Evaluation.** The evaluation of the artistic creativity of a machine, similar to that of the aesthetic taste of human-made artwork, is complex and subjective. If the artistic value of the creation is not considered, the system cannot be optimised to achieve a stable state of human empathy. Therefore, the end-to-end learning strategy is not appropriate for machine artists.

**Authenticity and novelty of machine-generated art.** If the output of an AI painting system is too realistic, it will resemble the machine recreation of a photo. If the output is too abstract and grotesque, it may be incomprehensible and considered a cliché of nihilism. Therefore, a balance must be ensured between human-oriented reality and machine-oriented novelty. Achieving this balance is challenging and a central issue in the study of machine creativity. In addition, artistic creativity involves several essential aspects of intelligence, such as abstraction, association (e.g. metaphors, contrasts and anthropomorphisms), exaggeration, irony and fantasy. However, understanding of the basic principles, mechanisms and corresponding machine imitation technologies is limited in the scientific community.

## 2 Machine Artist Models

To address the aforementioned issues, we propose a machine artist model based on state-of-the- art generative AI technologies, which implements a symbiotic creative process based on human– machine and machine–machine interactions.

The essence of a Turing machine is to abstract computations as a sequence of operations over an input data stream. A Turing machine is a closed system: given a programme, Turing computations represent an end-to-end mapping from the input to the output, corresponding to a deterministic function calculus. Notably, this abstraction can only describe a behavioural system depending on its environmental interactions.

In contrast, Wiener cybernetics puts emphasis on the evolution of a system through the response generated by its internal control mechanism to the environment. It highlights the importance of feedback from the environment to achieve the desired control objectives. Wiener's behavior-based intelligence model serves as the theoretical foundation for research in robotics and intelligent agents within the field of AI. This model also forms the basis for significant machine learning algorithms such as neural network backpropagation (BP), reinforcement learning (RL), generative adversarial network (GAN) **[145]**, and inverse RL (IRL) **[146]**. The roots of these algorithms can be traced back to the inverse optimal control strategy that was proposed by Kalman **[147]**.

The relevance of cybernetics in AI has been reevaluated recently. Li et al. **[145]** conducted a comparison between Turing automata and Wiener cybernetics to elucidate the relationship between control systems and machine/human intelligence. Ma et al. **[148]** proposed two fundamental principles of intelligent systems, namely, parsimony and self-consistency, using Wiener cybernetics as the theoretical foundation. They also presented compressive closed-loop transcription as a comprehensive description of contemporary deep learning methodologies. These studies have provided insights into the connections between cybernetics and AI, shedding light on control systems, machine intelligence, and the development of advanced learning approaches. Luyao et al. **[149]** recommended the use of interpretable bidirectional communication to achieve machine–human value alignment.

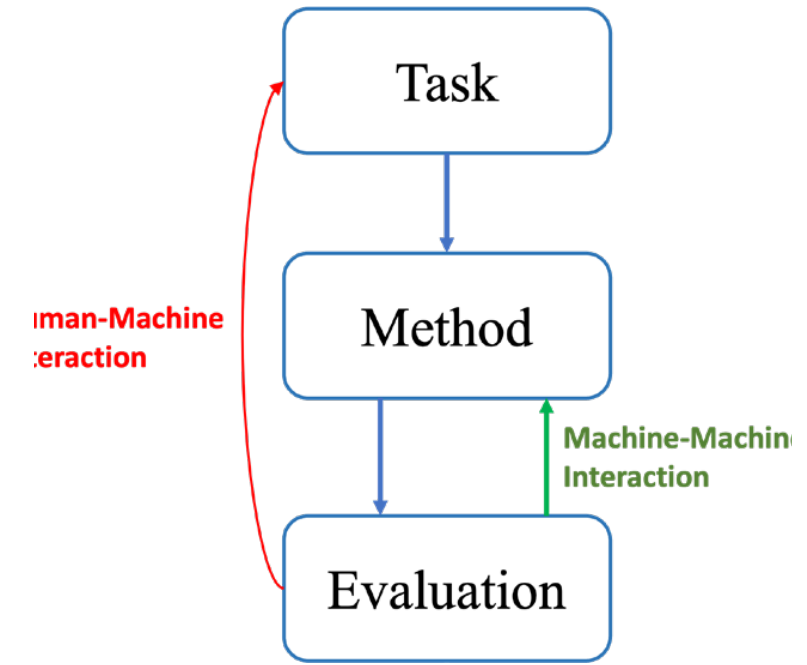

**Figure 23** General framework of a machine artist model

A stable system typically has a fixed control goal that is achieved through optimisation. Such a system obtains energy through interactions with the environment to attain a state of order. However, for creators, such as artists, such fixed-purpose assumptions regarding behaviours are often untenable. Artistic creation is the expression of one's emotions associated with environmental interactions, and people's emotions are imprecise and variable. Therefore, the design of a machine artist model requires the disruption of the traditional Wiener intelligent agent model and consideration of the dynamic changes in the creator's control goals, which are often guided by people's aesthetic feedback.

The interactive Natural Language Processing (NLP) paradigm emphasizes the symbiotic exchange of information between human actors and language models **[202]**. This approach aims to cater more effectively to user requirements and maintain adherence to human values, an idea encapsulated in the term Human-LM Alignment **[203]**-**[206]**. Historically, research on text generation predominantly focused on the intake and outflow of samples, neglecting considerations such as human preferences, experiential aspects, personalization, variegated requirements, and the process of text generation itself **[207]**. Recently, with the maturation of pre-trained language models (PLMs) and large language models (LLMs), the optimization of human-model interactions has emerged as a major focus within the research community. Strategies such as integrating human prompts, feedback, or configurations during either the model training or inference stages, utilizing actual or simulated users, have proven to be effective mechanisms to enhance Human-LM alignment [208]-[210].

Following this, we categorize human-in-the-loop NLP into three distinct types, based on the mode of user interaction, along with an extra segment that explores the simulation of human behaviors and preferences to facilitate scalable deployment of these systems. The categories can include

    (a) interaction through human prompts;

    (b) adaptation via human feedback;

    (c) control through human configuration;

    (d) education through human simulation.

This paper proposes a machine artist model that dynamically adjusts the task (i.e. objective function) subject to the artist's aesthetic standards by introducing dynamic human–machine interaction. Human-machine collaboration plays a crucial role in achieving a stable creative flow within the artist perception system. This is accomplished through the implementation of various components, including training based on multimodal embedding, generation that corresponds, characteristics of the system's response, and mechanisms for feedback at multiple levels. By combining these elements, the collaborative efforts between humans and machines enable the system to maintain a consistent and steady creative flow, promoting a harmonious and productive creative process. This approach is shown in **Figure 23**. Human–machine interaction leads to changes in the learning task (i.e. the utility function). IRL represents the first machine learning framework in which the utility function of a learning task can be autonomously learned [146]. The objective of RL is to learn a decision- making process to exhibit behaviour maximising a predefined reward function. IRL reverses the problem by learning the reward function from an agent's observed behaviour. Compared with the two frameworks, the utility function in the proposed model is learned as a more general manner in the context of machine creativity. Instead of being learned from the observation of the user's behaviour, the utility function is evolved from the aesthetic feedbacks of the artists and audiences.

## 2.1 Core Ideas

**Learn from human artists.** The dynamic human–machine interaction can guide the creative process and encoding of aesthetic values. Because the human artist is not expected to be extensively involved in the algorithm and system design, the aesthetics of the machine- generated artwork can be evaluated only by assessing the system input and output. The aesthetic evaluation process is iterative: in each iteration, the utility function of the system is changed by modifying either the system input or reward mechanism, based on the aesthetic judgement from the artist. These changes influence the model training and inference, and this complete process can be considered a generalisation of the IRL. The system must be designed such that its creations are consistent with human aesthetic values. Thus, human-in-the-loop is a core principle of the system design.

**Empathy embedding space with multimodal learning.** Robert Vischer [150] highlighted that the audience's appreciation of artwork depends on the empathy between the audience and artist, and a notable method of achieving empathy is synaesthesia. Wassily Kandinsky [151] described the essence of synaesthesia as 'heterogenous information from multiple sources inspiring inner beauty, fervour of spirit, and spiritual desire as inner necessity'. From the perspective of a machine artist, we believe that multimodal learning is an effective technique to achieve aesthetic empathy and synaesthesia, and its encoding of the semantic expression space corresponds to the 'inner beauty space' in aesthetics. Thus, an embedding space must be designed to encode the shared semantic meaning of the multimodal perceptual inputs. In other words, the dimensionality of the embedding space must be decreased, and the aesthetic semantics of the latent variables must be connected in the embedding spaces of different modalities. Such an embedding space is termed the 'empathy embedding space'.

**Creative learning.** In a transformer model, the encoder–decoder architecture must satisfy the self-consistency requirement. In other words, the learning process must ensure that, given the embedding generated by the encoder, the output of the decoder is sufficiently similar to the input of the encoder. Notably, such self-consistency is inadequate for creativity. Creativity refers to the expressed uncertainty in the intersection of multisource aesthetics-compliant information, as expressed by Henri Poincaré **[152]** 'Ideas rose in crowds; I felt them collide until pairs interlocked, making a stable combination.' In our design of the machine artist model, this uncertainty is introduced by random signals generated by sampling the embedded semantic space. The multimodal learning mechanism introduces synaesthesia and empathy, and the unsupervised/self-supervised learning and artist inputs ensure that the generated uncertainty conforms to the aesthetic standards of human art. Therefore, the only requirement is for the decoder to generate outputs from the empathic embedding space that are aesthetically related to the inputs and can be appreciated by humans.

**Multi-layered feedback learning.** As mentioned previously, when designing a machine artist model, the dynamic changes in the artist's creation objectives must be considered. In a symbiotic creative process, such changes are guided by aesthetic feedback from the artists. Therefore, the feedback system is multi-layered: it includes the inner gradient feedback (which decreases the local entropy) for the next creation iteration for a specific creation goal and the artist's feedback to the input, output and objective function of the complete system (which increases the global entropy). The traditional AI and control systems do not incorporate the aesthetics-compliant entropy-increasing feedback introduced by human aesthetic feedback.

## 2.2 System Architecture

Considering the abovementioned principles, the system architecture of a machine artist model based on human–machine interaction and multimodal learning is established, as shown in **Figure 24**.

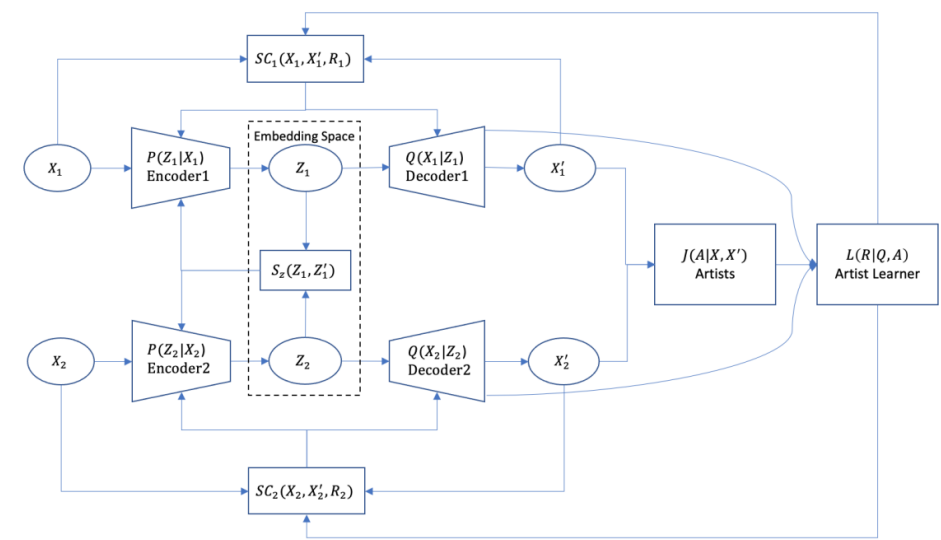

**Figure 24** System architecture of the proposed machine artist model

**Control Variables:**

- $X$: $X_1$ and $X_2$ are two data points from two resources, such as image, text, audio or motion information. $X_1'$ and $X_2'$ are the sampling results from the decoders.

- $Z$: $Z_1$ and $Z_2$ are the hidden variables or embedding semantic representation vectors generated by the encoders.

- $R$: $R$ is the output of the artist learner. This output may be a reward function or random variable representing how the artist's feedback influences the input guidance information and objective function of the generative model.

- $A$: $A$ is the artist's comments regarding the input and output in a time-series form, i.e. {$A$1, $A$2, $A$3, …}

- represents the model parameters to be optimised: $P(\cdot)$ and $Q(\cdot)$.

**Transfer Functions:**

- $P(\cdot)$ – Encoder, a learning model like a transformer or residual network, which is used for the compression of information.

- $Q(\cdot)$ – A decoder, a deep learning model utilized for generating content based on sampling methods.

- $S(\cdot)$ – Similarity function, which is a type of measurement metric (such as a dot product) of the similarity/consistency between two embedding semantic vectors (such as two hidden variables or input and output samples).

- $SC(\cdot)$ – Similarity and creativity. $SC_i(X_i, X_i', R_i)$ measures the similarity between two embedding semantic vectors as $S(\cdot)$ and introduces the *i*th reward function learned from the artist to optimise the output based on the artists feedback. This term can be considered a relaxation or regularisation factor of the similarity metric to incorporate the potential aesthetic meanings.

- $J(\cdot)$ – Artist judgement, which represents the subjective comments and modification suggestions provided by the artist regarding the image colour, content layout, melody, rhythm and text description, among other aspects.

- $L(\cdot)$ – Artist learner, which is an incremental and iterative learning system that learns the aesthetic metric from the artist. In the terminology associated with RL, the decoder can be considered the environment model, artist judgement $J(\cdot)$ is the action and aesthetic metric is the reward function. The output of $L(\cdot)$ is the reward function, which represents the aesthetic metric to optimise the encoder and decoder without the limitation of the strict or concrete matching in object recognition.

**Utility Functions and Optimisation**

The overall symbiotic creation can be formulated as follows based on Bayesian propagation:

$$Q_{t+1}^* = arg \max_Q p(R_t|Q_t)P(Q_t)$$

where $Q$ (or $Q(\cdot)$) represents the decoder, generator (in GAN frameworks) or policy (in RL frameworks). $R$ is the reward function $R(\cdot)$ (in RL frameworks) or discriminator (in GAN frameworks). The challenging posterior optimisation can be performed in a facile manner through adversarial IRL **[153]**. $Q$ represents the creative ability of the machine, and $R$ is introduced to iteratively incorporate the aesthetic feedback from the artists into the symbiotic creation loop.

The optimisation process involves two phases that are iterative and driven by the guidance of the artist.

**Phase 1:** Given a fixed reward variable, $R_i$, the encoder and decoder are optimised by maximising the weighted sum of the combination of the similarity and creativity metrics, termed the utility function:

$$\max_{\theta_{P,Q}} \{\alpha \cdot S_1(X_1, X_1', R_1) + \beta \cdot S_2(X_2, X_2', R_2) + \gamma \cdot S_z(Z_1, Z_1')\} \quad (2)$$
$$s.t., \alpha, \beta, \gamma \in [0,1], \alpha + \beta + \gamma =$$

The definition of $S_i(X_i, X_i', R_i)$ is not trivial; however, an intuitive and feasible solution exists:

$$S_i(X_i, X_i') = dot(X_i, X_i') \quad (3)$$

$$S_i(X_i, X_i', R_i) = S_i(X_i, X_i') + \lambda e^{-R} ||\theta_{P,Q}||_2, s.t., \lambda \in [0,1] \quad (4)$$

where $dot(\cdot, \cdot)$ is the dot-product operator, is the energy of reward $R$, $||\cdot||_2$ is the $L$2-norm and is the weight used to ensure a balance between the multimodal-learning-based machine creativity and human-aesthetics-based artist judgement.

Eq. $S_i(X_i, X'_i, R_i)$ can be intuitively explained as follows. When the artist is satisfied with the output of the machine, i.e. *R* is large, the regulator weight is small. In this scenario, the machine has a higher level of self-trust in terms of art generation and creation. When *R* is small, more generalisation and uncertainty are introduced to avoid overfitting in multimodal learning. This mechanism is similar to that of art professors encouraging students to expand their imaginations without being restricted to historical artwork.

**Phase 2:** Given the fixed reward variable of decoder *Q*(·), the reward function is optimised by solving the maximum likelihood problem:

$$Q^*_R = \arg\max_{\theta_R} E_{X'_t \sim [X'_0, X'_1, ..., X'_T]} [\log(p(X'_0) \prod_{t=0}^{T} Q(X'_{t+1}|X'_t, X_t, A_t) e^{\gamma^t R_\theta(X'_t, A_t)})] \quad (5)$$

where $Q_R$ represents the parameters of the reward function *R*, $R_t$ ´ is the content output of the decoder named 'state' in the RL framework) at time *t*, $(X'_{t+1}|X'_t, X_t, A_t)$ is the decoder (named 'transition distribution' in RL; the difference between this term and that in the proposed framework is that we also consider the multimodal input and artist evaluation or modification action $A_t$ at time *t*), is the decay factor and $R_\theta(X'_t, A_t)$ is the reward value given the content output $X_t$ ´ and artist evaluation $A_t$.

The artist's participation in the system can be summarised as follows:

- *Global artistic direction* in Phase 1: The reward value *R* reflects the artist's rating of the content output of the system and informs the machine how much to trust the strict matching between the sematic embedding vector in multimodal learning or the artist's judgement.

- *Local artistic direction* in the content generation by the decoder *Q*(·) in Phase 2. The artist's judgement in the form of the semantic embedding vector by multimodal learning directs the gradient optimisation of the iterative content generation. A representative example is presented in **[154]**.

### 2.3 Tasks and Process Flow

Given the control variables, transfer function and optimization objective function, we describe the processes of the three core tasks: multimodal training, content generation and understanding of the artist's aesthetic values, based on the architecture of the machine artist model.

**Multimodal Training**

The classical multimodal training process involves encoder and decoder training and similarity- based objective optimization. The novelty of our proposed multimodal training approach is that the self-consistency is not necessarily strict in the embedding semantic space. For example, the embedding vector of the word 'peace' may be similar to that of 'dove', or, more abstractly, 'circle' **[151]** owing to their intrinsic relationship related to aesthetics from the viewpoint of the artist. In classical multimodal training (e.g. contrastive language–image pretraining, CLIP **[154]**), the similarity between two concepts can be encoded, e.g. 'wave' may be similar to 'cloud' because the hierarchical layers of deep neural networks can capture the similarity in their physical characteristics. However, this encoding is too straightforward and not explicitly guided by the artist in terms of aesthetics. Therefore, we propose the function *SC*(·) to allow the reward function learned from the artist to be considered in the process of measuring the similarity.

**Content Generation**

To generate content in the inference process, the hidden variable *Z* is generally constrained by the models as the random noise, while the *Z* in multimodal models representing some compressed information with specific semantic meanings. Even so, both of the two models could be viewed in the unified framework of Encoder and Decoder. Thus, in addition to the encoder and decoder associated with multimodal training, another encoder and decoder should be trained for content generation. For simplicity, we only present the encoder and decoder associated with the multimodal training in **Figure 2**.

The input for content generation is the cross-modality information (such as text, image, music or motion) and embedding semantic vector from the artist's judgement, which guides the targeted gradient descent for iterative optimisation performed, for example, by the guided diffusion model **[154]** or GANs **[155]**. Similar to the process of a human creating art, the machine artist model must implement multiple trials to satisfy the artist. This iterative process allows the machine to automatically learn the artist's intent or aesthetic metrics through adversarial IRL. Consequently, the machine becomes smarter and more artistic, and beyond all expectation, may generate innovative and artistic content that may surprise the artist.

**Understanding the Artist's Aesthetic Values**

This process is implemented in Phase 2 in the optimisation process. To clarify the formulation of the utility function and optimisation, we show the correspondence between the considered problem and classical IRL. The control variables and transfer functions are related as follows by simplifying the quintuple $(S, A, P_{sa}, \gamma, R)$ in IRL **[153]**:

- State space $S$ corresponds to the set of the outputs of $\acute{X}$, i.e. artefact $\acute{X} \in S$.

- Action space $A$ corresponds to the subjective comments and modification suggestion $J(A|X, X')$ provided by the artist, in terms of the image colour, content layout, melody, rhythm or text description, among other aspects.

- Environment $P_{sa}$ corresponds to the decoder associated with content generation, i.e. $Q(\acute{X}t+1|\acute{X}t, Xt, At)$

- Discount coefficient $\gamma$ is the weight determining the significance of an artist's opinion on modifying the current learning task. This value is typically large at the beginning of the process and decreases towards the end.

Reward function $R$ corresponds to the mapping from the current state or output to the reward, subject to the distribution of $L(R|Q, A)$. (Note: $R$ is the reward function but may be the output variable of the reward function depending on the context.)

Despite these formal relations, machine creation involves several complex issues that must be addressed. Our proposed framework represents a starting point instead of a thorough solution to solve the problem of machine creation. The challenges that remain to be addressed can be summarised as follows:

Our framework involves many iterative loops such as those associated with the adversarial IRL, multimodal learning, content generator learning, guided and iterative inference for content generation, reward function learning based on super-sparse judgement by the artist, and human aesthetic learning. None of these processes can be managed by stable learning and inference based on the cutting-edge technologies in the multidisciplinary domain of AI, neuroscience and psychology. Additionally, these models must be trained in a unified framework.

A notable assumption is that the artist's judgement is the optimal strategy for content generation. Logically, in this case, the upper bound of the artistic taste of the machine artist model is expected to converge to that of the artist. In fact, the intrinsic uncertainty of the machine, either structural or random, can potentially extend its artistic initiatives, which may even surprise the artist with some artistic values. Moreover, the artist may be inspired by the machine's creativity, leading to the emergence of another novel creative idea. This process represents the essence of symbiotic creativity, which is different from the simple human– machine multimedia interaction encountered in the traditional techniques. Our target, although very challenging, is to enable the machine intelligence with adequate autonomous explainability and creative inspiration, along with a certain extent of controllability.

Several technical challenges remain, for example, the large computational cost of multimodal learning, sample sparsity, instability of adversarial training, sparsity of the artist's judgements, ambiguity of the reward function learning, error accumulation and propagation of long- sequence content generation, and uncontrollability of local object motion. These issues can potentially be addressed with further research in the domain.

**System Function**

The overall system operation is illustrated in **Figure 25**. First, multimodal training is performed across text, image, audio and motion information, depending on the creation requirements in step (1) This step involves a costly training process using a massive dataset or pretrained models such as CLIP **[154]**. Next, the generator model, such as a diffusion model or GAN **[155]**, is trained in step (2). Guided by the description or condition variable in the form of embedding vectors transformed by the multimodal model, the generator generates the corresponding content as the output of step (3). Subsequently, the artists evaluate this output and provide subjective comments for further modification in step (4). The comments are transformed by the multimodal model into embedding vectors that guide the gradient descent iteration in the content generation in step (5). The loop from step (3) to step (4) continues until the condition for the next IRL-based reward learning is satisfied in step (5). The new reward function, output by the reward leaning process, corresponds to two possible choices: return to step (1) to trigger a costly multimodal retraining process or return to step (2) to start a new round of generator training (indicated by the dotted line), which is preferable. After these iterations, the artist reviews the performance of the artist model and returns the process either to step (3) for generating additional samples or to step (2) for retraining the generator depending on the balance between the computational cost and artist's expectations.

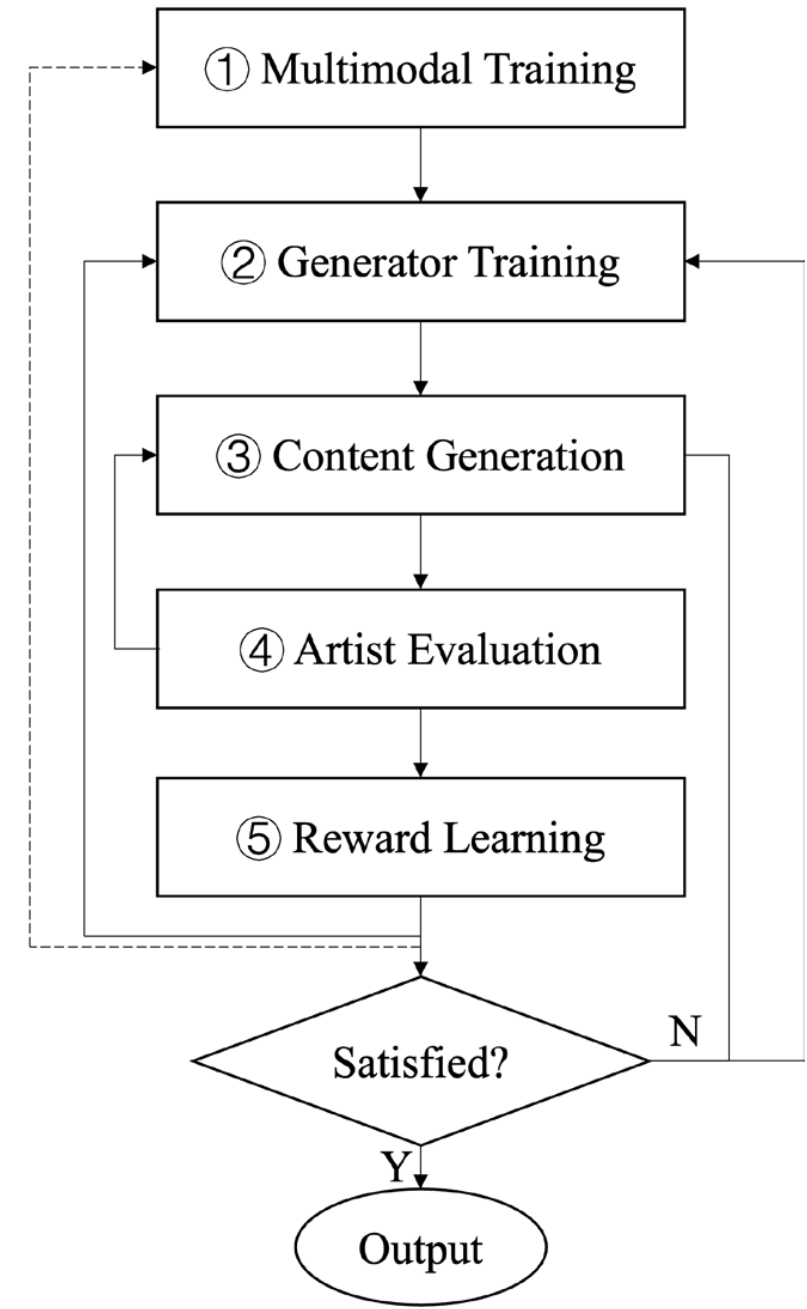

**Figure 25** Process flow of the proposed system

# 3 Comparison with Generative Models

## 3.1 Generative Arts with Large Language Models

Language is a structured communication system that incorporates grammar and vocabulary. It serves as the principal medium through which humans express meaning, in spoken, written, and other formats. Over thousands of years, humans have devised various communication methods, with language being fundamental to all forms of cultural and technological interactions. It supplies the words, semantics, and grammar necessary to communicate ideas and concepts [211]. Now, with the fast development of AI technology, especially with its capabilities in modelling human language, it is starting to permeate all fronts of human activities that are much wider than the scope of language itself. Especially, with the recent development of pre-trained language models, a new door has been opened. We will briefly review how language is playing its role in the era of AIGC.

All Language Models (LM) initially undergo training on a specific dataset. They then leverage different framework configurations to discern relationships and subsequently produce new content based on the data they were trained on. Based on the LM, we can have a lot of downstream NLP applications. As have been discussed in Part 1 of the report, when the language sequence to be modelled becomes longer and more complex, problems in context association and sequential processing efficiency will occur to sequential models such as RNN and LSTM. These situation have been changed with the introduction of Transformer [24], which provides an highly efficient alternative that enables full-attention and parallel context understanding in parallel enabled the development large-scale training for Large Language Models (LLM).

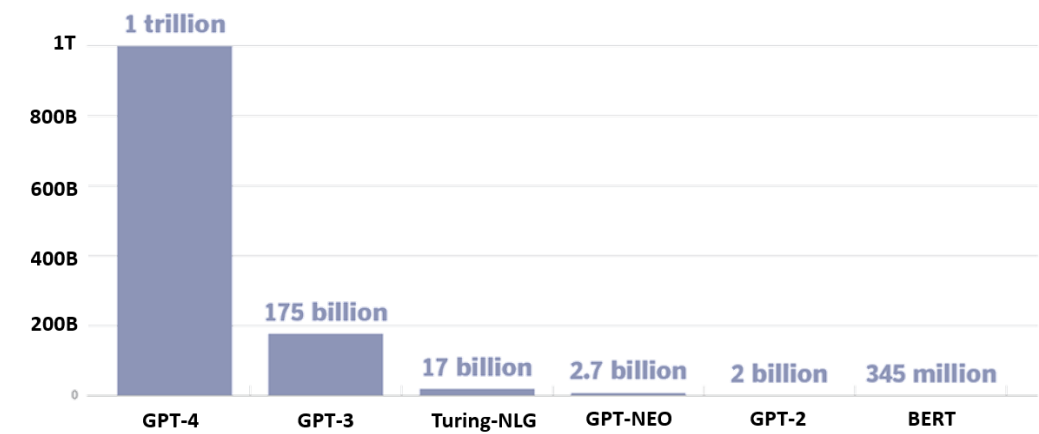

**Figure 26** Parameters of Transformer-based language models

Typically, a LLM has at least one billion or more parameters. As illustrated in **Figure 26**, the number of parameters for LLM have been sharply increasing in the past few years with the introduction of new LLMs. Certain Large Language Models (LLMs) are known as foundation models, a term introduced by the Stanford Institute for Human-Centered Artificial Intelligence in 2021. These foundation models demonstrate substantial knowledge capacity, serving as a base for further optimizations and specialized use cases. LLMs have gained substantial popularity due to their wide applicability across a range of Natural Language Processing (NLP) tasks. These tasks include text generation, translation, content summarization, content rewriting, classification and categorization, sentiment analysis, conversational AI, and chatbots, among others. ChatGPT [212] is one of the most well-known LLMs. What makes it different, besides using more data and computational resource to build the model, human feedbacks have been involved to fine-tune the model to give more correct and coherent answers to your question.

Language Models in Generative Arts. LLM starts to attract research attention in various human activities especially in generative visual arts. One of the foundation work is the introduction of CLIP [154]. CLIP is an advanced OpenAI model that connects images and text in a shared cross-modal embedding space. Relying on contrastive learning techniques, it learns to associate images with corresponding textual descriptions over a training dataset of 400 million image-text pairs. The semantically meaningful and continuous word embedding space enables to process or generate unseen visual features (creativity). Visual encoders/decoders learn passively from the supervision. Learning from natural language not only assists in acquiring a representation, but also connects that representation to languages. This connection enables flexible zero-shot inference and the generation of images that are semantically and stylistically controllable based on text prompts [213]. GLIDE [214] introduced text-guided diffusion models that successfully integrated textual information into the generation process [215]. Stable Diffusion [216] involves the text encoding from CLIP and deeply couples it to the denoising process in the latent space, which is more computationally efficient. Another commercialized text-based image generation tool is Midjourney [217], which enables user-controlled, prompt guided image generation. You can choose to which direction the diffusion process is heading, and then guide the model to further produce higher resolution images with more details.

The semantically meaningful and continuous word embedding space benefitted from large-scale language model pretraining and cross-modal context representation association proves to be a powerful tool which revolutionized how machines comprehend and work with multimodal data. The capacity and precision of the LLM help to re-distribute multi-modal embedding space for various semantically/style controllable content generation.

## 3.2 General Framework of Generative Models

The existing mainstream generative models **(Figure 27)** have an intrinsic relationship with the proposed machine artist model. A generative model typically has a coupled encoder–decoder architecture [148]. For example, the discriminator and generator of a GAN [155] correspond to an encoder and a decoder, respectively, both of which are iteratively optimised. The difference is that the discriminator serves as an objective function that is dynamically and iteratively optimised. This mechanism is representative of the feedback gaming mechanism advocated by Wiener cybernetics. A variational autoencoder (VAE) [4] constrains a hidden variable $Z$ as the embedding subjected to a Gaussian distribution learned by optimising the lower bound of Kullback–Leibler divergence. A flow-based model [156] directly optimises a log-likelihood objective function in which the flow (encoder) and inverse (decoder) form a generative pair.

Recently, diffusion models [157] have attracted considerable research attention as they can ensure diversity in generation. Diffusion models adapt training data by progressively incorporating Gaussian noise, and then they master the process of retrieving the data by reversing this noise-introduction procedure. This concept is similar to that of VAEs in that an objective function is optimised by first projecting the data onto the latent space and then recovering it to the initial state. However, instead of learning the data distribution, the system aims to model a series of noise distributions in a Markov *c*hain and 'decodes' the data by undoing/denoising the data in a hierarchical manner. Diffusion models outperform the other models in terms of diversity, but the computational cost for sample generation is high owing to the serial iterations. Nonetheless, the diversity of output makes diffusion models promising candidates for art generation. The variation can be regulated through assorted feedback mechanisms that mirror the semantic significance of the multimodal inputs and human aesthetic assessment.

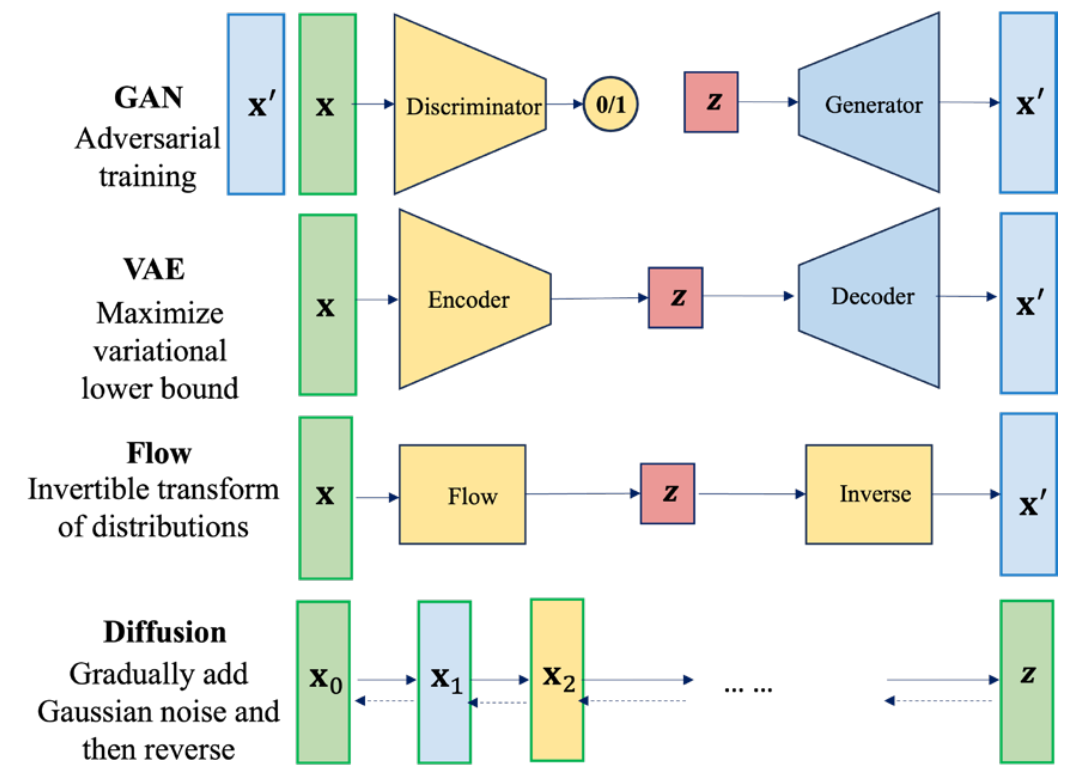

**Figure 27** Comparison of classic generative models [158]

# 4 Demonstration of the Proposed Framework

The functionality and performance of the proposed machine artist model were demonstrated through an innovative human–machine collaboration between the HKBU Symphony Orchestra and AI super-artists built based on the machine artist model at the Annual Gala Concert of HKBU at Hong Kong City Hall on 14 July 2022 [159].

In a few months, the machine artist model was engaged to instruct AI artists. After acquiring wisdom from an expansive "art data repository", these artists had the ability to sing, sketch, and dance in reaction to the music and lyrics they "heard". The AI virtual choir, which presented the song "Pearl of the Orient" employing the voices of 320 virtual performers, was trained by separating and deciphering key vocal elements from an anthology of songs documented by skilled singers.

Moreover, the AI media artist conceived a cross-media visual narrative of the song, aligning with its comprehension of the song's lyrical meaning. The visuals evolved in tandem with the melody and lyrics, delivering an immersive cross-media experience to the viewers. Contrary to traditional AI machines which use images as a guide for the algorithms to emulate, the AI media artist utilized textual lyrics as the unique input to generate vibrant artwork, demonstrating the progression of visual storytelling.

A ballet rendition of Ravel's "Daphnis et Chloé" was executed by AI virtual performers. Professional dancers from the "Hong Kong Dance Company" lent their expertise to the AI virtual dancers in deciphering the emotional and aesthetic ties between the music and dance. These dancers subsequently orchestrated the dance motions, drawing inspiration from a recently identified species of box jellyfish in Hong Kong.

These occurrences signify a noteworthy advancement in employing AI for crafting AI-enhanced performances that encompass not just a cooperative procedure between humans and machines, but also diverse art forms. The three pivotal accomplishments are encapsulated in the ensuing text.

## 4.1 Song-Driven 3D Narrative Video Generation

Based on the proposed machine artist model, we preliminarily implement a system of automatic 3D narrative video generation driven by the lyrics of a song, 'Pearl of the Orient', and its melody. The system framework is shown in **Figure 28**.

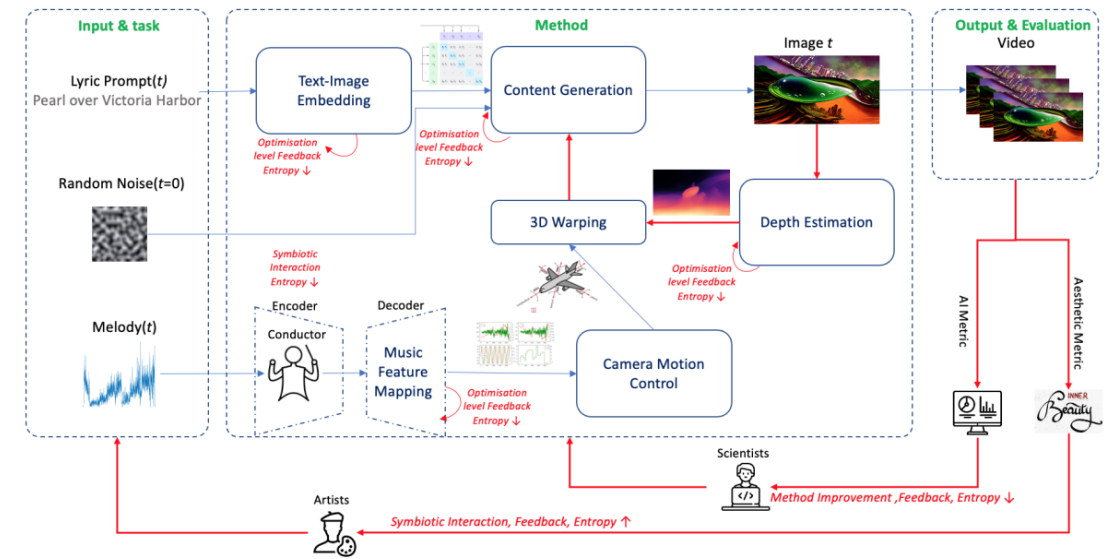

**Figure 28** Framework of song-driven 3D narrative video generation

This framework involves the following functional novelties:

Cross-modal artistic content generation. The input is a song (such as 'Pearl of the Orient' or 'On Wings of Song') consisting of melody and lyrics (with some minor manual modifications), and the output is a 3D narrative video consistent with the song content. No additional images are required, and all images are encoded in the pretrained multimodal models.

The output is a long serial of ~3000 frames lasting for ~4 min with a high resolution of 720P. No post-processing or human intervention is involved in the generation.

The objects in the video smoothly transform with magical visual effects, e.g. from stars to crystals, from clouds to Hong Kong bauhinia flowers, and from flowers to balloons.

The camera motion is fully driven by the melody and harmony. For example, a higher pitch corresponds to a higher viewpoint, and a faster rhythm corresponds to faster camera movement.

The system exhibits unique artistic creativity, and several creations may even surprise the artist. For example, as shown in the last image in Figure 29, the system generates a water-drop-like transparent pearl. The machine automatically transforms Victoria Harbour in Hong Kong from the Euclidian geometric space into the curved space of the pearl, which represents an artistic visual effect to match the subject of the song 'Pearl of the Orient'.

The content of the output video is narrative and consistent with the semantic meanings of the lyrics.

**Implementation and Results**

To ensure feasible engineering implementation, we prepare a simplified generation system based on the architecture shown in **Figure 2**. To minimise the computational cost, we try to avoid triggering the retraining of the multimodal and content generation models. If the evaluating artist is not satisfied with the results after several rounds of content generation, we consider finetuning the pretraining models based on samples closely related to the creation subject.

The generation is driven by the song. The lyrics are transformed by the multimodal models into the embedding semantic vectors to guide the iterative image generation. Because the sentences in the lyrics are logically related, the generated sequential images are naturally narrative.

The melody plays a unique artistic role in this system. The camera motion guiding the angle and focus of the generated image is driven by the melody. One intuitive method to link melody and motion is to map the acoustic features to the six freedom variables of the camera motion in the 3D space, based on the artist's guidance. We used a more innovative approach, in which the conductor in the concert is considered to be an encoder in the system. The complex musical signals are compressed into simpler gestural motions of the conductor. Subsequently, we design or train a decoder to map the conductor's gestures to the camera motion. This mapping results in an artistic resonance between the melody, camera motion, and image content and introduces a new dimension of performance through which the conductor can control the video.

To generate the 3D video content, we use depth estimation methods (such as MiDaS **[47]**) to obtain the depth information of the $t$th frame, followed by 3D warping. The warped frame is considered the initial conditional input to generate the ($t$+1)th frame, guided by the lyrics. The novelty is ensured by multiple trade-offs between the initial frame and guided lyrics, diversity and temporal consistency, motion consistency to melody and human perception, and realism and artistic abstraction.

Descriptions of other interesting demonstration systems can be found in **[160]**–**[162]**. We referenced the concepts and implementation skills of these systems to ensure appropriate operation of the proposed system.

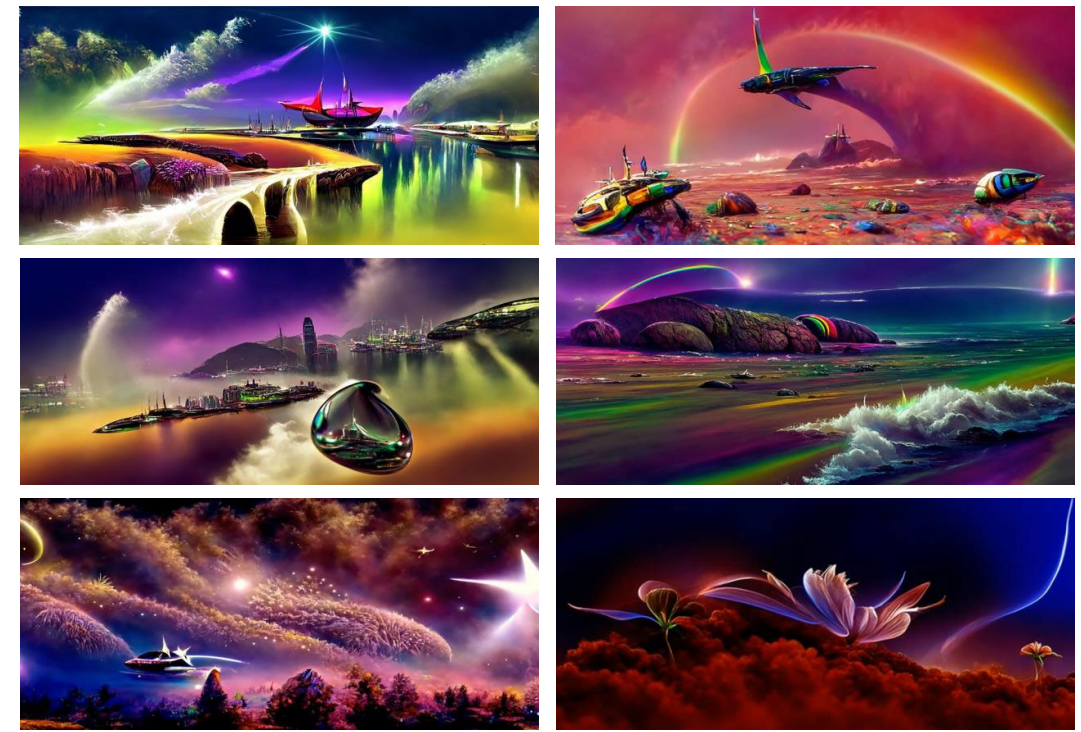

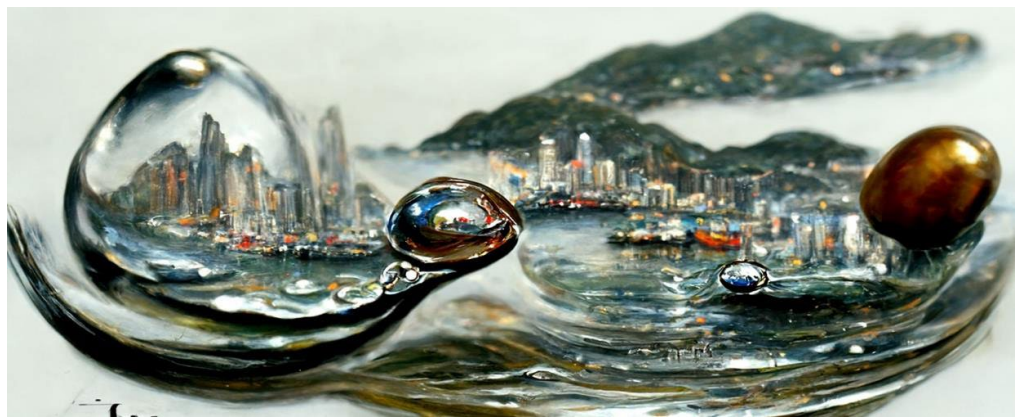

**Figure 29** Several frames of the output video for the song 'Pearl of the Orient'

## 4.2 Language models for motion and dance generation

Dance is a "language" composed of expressive gestures and body configurations, through which non-verbal communication can be accomplished **[218]**. This is a well-known statement that is often applied cross-culturally. In order to represent and model motion as language, we have to resort to an annotation system which could represent movement as symbols, phrases, and sentences.

For the professional choreographers and dancers, this choice is Labanotation, which is wildly accepted as the "language of dance". Just as in spoken language, dance has its own basic "units of expression." A well-structured grammar exists that outlines how these movement "units" relate to each other and their particular roles within the overall "phrase" of the dance. As shown in **Figure 30**, in this language of movement, the fundamental elements can be categorized as nouns, verbs, and adverbs, similar to verbal language. Movement signifies change, and to bring about this change, an action (the verb in this context) must take place. The body parts that are involved in the movement serve as the nouns. The manner in which the action is performed, the degree of change, or the style of performance is represented by the adverb. The figure below shows a diagram **[218]** where choreographers would relate textural elements with movement attributes.

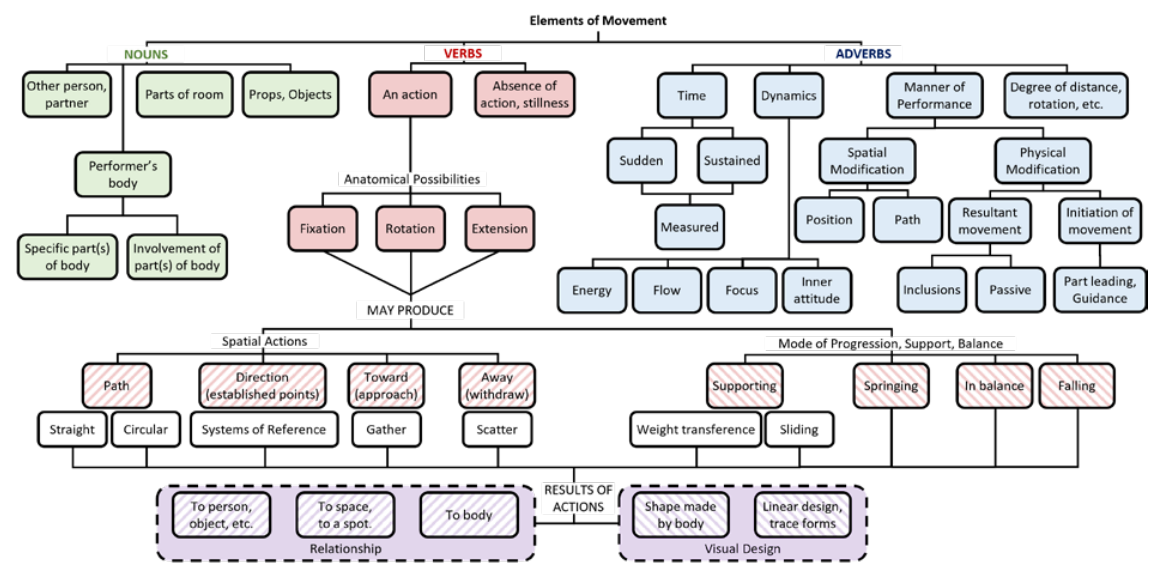

**Figure 30** Kinematic analysis of the relationship between textural and grammar elements with motion attribute concepts

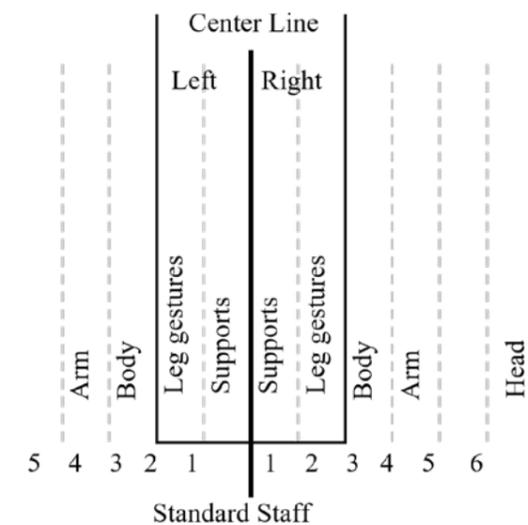

Labanotation is a system designed for the analysis and recording of human movement. When used to document dance, it is independent of genre and style. It is widely agreed among professionals to be capable of recording all types of human motion. As shown on the right and in **Figure 31**, Labanotation is written on a staff composed of three solid lines, and the scores start from the bottom and progress vertically upwards. The center line symbolizes the center of the body, separating the right and left sides. Vertical columns on each side represent major body parts as indicated. By positioning the movement event symbol in one of the vertical columns of the staff, an action for one of the body parts is denoted.

78    79

As further illustrated in **Figure 31**, each Labanotation symbol indicates attribute(s) of movement: the shape of the symbol in Labanotation indicates the direction of the movement; its shading specifies the level; its length represents duration (shorter symbols signify quicker movements, while longer ones denote movements that are more extended in time); and its placement on the staff identifies the body part that is in action. Additional symbols can be used to represent minor body parts (like hands, feet, chest, etc.), and other signs such as pins and hooks are used to denote details that modify the main action. An example Labanotation sequence is described in the right of **Figure 31.**

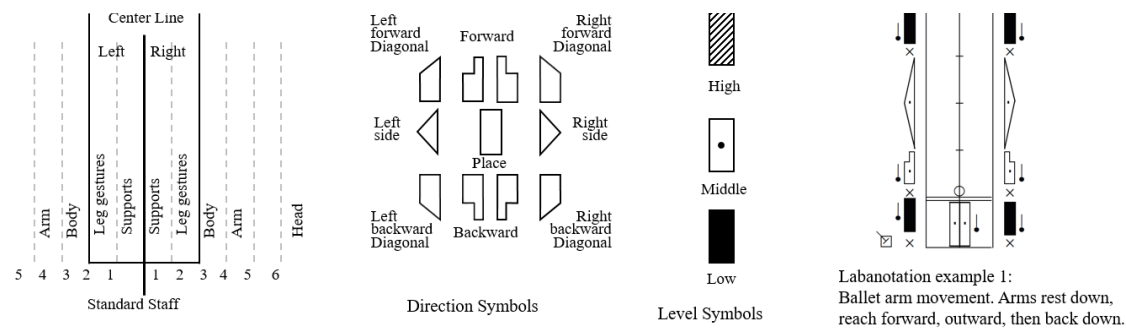

**Figure 31** Brief illustration of Labanotation description symbols with sequence examples

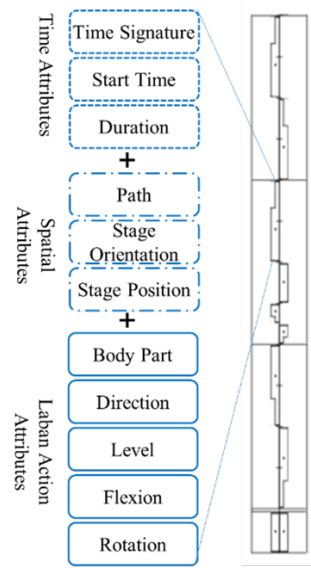

It is immensely challenging to *directly* model Labanotation using language models, as it does not strictly follow temporal sequential order (e.g., movement notes for different body parts often overlap), thereby flattening into a 1D sequence undermines the local structure of motion. As such, a *note-based* Labanotation Symbolic Token Representation (*LabanSTR*) is required for modelling movements without presumption of sequential orders among the tokens. As shown on the right, The *LabanSTR* is designed as a tuple of 3 attribute groups, which encodes different temporal, spatial, and movement characteristics of a Labanotation. The time attributes specify the time signature (similar to the same concept in music notes), start time, and duration; spatial attributes describe progression path, stage orientation and position etc., and action attributes specify the execution body part, the direction, elevation level, rotation and flexion for the action. Note that these *LabanSTR* events are *permutation-invariant*, as each symbol contains attributes of start time and duration, therefore the processing and generation orders do not affect the whole score's representation. Based on these elements we can now model the motion using language models.

### 4.2.1 Textural element composition for motion generation

Previous efforts have been made in analyzing the dynamics of motion embedded within the sequential combination of the Labanotation. For instance, the Laban effort graph [219] provides an inspiring study of how dynamics in symbol combinations can indicate physical and emotional dynamics.

Based on the unique textural and motion attribute relationship established by professional choreographers in **Figure 30**, we propose to directly learn the textural-motion correlation in a compositional manner, i.e., analyzing the motion (verb) and its attributes (associated nouns and adverbs) individually in the joint-embedding space. As illustrated in **Figure 32**, the *LabanSTR* sequence are organized as tokens and analyzed by the motion attribute encoder into the compositional latent vector space $\{\mathrm{mv}_k\}_{k=1}^{n}$, where each vector stands for a motion attribute that associates with the textual element word embedding.

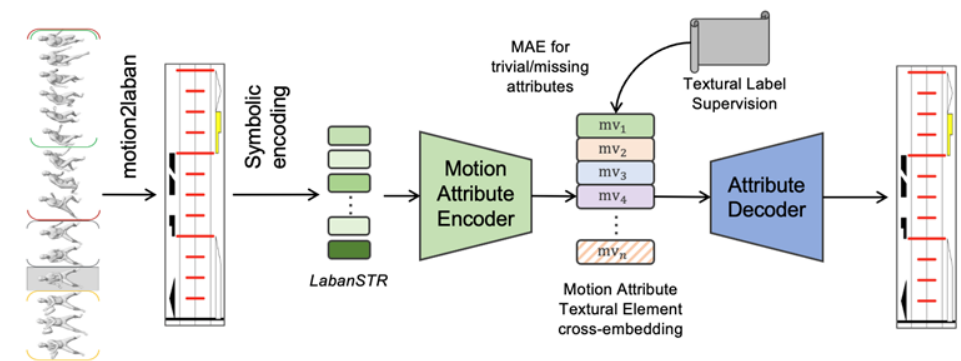

**Figure 32** Proposed framework for textual element and motion attribute learning

Text-labelled motion datasets [220] can be used directly for the model training. With this framework, complicated motion can be generated with detailed and delicate text control.

### 4.2.2 Symbolic Deep Phase embedding for motion progression and transition

On top of textural element studies, we can also rely on labanotation analysis for motion interpolation, transition, and morphing. To translate a long streams of text scripts into human skeletal motion, we need to first analyze and segment the script into textural elements and generate a sequence of motion clips based on the framework discussed in 4.2.1. To animate and cycle these elementary motions over the desired time interval, and then transit to the next elementary motion at the specified key frame, we need to make sure that the transition is natural, smooth, and conforming to physical constraints.

Based on the assumptions of DeepPhase [221], all motion can be modelled as combinations of cyclic elements. It is desirable to study the frequency, magnitude, and phase for these latent elements so that natural animation and transition can be conveniently achieved via exploring the learned phase manifolds. Note that while frequency and magnitudes are more related to motion intrinsic kinematic attributes, phase indicates temporal attributes that directly decides the progression and main controlling factor for transition. DeepPhase achieved projection to such latent space via a simple 1D among the human joints which is straightforward while too naïve. We believe by more explicit statistical analysis into motion attribute dynamics reflected in the combination of *LabanSTR*, and by analyzing the correlations between different body parts as separated by the Laban staff, such a deep phase space can be more accurately and easily modelled. This will be a key step for us to programme a long and natural action based on script streams.

### 4.2.3 The MotionGPT

Based on the symbolic representation as well as the natural connection between labanotation and natural language as analyzed in 4.2.1 and 4.2.2, we will work to develop a model that enables us to programme motion just like programmers are using Python, and to choreograph new pieces of dance just like chatGPT write an essay. As illustrated in **Figure 33**, MotionGPT, which is a motion foundation model, will not as large as common LLMs, as the grammars and vocabulary of labanotation is much smaller than natural language. However, we are working towards a pre-trained model to connect cross-modality context for motion programming: given conditions such as prompt text script, visual signals, music, etc, the MotionGPT which directly writes out labanotation.

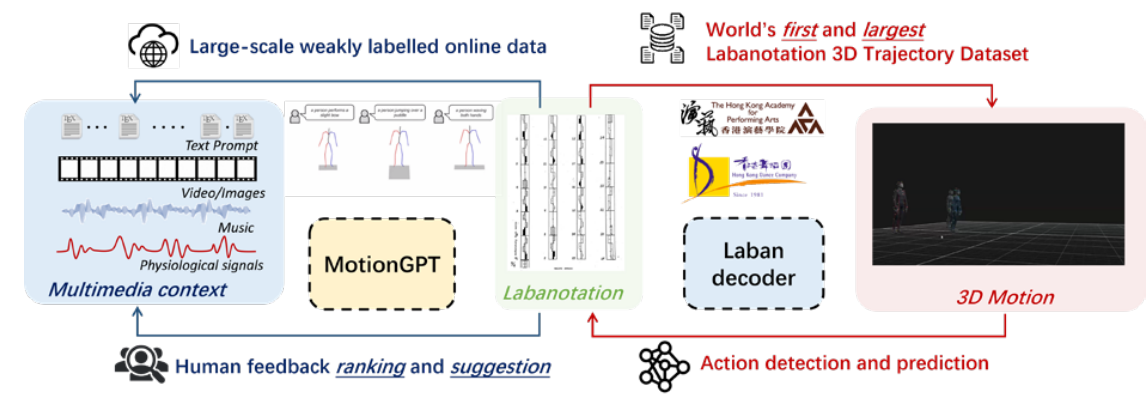

**Figure 33** The MotionGPT framework

We are now building a large dataset with multimedia context with paired labanotation. This labels is generated by our action recognition algorithms. **In addition, the labanotation works as a window and an interface for professional choreographers to provide human feedback to improve the intelligence of MotionGPT.** To learn a decoder to translate Labanotation to actual human motion trajectory, we are also building the world's first and largest Labanotation dataset. This is an ongoing effort in collaboration with the Hong Kong Academy of Performing Arts.

**Concluding remark.** Labanotation, unlike direct 3D trajectories, is a compact, and semantically meaningful symbolic language. Therefore, it is a much more semantically meaningful representation, that is dense, lower-dimensional, which can be easily projected to word embeddings and connect with the embedding space of other pretrained LLMs.

## 4.3 AI Choir with Human-in-the-Loop

Following the principles of the machine artist model, an AI choir can be generated by the feedback-controlled encoder–decoder framework involving both the machine and humans. Given a large amount of audio data collected either online or in studios from different singers, the machine aims to learn the embedding space in an unsupervised manner, expressing and disentangling the features related to the melody, emotion, timbre and rhythm of the song. The human artist provides essential initial guidance on how to control the embedding vectors in the embedding space for artistic music generation as well as continuous feedback regarding the different aspects of the generated music in a looped process.

Besides the accompaniment, the choir comprises dozens to hundreds of vocalists. In actuality, to yield a pleasing choir effect, genuine singers must execute with high synchronization concerning timbre, rhythm and expressiveness. Remarkably, an aggregate of identical sounds doesn't result in a choir effect; it emerges from the thoughtfully planned diversity of timbre, rhythm and expressiveness of each individual singer in a group performance. This aspect signifies a regulation issue for the AI choir: the local convergence of the coherence of a singer's voice and the diversity of timbres and universal harmony of the choir must be jointly enhanced. Throughout this procedure, the human must proactively furnish feedback to assure that the machine can maintain a balance between global and local enhancement and meet the human artist requirements.

The following text describes the AI choir generation framework and interaction between humans and the machine in the machine artist model for generating the AI choir. The AI choir generation framework consists of a universal virtual-singer voice generation model that yields singing voices with different timbres, and a singing-voice combination model that optimises the timbre proportions of the choir to obtain a satisfactory output.

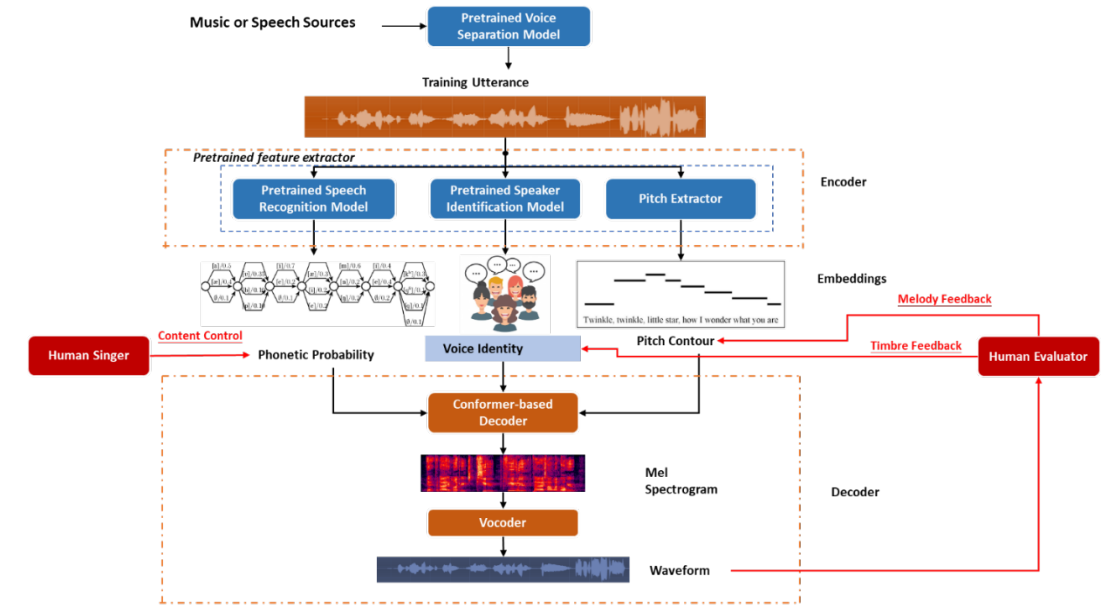

**Figure 34** Universal virtual-singer voice generation model with human-in-the-loop

**Figure 34** shows the universal singer voice generation model. The model has an encoder–decoder architecture, and the produced waveform is controlled by modifying the embeddings output by the encoder. Three pretrained encoders are used to comprehensively describe the input utterance, yielding three embeddings, which indicate the lyrical information expressed by the phonetic probability, speaker voice identity information and pitch melody. The embeddings are fed into the decoder to synthesise the time-domain waveform. By modifying the speaker voice identity embeddings, the timbre of the singing speech can be changed. The model can be trained by minimising the loss functions defined considering the consistency between the waveform outputs and original signal. However, because the signals will eventually be appreciated by humans, the human evaluator must be integrated into the framework to evaluate the artistic/aesthetic value of the melodies and timbre, corresponding to the function shown in **Figure 34**. The feedback is used to update the embeddings to be fed into the decoders **(Figure 34)**. Moreover, the embedding can be affected by the human singer, and this feedback provides the 'seed' song to control the aesthetic taste of the synthesised singing voice.

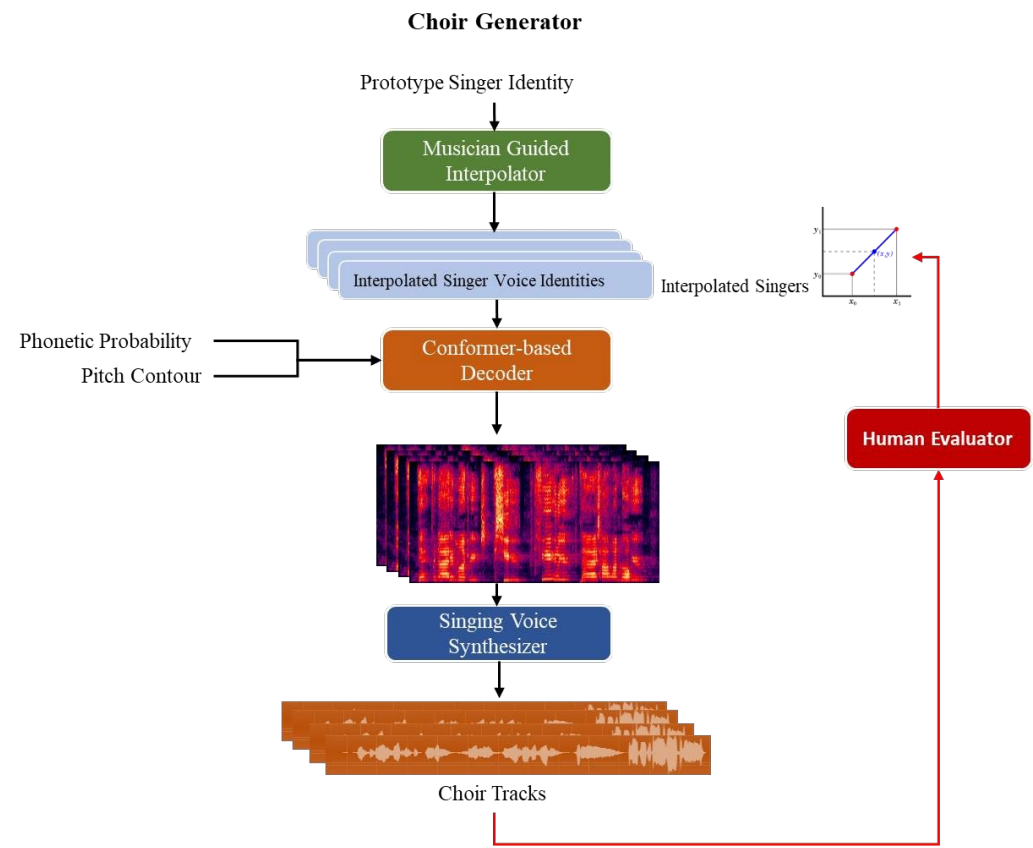

**Figure 35** Singing voice combination with human-in-the-loop

The human evaluator also holds a significant role in amalgamating the virtual singers to form the choir. As stated, the AI choir strives to enhance the diversity of each generated singing voice while still delivering a highly synchronized performance. As depicted in **Figure 35**, through the speaker identity embeddings extracted from the prototype singers in the training set, an indefinite number of interpolated singers can be created by linearly interpolating the embeddings of two prototype singers. Even though the combination of a considerable number of virtual singers to form the choir seems like a direct process, a human evaluator must review whether the choir possesses a satisfactory combined timbre and adjust the proportions of the prototype singers.

The proposed machine artist model represents a starting point for the development of art- generating AI. The core idea and underlying philosophy pertain to the belief that human–machine interaction is essential to promote the field of machine creativity. In this context, the traditional end-to-end learning model does not allow the machine to acquire knowledge of human aesthetics or the inspiration to express human emotions. An art-generating machine must be an open system that can understand humans and can be understood by humans. Therefore, the development of an art machine is expected to be a notable challenge for future AI development. The development of AI-based art technologies will not only be beneficial to the art community, but also facilitate the realisation of truly human-compatible AI.

# Part 4
# NFTs and the Future Art Economy

A conventional art market starts with galleries, which sell artworks to museums and collectors, followed by the secondary market, centred around auction houses. Art galleries and auction houses have long been the marketplaces of the art world, where museums, collectors and artists buy and sell artworks. Recently, this traditional market model has been disrupted by the innovation of non-fungible tokens (NFTs), which provide a revolutionary way of trading digital assets, including machine-made artworks, on the Internet-enabled global decentralised market.

In the realm of the digital world, particularly the 'metaverse', the economy hinges on the validation of digital property. NFTs are conceived to signify digital ownership by facilitating authentication of possessions, property, and even identity as *unique* and *non-exchangeable* assets. Given each NFT is preserved in a blockchain and safeguarded by a cryptographic key, it's impossible to erase, duplicate, or annihilate it. The absence of *fungibility* (exchangeability) sets NFTs apart from other blockchain cryptocurrencies. It enables the sturdy, decentralized confirmation of one's virtual identity and digital belongings. NFTs introduce rarity into an online environment where duplication has been consistently detrimental to content creators. As transactions are perpetually tracked on a digital ledger, it's feasible to identify who made any given purchase, when and for what amount. For content creators, an NFT can be divided to signify the fraction of shared contributions. In this context, NFTs challenge the concepts of copyright and ownership. They represent an economic innovation for the creative industry, as they permit creators anywhere across the globe to share and be compensated for their artwork. NFTs dismantle barriers for creators and collectors alike; they construct a new cosmos of digital content that can be bought and sold worldwide in an instant.

NFTs unlock the future of digital asset ownership. The NFT trading volume surged to new highs with $2.5 billion in sales as of the second quarter of 2021, up from just $13.7 million in the first half of 2020 **[163]**. Mainstream artists are increasingly launching their own NFT products, and well-known brands have also entered the field. Decentralised finance projects have made NFTs a core part of their business. Notably, the cultural and creative industries are greatly engaged with NFT development, as NFT technology has the potential to add considerable value.

NFTs enable a new way of working for artists and the art community, particularly in terms of obtaining copyright protection and monetary income for their creations. Shaped by blockchain platforms, NFT technology provides a direct communication channel between artists and art collectors. Blockchain solutions specifically designed for the art community will ensure that artists, galleries and art collectors can easily engage in this innovative process.

Mirroring the transformation of many other traditional industries, the art world will witness a new form of art economy with the coming of Web 3.0 (the metaverse), benefiting from its augmented creativity and sophisticated marketplace. In the art world, the new medium of blockchain technology endows digital copyright with additional commercial value above that of a mere currency or collection. NFTs are an open canvas. The development of artificial intelligence (AI) technology is changing artists' creative processes. AI can cooperate with artists to unleash the creative potential of NFTs and promote the development of a digital marketplace of AI-powered blockchain art.

The advent of NFTs is facilitating a myriad of applications. We list the most representative applications of NFTs as follows.

- *Boosting the gaming industry*: many NFT-based games are now available, including CryptoKitties, Cryptocats, CryptoPunks, Meebits and Gods Unchanged.

- *Fostering virtual events*: Virtual events facilitated by NFTs circumvent the issue of counterfeit or invalid tickets as an NFT ticket is exclusive to the holder, who is unable to resell the ticket post-purchase.

- *Protecting digital collectibles*: digital collectibles include digital images, videos, virtual estate and domain names. All of them can be essentially encoded as NFTs.

- *Supporting the metaverse*: users can earn profits from the virtual economy supported by the metaverse. During this process, NFTs play a critical role in implementing the metaverse.

Compared with traditional art markets, NFT-enabled digital art markets have the following advantages.

1) The difference between digital art and traditional art markets lies in copyright issues. As NFTs guarantee copyright, they enable easy circulation of digital artworks.

2) NFTs create transparent, reliable and cost-effective markets, thereby allowing users to trade collectibles easily and conveniently. This is useful for boosting the 'fan' economy.

3) Through the underlying blockchain technology, NFTs ensure the full traceability of digital artworks throughout their life cycle from creation to circulation.

4) NFTs can be integrated with physical assets, property, securities and insurance, thereby promoting a diversity of applications in art markets.

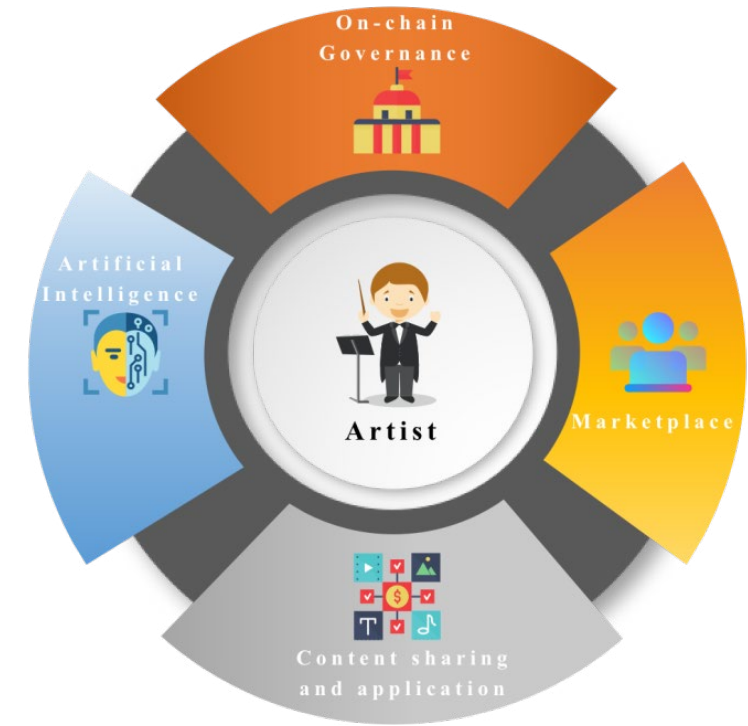

**Figure 36** New production relations based on blockchain

We should also carefully examine how NFTs could impact individual stakeholders in the art ecosystem:

- **For collectors:** The NFT market has passed authenticity review, and provides collectors with access to NFTs created by artists from the world's top art galleries.

- **For art copyright users:** NFTs help to bypass engagement with intermediary agencies so that users can not only enjoy an exclusive artistic experience but also circumvent the high-cost intermediate links in the chain of copyright use.

- **For artists:** NFTs establish a human–machine co-creating base that allows artists to customise their art metadata and enforce their copyright royalties on the blockchain.

- **For galleries:** NFT casting and sales platforms can be deployed online to empower artists and collectors using new technologies.

- **For museums and cultural institutions:** Digital art management on NFT platforms is an ideal form of collaboration. Donations and partnerships with cultural institutions will be made more flexible and diverse by the use of NFT platforms.

At the same time, a series of hurdles must be addressed, as with any emerging technology. These include energy consumption, cross-chain trading, regulation and oracles, covering both the system-level issues and human factors.

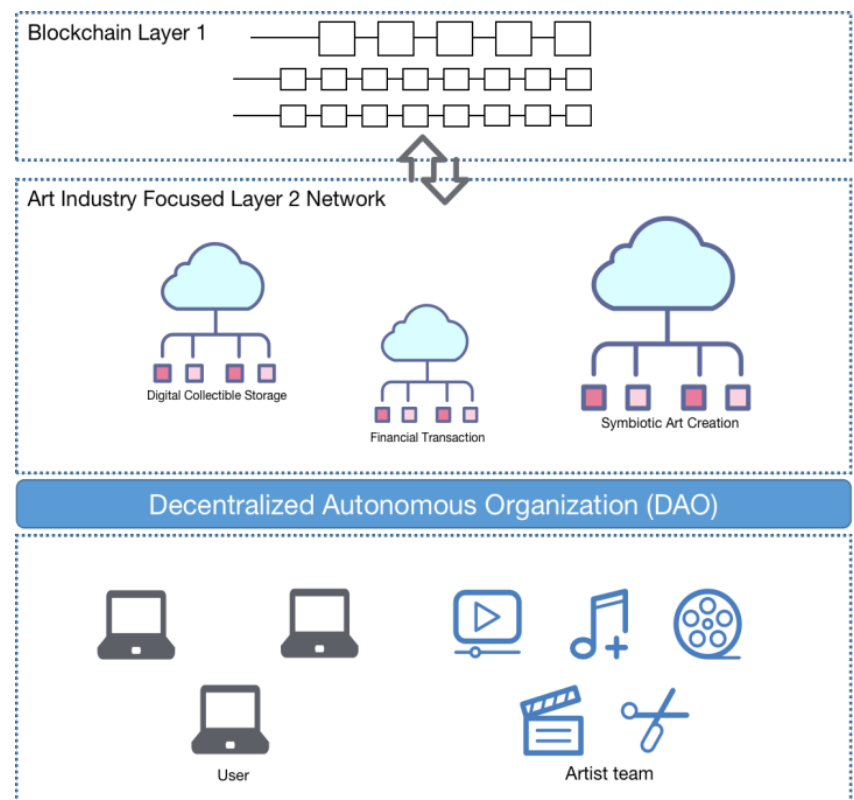

**Figure 37** Diagram of the art industry focused blockchain infrastructure

- **Energy consumption:** The substantial energy consumption of blockchain technology is a challenge that necessitates a more environmentally friendly NFT ecosystem. At present, in a typical NFT implementation based on Ethereum, up to 130 kilowatt-hours could be consumed to process a single transaction. The energy consumption associated with NFTs derives from two aspects, i.e. computation and storage.

  *A. Computation*: Currently, most blockchain systems rely on the Proof of Work consensus mechanism to realise decentralisation and guarantee system security. To process transactions and generate a valid block, all miners in the blockchain system compete with each other to solve a cryptography puzzle. This crypto-mining procedure is the main reason for the high energy consumption.

  *B. Storage:* To maintain the integrity of blockchain data storage and the execution of smart contracts, existing blockchain systems frequently necessitate each node in the network to store a complete copy of the transaction history and the ledger states. As of July 2021, the entire blockchain ledger for Ethereum surpasses 7 TB. With these ever-growing data volumes, such a cumbersome storage requirement also poses a significant energy cost. Therefore, to reduce the carbon footprint of the blockchain system, an energy-efficient NFT platform should be developed to enable a sustainable, green NFT ecosystem.

- **Cross-chain Trading:** Current NFT ecosystems are segregated from each other because of the presence of multiple underlying blockchain systems, like Ethereum and Flow. When users select a specific ecosystem, they can only buy and sell NFTs within the same platform, leading to data and value silos. With the proliferation of blockchain systems, cross-chain communication and interoperability have become barriers to the evolution of NFT marketplaces. Currently, most solutions leverage trustworthy external parties to facilitate cross-chain trading of NFTs. However, these inevitably violate the inherently decentralised property of blockchain. Worse, they may give rise to security issues. Consequently, it is worth exploring how to build a uniform trading platform to achieve seamless interoperability among multiple blockchain systems. Challenges to such a platform may come from many directions, including transaction atomicity, efficiency, security, semantic diversity and implementation friendliness.

- **Regulation:** While NFTs spark growing excitement, their legal status continues to evolve and remains unresolved. Typical legal issues include government management and market regulation.

    *A. Laws and policies:* Legal issues arise in a wide range of areas, including exchange, cross- border transactions, financial derivatives and data privacy protection. Various jurisdictions (e.g. China, the US and the UK) have reacted to NFTs in different ways. With new regulations being continuously enacted, compliance with regulatory terms is the first priority of NFT marketplaces. Therefore, it is necessary for the government to undertake due diligence and adjust marketplace construction accordingly.

    *B. Intellectual and taxable property:* Users who are unfamiliar with the legal constraints pertaining to NFTs might incur potential infringement liability when participating in the NFT market. For example, an NFT buyer may believe that they have acquired the underlying artwork linked with the NFT. In fact, the original creator is the copyright owner, retaining the sole rights, while the buyer typically only obtains the token and the permission to use the copyrighted artwork for personal use. In addition, under the present legislative framework, intellectual property-related items are considered taxable property. NFT-based sales, however, are not covered by this framework. This might significantly increase the number of financial crimes committed under the guise of NFT trading. Consequently, it's recommended that traders involved with NFTs seek further counsel from proficient tax departments.

- **Oracle:** An oracle is a mechanism to feed off-chain data (from the outside or real world) into a blockchain's virtual machine so that relevant smart contracts can be invoked to trigger predefined state changes on the blockchain. The condition invoking the smart contract could be represented by any type of data, e.g., artwork generation, bidder enrolment, payment status or price fluctuations. The main challenge facing oracles is how to verify the correctness of off-chain data. Existing oracle mechanisms can be categorised into centralised and decentralised models. A centralised oracle relies on a single source of truth, which could make it susceptible to single-point failures and insider attacks. Decentralised oracles increase the reliability of the data by querying multiple data sources and offering incentives for honest reporting. However, they remain vulnerable if supplying manipulated data could be more profitable than behaving honestly. Decentralised oracles are not involved in the main blockchain consensus, and thus data security cannot be guaranteed. These considerations open up new opportunities to design secure oracle mechanisms for the blockchain-based NFT marketplace.

# Part 5
# Ethical AI and Machine Artist

As the applications of AI become increasingly pervasive and consequential, questions about their social impacts and ethical implications are frequently raised and discussed. These discussions are not confined to the academic circles of ethicists or scholars of technology in humanities and social sciences. Not only have many governments, organizations, and corporations issued guidelines or policy documents on AI ethics and governance, AI scientists and engineers are routinely asked to include a reflection on the social and ethical dimensions of their work in scientific papers. It is now widely recognized that AI technologies present distinctive ethical challenges that need to be carefully taken into account in their further development.

A host of issues have been highlighted and examined in the ethical reflections on AI. Some of these issues have primarily to do with potential negative impacts on individuals, such as drastic invasion of privacy by AI-enabled surveillance, subtle but extensive manipulations of behaviour by recommender systems or targeted advertising, and unfair treatment of individuals due to entrenched or amplified biases in algorithmic classification and decision-making. Other issues are formulated and discussed at an aggregate level, including debates about the environmental costs and sustainable development of AI technologies, about the disruptive macroeconomic effects of large-scale automation, and about the possible exacerbation of socioeconomic inequality resulting from the unequal distribution of benefits. Still others are intimately related to deep philosophical questions on the defining characteristics of moral agents and patients, concerning, for example, the status of intelligent programs and robots in moral deliberations and the nature or ideal of human-machine relations.

We are concerned with AI-based art technologies rather than AI technologies in general. Even with this restriction, many issues highlighted in general AI ethics still arise in one form or another. However, some of them take essentially the same form in the context of art technology as they do in other domains of AI applications. For example, some generative techniques for symbiotic art creation require biometric and brain data that encode aspects of people's cognitive and affective states. Such data are sensitive in that they may be used to identify individuals. Moreover, even when these data are appropriately anonymized, their potential to reveal emotional conditions still raises questions about privacy, at a group level if not an individual one, as some scholars have argued. Important as it is, however, this matter does not assume a distinctive shape for generative art, and should be addressed in the same way as it is in affective computing or emotional AI. As another example, the issue of energy consumption in building large-scale machine learning models or blockchain-based marketplaces is certainly relevant to sustainable development of art technologies. But this concern is no different for the applications pertaining to art than for other applications of these techniques.

For the present purpose, we will set aside those issues that do not raise distinctive questions for artistic AI, and focus instead on three topics that are either specific to AI-based art technology or display special features in this context. First, we consider the authorship and ownership of artworks generated by AI systems, a topic that has obvious practical relevance, as well as connections to debates about the moral standing of AI and about the nature of the human-machine relationship. Second, we examine certain kinds of algorithmic bias that have been observed in generative art and raise questions about their significance. Third, we discuss the tendencies to democratize art of the new and emerging technologies, and stress the feasibility and importance of adopting a rich conception of democratization.

# 1 Authorship and Ownership of AI-generated Works of Art

The notion of authorship may appear fairly straightforward. In both common sense and legal practice, the authorship of an expression, be it a piece of artwork or a mundane instance of communication, seems simply to be aligned with the *production* of the expression. The author of a work is whoever is responsible for producing the work. Even if we adopt this simple idea, however, it is already unclear who the author is for a work that is generated by a machine learning model such as a pretrained GAN. On the one hand, the computer program appears to have a better claim for producing the work than a user of the program or the programmer, for the computer program is the proximate cause, and neither the user nor the programmer has much control over its output. On the other hand, the kind of production implicated by authorship seems to require more than a causal process. Intuitively, an author of an artwork must have *intended* the work to be produced, in order to be "responsible for" the production of the work. It may seem far-fetched to say that the current generative systems, impressive as they are, have artistic intentions.

The emphasis on the necessity of artistic intentions for authorship is not just an intuition. It is also a prominent, considered view in the philosophical literature on authorship. According to this view, authorship is important for mainly two reasons. First, authorship is needed to attribute responsibility; it is the author of a work who can be appropriately praised or blamed, awarded or punished for the content and quality of the work. Second, information about authorship is relevant to art appreciation. To interpret an artwork, in this view, is at least in part to figure out some meanings intended by the author. Neither of these reasons seems to make much sense if the author of an artwork does not need to harbour artistic or expressive intentions.

Does this intentionalist view, if correct, disqualify the current AI systems from authorship, especially when we move more and more to the era of autonomous artists? Not necessarily. It depends on whether the apparently far-fetched attribution of artistic intentions to an AI system is in fact tenable. There is at least one influential theory in the philosophy of mind and cognitive science that is friendly to attributing intentions to sufficiently complex machines, i.e., the intentional systems theory championed by Daniel Dennett. On this theory, a system is intentional just in case its behaviors can be efficiently and successfully predicted or accounted for from an intentional stance, a stance that predicts or explains behaviors by attributing such mental states as beliefs and desires. Thus a highly competent game playing program such as AlphaGo can be attributed intentions, because its moves can be largely predicted or rationalized from an intentional stance. Even a move that appears exceedingly novel to human players is at least explicable in terms of a "belief" that the move has the highest chance of winning the game and a "desire" to win the game. In principle, more specific and short-term intentions are also attributable when AlphaGo places a stone, such as the intention to create a ko threat.

Can an art generation system be similarly treated from an intentional stance? One difficulty is that unlike the setting of a strategic game, in which there are clear final and intermediate goals, it is usually unclear what artistic or aesthetic objectives an autonomous art system is built to achieve, and what moves or choices in artistic terms are available to the system. In the absence of such interpretations, an audience cannot begin to attribute or decipher intentions, and the notorious opaqueness of the state-of-the-art generative models based on deep learning aggravates the difficulty.

This perspective suggests from another angle the importance of human-machine interaction in developing machine artist systems employing the technologies from explainable AI and aesthetic computing by attributing artistic intentions to AI systems. As stated in part 3, in analogy to strategic games, we can reformulate the task of art content generation as a (sequential) decision problem, with a space of artistic options and an aesthetic evaluation network, and develop techniques based on the reformulation to interpret or explain the output of a machine artist system. If an output artwork can be reconstructed or mimicked by a series of artistic choices that contribute to the aesthetic value according to the evaluation network, that would provide compelling evidence that the system is interpretable from an intentional stance and hence is an intentional system. In addition to making room for AI authorship by the intentionalist standard, an interpretable or explainable generative system will probably be useful for other purposes as well, such as improving the user or audience's aesthetic experience or exposing subtle biases.

Even if all these are achieved, it is unlikely to resolve all controversies regarding AI authorship. After all, the intentional systems theory is a highly controversial account. A common complaint is that this account theorizes about "as if" or metaphorical intentionality rather than true or literal intentionality. For the present purpose, there is no need to consider this objection in any detail. But it is worth noting that the intentionalist account of authorship is of course not the only game in town either. One alternative approach, interestingly, takes authorship as a fictional or "as if" matter. In this approach, the author of a work is in an audience's imagination. If so, whether an AI system can be an author is more directly a matter of interpretation.

An even more radical view regards authorship as a social construction, having little to do with the actual creative intentions and having everything to do with the prevailing social institution and cultural practice of authorship attribution. This view strikes many as counterintuitive, but it motivates useful studies of the practice of authorship attribution. In the context of AI art, it is potentially illuminating to study a person's tendency to grant or deny authorship to AI, in relation to the person's aesthetic experience on the one hand and the person's perception of AI's moral status on the other. There is some empirical evidence suggesting that people tend to underappreciate artworks *labelled* as AI arts, in comparison to the same works *labelled* as human- made. A hypothesis worth testing is that such underappreciation is correlated with a tendency to deny AI authorship and a tendency to treat an AI system as an advanced tool rather than an autonomous agent.

To summarize, we believe that there is room for attributing authorship to current AI systems, even by the intentionalist standard, which is probably the most demanding. However, to make a convincing case, the systems need to become aesthetic-aware and explainable. A separate question is whether it is desirable or beneficial to attribute authorship to AI. There are vocal objections to, for example, developing robots to be treated as moral agents or patients. These objections probably weigh against attribution of AI authorship, because an author is in common parlance someone with certain rights and obligations. However, our view is that attribution of AI (co-)authorship will probably help to enhance appreciations of generative art and of art technologies. It seems to us harmless to assign authorship for the sake of art appreciation, without at the same time attaching moral responsibility.

In fact, the role of taking responsibility can be served by the related notion of ownership. In a way, ownership is an even more pressing problem for AI art than authorship, because it has more immediate connections to financial and other benefits. The recent years have seen several widely publicized controversies. In 2018, for example, a painting generated by a GAN model, *Edmond de Belamy*, was sold for $432,500 in a Christie's auction. However, the alleged human authors of the painting only contributed by tweaking some open source code. The bulk of the work, including the design and implementation of the network, and even the scraping of training data, was done by others who did not get any share in the lucrative gain. This case is often framed as a controversy about authorship, but the underlying question is who should enjoy the rights of ownership regarding the generated work.

Another case that explicitly raised copyright questions happened in 2019, when a conceptual artist sampled images from GANbreeders, an online tool that allows users to mix different images as "parental genes" and produce novel images as "children". The artist thought the images he selected were randomly generated by the tool, and he hired painters to paint them on canvas and planned an exhibition in an art gallery. However, some of the images he used were actually designed by GANbreeders users, who chose the "genes" for the images. Whether there is any copyright infringement in this case (or whether it should be regarded as infringement) attracts quite some discussions, and is again a question about ownership.

Ownership and authorship are related but distinct notions. As things currently stand, it seems pointless to consider ownership by AI, even though as suggested above, it is possible and potentially desirable to sometimes attribute authorship to AI. The current copyright laws in most jurisdictions do not protect AI-generated arts unless there is a sufficient degree of human intervention. This seems sensible to us, both because an ownership by AI is not yet meaningful and because human intervention is typically present. The difficult question is how to allocate rights and benefits among contributors to an AI-generated artwork, which usually comprise people working at different places and times. In the case of *Edmond de Belamy*, potential claimants include at least the designer of the network and the primary coder who implements the network, and perhaps also the contributors to the training data and even the original inventors of GAN. How to make copyright laws suitable for the age of AI art is an urgent challenge for the legal community to meet. As we described in the previous part on NFT, perhaps the new technology such as smart contract and NFT can provide new insights and mechanisms to provide a solution. It is feasible to consider that new paradigm of "computational copy right" where the machine creativity as well as the human contributions can be quantified during a creation process so that copy right of a generated art work can be divided accordingly.

## 2   Algorithmic Bias in Art Generation

As one of the most salient ethical issues in AI applications, algorithmic fairness has been the subject of much discussion in recent years, attracting both technical treatments and philosophical inquiries. A variety of algorithmic biases are distinguished in the literature. Some, such as aggregation bias and confounding bias, are primarily epistemic and affect the quality of inferences. Others are rooted in datasets and are ethically speaking more significant. For example, a common worry is that datasets may miss under-represent minority or marginalized groups. As a result, models trained on such datasets deliver much worse performances on some groups than on others, leading sometimes to embarrassing, offensive, or harmful errors. A different concern is historical bias: even though data are representative, the reality being accurately represented may be shaped by past prejudice. Such biases can be reinforced or amplified in algorithmic decision making.

In the context of content generation, the best known examples of bias are those observed in large- scale language models. These models are shown to pick up biased language, including racist and sexist vocabularies and expressions, as well as misleading and undesirable stereotypes. Very recently, some biases in generative visual art are also documented. One conspicuous example is that in several GAN-based style transfer apps, black faces in an input are typically lightened up or transformed in other noticeable ways in the output, whereas white faces are preserved.

It is not entirely clear how such biases in generative art should be treated. In terms of the risk of causing harm, these biases are not nearly as consequential as biases in other domains, such as the algorithmic assessment of the chance of recidivism or that of default. Moreover, it may be argued that aesthetic evaluation is inherently subjective and biased to a good extent. An aesthetic preference for light skin is as normal and unproblematic as a taste for vanilla ice cream. Still, if such biases are demonstrably instrumental in the further entrenchment of false or misleading societal stereotypes or the further marginalization of non-mainstream culture or the compromising of diversity in aesthetic experiences, it is probably reasonable to regard them as undesirable and to look for ways to mitigate them. Once again, to make generative models more explainable or interpretable is likely to be useful for detecting and treating biases.

When biases in AI art lead to offensive or morally flawed works, a question about censorship may arise. It is a contentious question when it is reasonable to restrict or censor offensive speeches, and there is a perennial debate in philosophical aesthetics about the value or disvalue of immoral art. In the human world these questions concern the freedom of expression, and are infected with all the heated controversies over the proper bounds or limits of that freedom. In the current stage of AI, however, these questions are probably easier to answer. As is widely believed, the current AI systems do not have expressive intentions (even if they can be attributed artistic intentions, as discussed in the previous section). If so, the issues regarding the freedom of expression and the legitimacy of censorship do not arise. It is thus no surprise that offensive chatbots, once spotted, are taken offline immediately.

## 3   Democratization of Art with New Technologies

Good technologies empower human beings. Depending on how selective or inclusive the empowering is, a new technology may exacerbate inequality or facilitate democratization. It is an uncontroversial dictum that art technologies, as is the case with most other technologies, should be made as widely accessible as possible so that people who are willing to take advantage of these technologies have decent opportunities to master and use them. This equity of access is no doubt difficult to achieve, but there is a general consensus that it is an ideal to aim for. Assuming this ideal is approximately achievable, it is natural to ask how various art technologies may serve to promote the democratization of art and whether the resulting forms of democratization, if realized, are desirable.

The notion of democratizing art is now multifaceted. A traditional conception of this notion, which can be traced back at least to the 19th century, has to do with the enlargement and diversification of the audience of high art. In this conception, art making remains elitist, and so does the collection of or investment in artworks; it is only the appreciation or consumption of art that needs to be democratized. However, with the advent of powerful tools for symbiotic art creation and NFT based decentralised marketplace, the democratization of art production and art collection is also a salient prospect today. Below we outline some ways in which art technologies are likely to democratize various elements of the art world. In our view, if the ultimate goal of democratizing art is to enable more people from diverse walks of life to engage with arts and benefit from such experiences, it is best served with a multi-pronged approach that covers art making and art collection as well as art appreciation.

Since the traditional target of democratization is art appreciation and education, a main approach to democratizing art or high culture centres around art museums or galleries, especially publicly funded ones. The idea is to incentivize museum curators to design blockbuster exhibitions that would attract a large number of visitors, with a broad range of demographic profiles. The success of such attempts rarely has unambiguous evidence; even when the number of visitors increases significantly, the demographic profiles tend to be still skewed toward socially and economically privileged groups. However, there is a widespread perception that interactive elements and immersive experiences tend to attract audiences who are otherwise uninterested in visiting museums. As a result, there has been a steady increase of art institutions that aim to enrich and personalize visitors' experiences with substantial interactive and immersive modes. Digital technologies of course play essential roles in these efforts, including especially advanced techniques of VR and AR. When combined with internet technologies and digitalization of artworks, they enable online virtual exhibitions that are comparable to physical ones in terms of the experiences they stimulate. The value of such technological capabilities has been amply demonstrated during the covid pandemic. In the post-pandemic world, it's probable that such technology-facilitated virtual and distributed exhibitions will continue to play a vital role in enhancing the accessibility of the art museums and galleries' offerings in a sustainable way. After all, blockbuster exhibitions in brick-and-mortar institutes are too costly from an environmental and ecological point of view.

Regarding the incorporation of interactive elements in art appreciation, a theoretical perspective is that an audience thereby takes control of some aspects of their aesthetic experience and becomes to a certain extent a performer as well as a spectator in their engagement with the artwork. This double-role status is supposed to contribute to the ideal of democratization because it implies more autonomy or freedom on the part of the audience in the process of interpreting and appreciating art, as compared to a purely passive observer. It is unclear whether and when interactive experiences in a museum or a gallery are really sufficient to elevate the status of an audience to that of an active performer in an artistically relevant sense, but this perspective suggests an important insight in our view, the insight that genuine democratization of art is not simply a matter of widening the range of passive audiences or consumers of art. It should also pay attention to the creative and productive side of the process.

Why then is art making not given due attention in the discourse on democratizing art? Presumably because the production of art is conceived of as a highly professional, creative, and skill-intensive enterprise. It is already a formidable task to extend the base who possess the relevant knowledge and temperament to understand and appreciate high art; it is next to impossible to make serious art creation an opportunity for all. Moreover, it appears doubtful that democratization of art making, even if practicable, is a desirable goal to pursue. One may worry, for example, that democratization in this vein will potentially encourage mediocrity in art practice.

However, in our conception, to democratize art making is not to endow everyone with a potential to become a full-blown artist in the traditional sense, let alone a lonely creative genius, but rather to distribute widely the opportunities for playing a non-trivial part in some process of art creation. The rise of generative art shows that artistic creativity in our era may be better reflected in a generative process than in the final outputs of that process. Anyone who plays an essential role in the process, such as the designer or programmer of a generative algorithm, is a contributor to art making. More generally, with AI-based smart tools for symbiotic art creation, the technical threshold for making a novel contribution to a creative process can be significantly lowered. The hope is that without much formal training in art making, a reasonably educated person can work with user-friendly and affordable art generation tools and make an interesting contribution in a collaborative creative process.

This is a feasible sense of democratizing art making, and also a desirable one, given the inevitable trend of accommodating AI in the art world. If the risk of encouraging mediocre arts is real (which we doubt), it is already carried by the rapid development of AI-generated art. Since AI art is definitely here to stay, there is no extra worry of this kind to exploit the technology for the purpose of democratizing creativity in a modest sense. On the other hand, the experience of participating in a creative process is likely to be useful for cultivating aesthetic sensitivity and artistic interest. It will therefore serve a valuable educational function that is instrumental to enhancing art appreciation. Moreover, the aforementioned rationale for building substantial interactive modes into the experience of art appreciation lends even stronger support to the idea of enabling a wide participation in art making. If democratization of art is in part a matter of nurturing a degree of artistic autonomy or freedom in diverse types of people, involving them in art creation is probably an even better strategy than merely giving them choices on how to engage with existing artworks.

The emphasis of freedom in artistic activities can also mitigate the following criticism of the traditional approach to democratizing art, namely that it is based on the premise that art or high culture as presented in major museums ought to be valued by everyone, that we all should aspire to understand and appreciate high art as defined by the experts. According to the critics, this premise is both questionable and condescending. Whether or not this perception of presumptuousness is fair, it is desirable to enrich the conception of democratization to overcome this perception. To highlight the importance of active and free participation in creative processes is a promising way to serve this purpose. This is another reason why democratization of art making is important for that of art appreciation and consumption.

In addition to art consumption and art creation, a third pillar in the democratization of art is art collection. Compared to the other two, art collection is by far the most elitist or exclusive activity in the traditional art world, due to the financial resources it requires. However, as is now widely anticipated, asset tokenization will trigger a radical transformation of the patterns and modes of investment in artworks. In the form of NFTs, it is now possible to own a certain share of a piece of fine art at an affordable cost. The foremost bottleneck for democratizing art collection is thus breakable. In addition to the benefit of enabling vastly more individuals to invest in art, the tokenized art market is likely to have several other positive effects. For example, art galleries can now choose to sell only parts of an artwork as tokens and still keep the artwork for offline exhibitions. This added liquidity implies more flexibility in the operation of an art gallery, which, in turn, will probably benefit artists and audiences who rely on galleries. Similarly, the rich people can now further diversify their art collections or investment portfolios, which, in turn, will support more artists and promote diversity in art practice. More generally, democratization of art collection is expected to provide extra impetus to democratization of art making. After all, few things work better than recognition in financial terms to stimulate interest and confidence.

Needless to say, it is again the new technologies that put these exciting prospects on the table. The blockchain technologies on which NFTs are based, and the envisaged Web3.0 that promises to realize a truly decentralized yet trustable P2P global network, are expected to be the major catalysts for democratizing many aspects of the internet. The optimism derives from the fact that with smart contracts, consensus mechanisms, and cryptographically secured transactions, we now seem to have the technological means to meet a fundamental challenge in human interaction, the challenge of ensuring the trustworthiness of total strangers without appealing to a central authority. When this problem is solved, a collective and decentralized mode of governance, that is, a democratic mode of governance, becomes feasible.

For the art world, therefore, the blockchain and related technologies enable new settings of the art market that have the tendency to democratize art collection and investment. This third prong of democratization is not only important in its own right, due to its connection to economic inequalities; we expect it to be also instrumental to the other two prongs of the democratization of art. The potential reinforcement between art making and art buying is intuitive. Moreover, it is a plausible hypothesis that opportunities to invest profitably in artworks usually lead to or increase aesthetic interests.

To summarize, we favour a thick conception of democratization of art that does not focus exclusively on enlarging and diversifying the audiences of art. It is instead conceptualized as a three-pillar enterprise, spanning art appreciation, art creation, and art investment. New and emerging technologies play important roles along all these dimensions and are indispensable for the second and the third.

# Acknowledgement

**CO-AUTHORS**

Yike GUO, *HKUST*
Qifeng LIU, *HKUST*
Jie CHEN, *HKBU*
Wei XUE, *HKUST*
Jie FU, *HKUST*
Henrik JENSEN, *Imperial College London*
Fernando ROSAS, *Imperial College London*
Jeffrey SHAW, *HKBU*
Xing WU, *Shanghai University*
Jiji ZHANG, *HKBU*
Jianliang XU, *HKBU*

**MAIN CONTRIBUTORS** (in alphabetical order)

Caroline LI, *City, University of London*
Drew CAPPOTTO, *CityU*
Haoqi WANG, *SenseTime*
Hongning DAI, *HKBU*
Johnny POON, *HKBU*
Junkun JIANG, *HKBU*
Likai PENG, *HKU*
Louis NIXON, *Norwich University of The Arts*
Maosong SUN, *Tsinghua University*
Pan WANG, *Delft University of Technology*
Qi CAO, *HKUST*
Rongjun YU, *HKBU*
Ryan AU, *HKBU*
Sara WANG, *HKBU*
Wanyu CHENG, *HKUST*
Yiwen WANG, *HK Chu Hai College*
Zhe PENG, *PolyU*
Zhizhong LI, *SenseTime*

## Team Members (in alphabetical order)

Chi-min CHAN, *HKUST*
Jiahao PAN, *HKUST*
Jianyi CHEN, *HKUST*
Junkun Jiang, *HKBU*
Lujun LI, *HKUST*
Mengfei LI, *HKUST*
Min ZENG, *HKUST*
Peng LI, *HKUST*
Ryan Au, *HKBU*
Ruibin YUAN, *HKUST*
Shixin CHEN, *HKUST*
Wanyu CHENG, *HKUST*
Wenpeng Xing, *HKBU*
Xingwei QU, *HKUST*
Xingqun QI, *HKUST*
Yiwen LU, *HKUST*
Xiaowei CHI, *HKUST*
Xi CHEN, *HKUST*
Ziya ZHOU, *HKUST*
Zeyue TIAN, *HKUST*
Zeyu HUANG, *HKUST*
Zheqi DAI, *HKUST*
Zhen YE, *HKUST*
Zhenghao ZHU, *HKUST*